\pdfoutput=1

\documentclass[11pt]{article}

\usepackage[final]{coling}

\usepackage{times}
\usepackage{latexsym}

\usepackage[T1]{fontenc}

\usepackage[utf8]{inputenc}

\usepackage{microtype}

\usepackage{inconsolata}

\usepackage{graphicx}
\usepackage{xcolor}

\usepackage{amsmath}
\usepackage{amssymb}

\usepackage{subcaption}
\usepackage{graphicx}

\usepackage{tabularx,tabulary, booktabs}
\usepackage{siunitx}
\usepackage[overload]{ragged2e}
\usepackage{array}
\newcolumntype{L}[1]{>{\raggedright\let\newline\\\arraybackslash\hspace{0pt}}m{#1}}
\newcolumntype{C}[1]{>{\centering\let\newline\\\arraybackslash\hspace{0pt}}m{#1}}
\newcolumntype{G}[1]{>{\raggedleft\let\newline\\\arraybackslash\hspace{0pt}}m{#1}}

\usepackage{enumitem}

\definecolor{figure-blue}{RGB}{108, 142, 191}
\definecolor{figure-yellow}{RGB}{215, 155, 0}
\definecolor{figure-green}{RGB}{130, 179, 102}

\newcommand{\change}[1]{#1}

\title{Analyzing the Attention Heads for Pronoun Disambiguation\\
in Context-aware Machine Translation Models

}

\author{Paweł Mąka \and Yusuf Can Semerci \and Jan Scholtes \and Gerasimos Spanakis\\
        Department of Advanced Computing Sciences \\ 
        Maastricht University \\ 
        \texttt{\{pawel.maka, y.semerci, j.scholtes, jerry.spanakis\}@maastrichtuniversity.nl}}

\begin{document}
\maketitle

\begin{abstract}
In this paper, we investigate the role of attention heads in Context-aware Machine Translation models for pronoun disambiguation in the English-to-German and English-to-French language directions. We analyze their influence by both observing and modifying the attention scores corresponding to the plausible relations that could impact a pronoun prediction. 
Our findings reveal that while some heads do attend the relations of interest, not all of them influence the models' ability to disambiguate pronouns. We show that certain heads are \textit{underutilized} by the models, suggesting that model performance could be improved if only the heads would attend one of the relations more strongly. 
Furthermore, we fine-tune the most promising heads and observe the increase in pronoun disambiguation accuracy of up to 5 percentage points which demonstrates that the improvements in performance can be solidified into the models' parameters.

\end{abstract}

\section{Introduction}

In Context-Aware Machine Translation (MT), the context sentences are available to the system and can be used to maintain coherence of the translation and to resolve ambiguities \citep{agrawal2018contextual, bawden-etal-2018-evaluating, muller-etal-2018-large, voita-etal-2019-good}. Both the source-side (sentences in the source language) and target-side context (the previously translated sentences) can be used as context. Although many novel architectures have been proposed to tackle the task of Context-aware MT \citep{tu-etal-2017-context, bawden-etal-2018-evaluating, miculicich-etal-2018-document, maruf-etal-2019-selective, huo-etal-2020-diving, zheng2021towards}, we limit our investigation to the standard Transformer \citep{vaswani2017attention} architecture (often referred to as the \textit{single-encoder} architecture), because of its simplicity and demonstrated high performance \citep{sun-etal-2022-rethinking, majumde2022baseline, gete-etal-2023-works, post2023escaping, mohammed-niculae-2024-measuring}.

\begin{figure}
    \centering
    \includegraphics[width=0.93\linewidth]{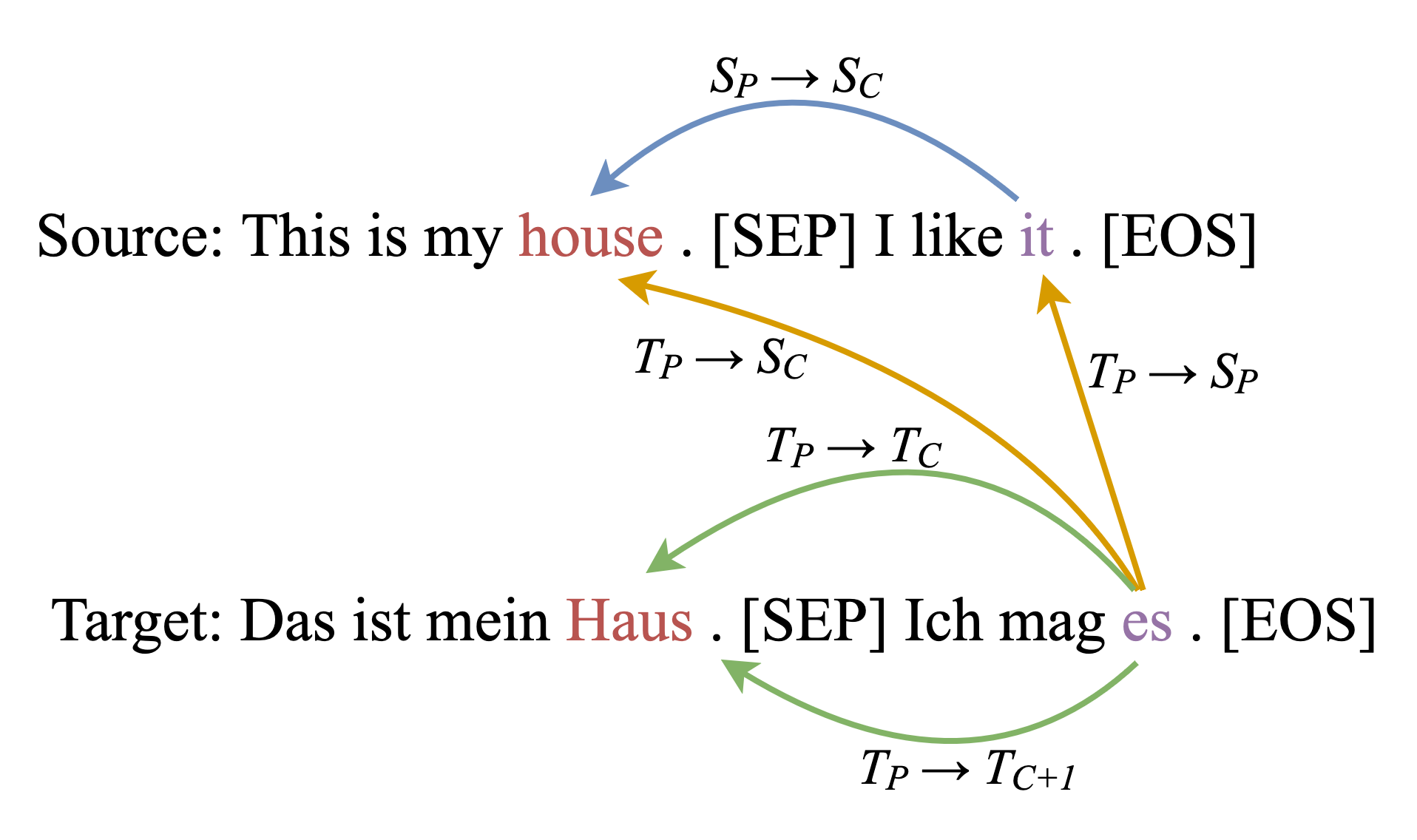}
    \caption{The types of relations we investigate. $S_P$ and $T_P$ mark a context-dependent word (e.g., pronoun) on the source- and target-side respectively, and $S_C$ and $T_C$ mark the source- and target-side context cue (e.g., antecedent). For the target side, the words represent tokens predicted by the model, and $T_{C+1}$ is the token corresponding to the antecedent as the input. Colors of arrows designate self-attention (\textcolor{figure-blue}{blue}), cross-attention (\textcolor{figure-yellow}{yellow}), and decoder-attention (decoder self-attention, \textcolor{figure-green}{green}).}
    \label{fig:relation-types}
\end{figure}

For the Transformer model to be successful in the task of Context-aware MT, it has to integrate the contextual information and utilize it to produce the translation autoregressively. During inference, the contextual information has to come through the mechanism of multi-head attention. Motivated by the findings that attention heads can learn to perform seemingly specific functions \citep{clark-etal-2019-bert, voita-etal-2019-analyzing, jo-myaeng-2020-roles, Olsson2022IncontextLA}, we hypothesize that certain heads in a model can be crucial for the context utilization. We study the case of pronoun disambiguation, where the selection of the correct pronoun is dependent on the antecedent (contextual cue), which can be present in the previous (context) sentences. Therefore, we analyze the translation models through the lens of the attention given by the attention heads to the plausible relations that could influence the prediction of a pronoun: pronoun-antecedent on the source and target sides, and pronoun-pronoun and pronoun-antecedent between target and source sides. 
Figure~\ref{fig:relation-types} illustrates the relations we consider.


Our analysis starts with measuring the attention scores given by each head to the relations we consider when evaluated on the contrastive datasets - ContraPro \citep{bawden-etal-2018-evaluating} \change{for the English-to-German direction} and Large Contrastive Pronoun Testset (LCPT; \citet{lopes-etal-2020-document}) \change{for the English-to-French direction}. In these datasets, the model is presented with the task of ranking several translations of the same source sentence with the same context. The provided translations differ only partially, and the provided context is required to choose the correct translation. Next, we correlate the scores with the model being correct when evaluated on the contrastive example. Furthermore, we artificially modify the attention scores of the heads and measure the difference in the accuracy on these datasets. Lastly, we fine-tune the selected heads to attend the relations more strongly.


\section{Related Work}

\subsection{Context-aware Machine Translation}

A direct approach to include previous sentences as context in MT is to concatenate them with the current sentence. This method is often called the \textit{single-encoder} architecture, as the basic encoder-decoder architecture is used \citep{tiedemann-scherrer-2017-neural, ma-etal-2020-simple, zhang-etal-2020-long}. The \textit{multi-encoder} approach is to encode the context sentences by a separate encoder \citep{jean2017does, miculicich-etal-2018-document, maruf-etal-2019-selective, huo-etal-2020-diving, zheng2021towards}. Existing studies also investigated more exotic architectures \citep{miculicich-etal-2018-document, bao-etal-2021-g, chen2022one, maka-etal-2024-sequence}, post-processing translation \citep{voita-etal-2019-good, voita-etal-2019-context}, and employing a memory mechanism \citep{feng-etal-2022-learn, bulatov2022recurrent}. \citet{yin-etal-2021-context} applies attention regularization that concentrates on the context phrases marked as important by human translators which is similar to Head Tuning method in our work. 

In recent years, the single-encoder architecture has seen increased prominence in the literature because of its simplicity and robust performance \citep{majumde2022baseline, gete-etal-2023-works, post2023escaping, mohammed-niculae-2024-measuring}, even on long context sizes (of up to 2000 tokens) when data augmentation was used \citep{sun-etal-2022-rethinking}. For this reason, this paper focuses on this architecture.


Studies have shown that the sentence-level metrics (such as BLEU \citep{papineni-etal-2002-bleu}) are not well suited to measure the translation quality with respect to context usage \citep{hardmeier2012discourse, wong-kit-2012-extending}. To address this issue, researchers introduced new metrics \citep{fernandes-etal-2021-measuring, fernandes-etal-2023-translation} and context-sensitive datasets \citep{wicks-post-2023-identifying, fernandes-etal-2023-translation}, including contrastive datasets \citep{muller-etal-2018-large, bawden-etal-2018-evaluating, voita-etal-2019-good, lopes-etal-2020-document}. In this study we employ two of them: ContraPro \citep{muller-etal-2018-large} and Large Contrastive Pronoun Testset (LCPT; \citet{lopes-etal-2020-document}).

\subsection{Explaining Models}

Explaining models' decisions or behaviors is an important issue from the safety and ethical considerations \citep{madsen2022posthoc}. Additionally, we argue that they can also be valuable from the engineering perspective, where the shortcomings of the models can potentially be addressed when they are brought to light. 
Numerous methods explaining Natural Language Processing models have been proposed \citep{bau2019identifying, toneva2019interpreting, ferrando-etal-2022-towards, langedijk-etal-2024-decoderlens, meng2024locating}.

Even though some works have argued against using raw attention scores as explanations themselves \citep{jain-wallace-2019-attention} the attention mechanism is an important part of the Transformer and many researchers concentrated their efforts to understand its influence on the model's behavior \citep{wiegreffe-pinter-2019-attention, abnar-zuidema-2020-quantifying, kobayashi-etal-2020-attention, kobayashi-etal-2021-incorporating, bogoychev-2021-parameters, gheini-etal-2021-cross, mohebbi-etal-2023-quantifying}.
Our work is motivated by the findings suggesting that some heads have specific roles or functions \citep{clark-etal-2019-bert, voita-etal-2019-analyzing, Olsson2022IncontextLA} that can be linked to linguistic relations \citep{vig-belinkov-2019-analyzing, tenney-etal-2019-bert, jo-myaeng-2020-roles}. 


Several works have investigated the models in MT \citep{goindani-shrivastava-2021-dynamic, voita-etal-2021-analyzing, sarti2023quantifying, mohammed-niculae-2024-measuring}. In contrast, our work focuses on identifying where in the model's architecture the contextual information is integrated. 
We also investigate the influence of the alterations to the model 
(i.e., modifying the attention scores corresponding to the plausible relations) on the context usage.

\section{Methods}
\label{sec:methods}

We denote the tokenized source and target sentences of example $d$ in the dataset as $S^d$ and $T^d$ respectively. We define the sets containing the indices of contextually important tokens (context cues) on the source $S_C$ and target side $T_C$, and the sets containing the indices of contextually dependent tokens on the source $S_P$ and target side $T_P$. 
On the target side, the token indices denote the predicted tokens, rather than the input tokens, which are shifted right during training and contrastive evaluation. This way we analyze the model during the inference step when the context-dependent tokens $T_P$ are being predicted. Consequently, the context cue tokens on the target side $Y_C$ also mark the steps where those tokens are being generated by the model.
Because the input tokens have been found to retain their identity throughout the layers of the Transformer \citep{brunner2020identifiability}, we add the set $T_{C+1}$ referring to the contextually important tokens on the input. In the case of ambiguous pronouns, the context cues tokens are the antecedents and the contextually dependent tokens are the pronouns, both on the source and target side.

We investigate the relations between the following token sets (see Figure~\ref{fig:relation-types} for an example):
\begin{itemize}[topsep=3pt,itemsep=3pt,partopsep=0pt, parsep=0pt, itemindent=15pt, leftmargin=0pt]
    \item from source dependent to source important tokens ($S_P \rightarrow S_C$) - in the encoder's self-attention;
    \item from target dependent to source important tokens ($T_P \rightarrow S_C$) - in the decoder's cross-attention;
    \item from target dependent to source dependent tokens ($T_P \rightarrow S_P$) - in the decoder's cross-attention;
    \item from target dependent to target important tokens ($T_P \rightarrow T_C$) - in the decoder's self-attention;
    \item from target dependent to target important tokens on the input ($T_P \rightarrow T_{C+1}$) - in the decoder's self-attention.
\end{itemize}


We examine the models through three methods. \textit{Attention Scores} (Section~\ref{sec:attention-scores}) and \textit{Score-Accuracy Correlation} (Section~\ref{sec:correlation}) are based on observing the model's behavior during the task of contrastive disambiguation. The method \textit{Modifying Heads} (Section~\ref{sec:modifying-heads}) introduce a form of disturbance into the functioning of an attention head. 

\subsection{Attention Scores}
\label{sec:attention-scores}

To find the heads that learned to pay high attention to the relations of interest we measure and average the attention scores $Z^{l,h}$ of each head $h$ in every layer $l$
for all relations of interest. For the contextually important and dependent phrases that span over multiple tokens, we take the maximum score.


\subsection{Score-Accuracy Correlation}
\label{sec:correlation}

The fact that a particular attention head - on average - pays attention to the relation of interest does not necessarily mean that the head is crucial or helpful for disambiguation. Therefore, we want to measure how the head's attention scores of a relation correspond to the model correctly disambiguating a particular example. We define a variable $I$ as follows:
\begin{equation} \label{eq:correct-variable}
I^{d} = \begin{cases}
1, \text{if correctly scored}, \\
0, \text{otherwise}, \\
\end{cases}
\end{equation}
where $d$ is an example from the contrastive dataset $\mathcal{D}$. We calculate the point-biserial correlation coefficient between attention scores given by each head $Z^{l,h}$ and the variable $I$. The accuracy on the whole dataset can be calculated as $\sum_{d\in\mathcal{D}} I^d / |\mathcal{D}|$.



\subsection{Modifying Heads}
\label{sec:modifying-heads}

The goal of Modifying Heads is to adjust the behavior of the head in a controlled manner. It modifies attention scores from a particular token ensuring that the total attention score given to a target subset of tokens equals a desired value $C$ while preserving the pre-softmax attention scores $H$ for all other target tokens. Modifying Heads allows us to experimentally test the behavior of the model if one (or more) of its heads were better or worse at the function of attending to the context-informative tokens.
For simplicity, we jointly label the attending tokens (left-hand-side of the arrow $\rightarrow$) as $\mathcal{Y}$ and attended tokens (right-hand-side of the arrow $\rightarrow$) as $\mathcal{X}$. Consequently, $\mathcal{Y} \rightarrow \mathcal{X}$ can represent all investigated relations inside the corresponding attention module. 
Modifying Heads can be formulated as follows:
\begin{equation} \label{eq:modified-heads}
\begin{aligned}
&\tilde{H}^{l,h,d}_{i,j} = \log \Bigl( \frac{C}{|\mathcal{X}^d| (1-C)} \sum_{k \in X^d \setminus \mathcal{X}^d} \exp \bigl( H^{l,h,d}_{i,k} \bigr) \Bigr) \\ 
&\forall i \in \mathcal{Y}^d, j \in \mathcal{X}^d,
\end{aligned}
\end{equation}
where $\tilde{H}$ represent the updated pre-softmax attention scores,  $H$ are the original pre-softmax scores, $k \in X^d \setminus \mathcal{X}^d$ are the attended tokens not present in the subset of interest $\mathcal{X}^d$. The derivation of the formula can be found in Appendix~\ref{sec:modifying-heads-derivation}.

\section{Experiments}
\label{sec:experiments}

All our experiments are implemented\footnote{The code for this paper can be found on Github \url{https://github.com/Pawel-M/context-mt-attention-analysis}.} in \textit{Huggingface transformers} framework \citep{wolf-etal-2020-transformers}.

\subsection{Models}
In this  study, we use pre-trained single-encoder models for two reasons: to assess how models' abilities learned on intra-sentential phenomena will translate into the inter-sentential regime, and to analyze the robustly trained and widely tested models instead of training models from random initialization.
The first model we experiment with is \textit{OPUS-MT en-de}\footnote{\url{https://huggingface.co/Helsinki-NLP/opus-mt-en-de}} \citep{TiedemannThottingal:EAMT2020, tiedemann2023democratizing}. It is a relatively small (6 layers, 8 heads) encoder-decoder Transformer model trained on English-to-German translation. The second model is \textit{No Language Left Behind} (NLLB-200) \citep{nllb2022}, which is a multilingual MT model. We use the small distilled version with approximately 600 million parameters \footnote{\url{https://huggingface.co/facebook/nllb-200-distilled-600M}}, which consists of 12 encoder and decoder layers and 16 heads.

\begin{table*}[!ht]
\noindent
\begin{tabulary}
{\linewidth}{@{}l*{7}{R}@{}}
\hline
        \textbf{Model} & \textbf{Context} & \textbf{ContraPro Accuracy} & \textbf{ContraPro BLEU}  & \textbf{IWSLT en-de}  & \textbf{LCPT Accuracy} & \textbf{LCPT BLEU} & \textbf{IWSLT en-fr} \\
\hline
        Opus MT en-de &  0 &  ${}^\lozenge81.46\%$ &  ${}^\lozenge30.24$ &  $32.42$&  - & - & -\\
         &  $1$ &  $78.35\%$ &  $31.00$ &  $34.40$ &  - & - & -\\
         &  $3$ &  $79.08\%$ &  $31.35$ &  $34.59$ &  - & - & -\\
         \hline
         NLLB-200 &  $0$ &  ${}^\lozenge80.88\%$ &  ${}^\lozenge29.46$ & 32.31 &  ${}^\lozenge95.02\%$ & ${}^\lozenge33.64$ & 43.34\\
         &  $1$ &  $77.95\%$ &  $29.18$ &  $34.01$ &  $91.89\%$ & $31.49$ & 43.86 \\
\hline
    \end{tabulary}
    \caption{The results of the unmodified models. For the sentence-level models (with the context size of zero) we report accuracy and BLEU on contrastive datasets with the "antecedent distance" of zero (marked with ${}^\lozenge$; meaning that the antecedent is located in the current sentence.}
    \label{tab:base-results}
\end{table*}

\subsection{Context-aware Fine-tuning}

We fine-tuned both models for Context-aware MT by concatenating the previous sentences with the current sentence on both the source and target side. Each sentence is separated by the \verb|[SEP]| token (before fine-tuning, we expanded the vocabulary of the OpusMT en-de model with this token). We trained two versions of the OpusMT en-de model with maximum context sizes - the number of context sentences - of one and three (refered to as context-aware-1 and context-aware-3 models), and one version of NLLB-600M with the maximum context size of one. 
\change{We train the models with all the context sizes from zero to the maximum context size while keeping the context size the same on the source and target side for each example.}
During inference, we provide the models with the number of sentences equal to the maximum context size it was trained with.

For Opus-MT en-de, we used the train subset of the IWSLT 2017 \citep{cettolo-etal-2017-overview} English-to-German dataset, and interleaved English-to-German and English-to-French train subsets for NLLB-200. IWSLT 2017 is a document-level dataset (the details of the datasets are presented in Appendix~\ref{sec:datasets-details}). Further details and the hyper-parameters of the fine-tuning process are presented in Appendix~\ref{sec:fine-tuning-hyperparameters}. 

\subsection{Contrastive Datasets}

We used two contrastive datasets: ContraPro \citep{muller-etal-2018-large} for the English-to-German direction, and the Large Contrastive Pronoun Testset (LCPT; \citet{lopes-etal-2020-document}) for the English-to-French direction. Both are based on the OpenSubtitles 2018 dataset \citep{lison-etal-2018-opensubtitles2018}, and consist of the source sentence, the source- and target-side context with several translations differing only in a pronoun that requires context to be correctly translated. 
The details of the contrastive datasets are presented in Appendix~\ref{sec:datasets-details}.

For each model, we first calculated the average attention scores assigned by each of the model's heads to the relations of interest (as described in Section~\ref{sec:attention-scores}) for the examples from the contrastive datasets and calculated the point-biserial correlation between the measured attention scores given by each head to the relations of interest and the model correctly disambiguating a particular example (see Section~\ref{sec:correlation}). 
Next, we applied Modifying Heads method (see Section~\ref{sec:modifying-heads}) to each head of the models. We used the following values of $C$ (eq.\ref{eq:modified-heads}) $[0.01, 0.25, 0.5, 0.75, 0.99]$ (uniformly probing the available range) for OpusMT en-de models, and $[0.01, 0.99]$ for other models after observing that the relation between model's ContraPro accuracy and modified value was mostly monotonic for all heads (see Appendix~\ref{sec:expanded-results-opus-ende} for the detailed results).
Additionally, we disabled each head by assigning the same probability to all tokens (see Appendix~\ref{sec:disabling-heads} for the detailed formulation) but we found that the results match the results for modifying the head to $0.01$ without distinguishing the relation of interest.

\section{Results}
\label{sec:results}

The base results in terms of accuracy and BLEU \citep{papineni-etal-2002-bleu} on contrastive datasets and BLEU on test subset of the IWSLT 2017 dataset are presented in Table~\ref{tab:base-results}. Since sentence-level models receive only the current sentence as input, we evaluate them by extracting examples from contrastive datasets where the antecedent is in the same sentence as the pronoun. 
This way we ensure that the sentence-level models are tested exclusively on examples that they are capable to disambiguate. \change{We do not filter the examples for context-aware models to allow the comparison between the tested models (as well as the results in the literature).}

We analyzed the results in terms of average attention scores (Section~\ref{sec:attention-scores}), correlations between attention scores and correctly disambiguating an example (Section~\ref{sec:correlation}), and modifying heads (Section~\ref{sec:modifying-heads}) to $0.01$ and $0.99$.
The correlation is only partially corresponding to the changes in accuracy when modifying the attention scores, meaning that we find heads that demonstrate low correlation but have a high impact on the model's performance when modified. We ignore the losses in accuracy when modified to $0.99$ as possibly resulting from the decreased attention scores for other token-to-token relations, not investigated in this work. Similarly, we discard the increases in accuracy when modified to $0.01$. We broadly categorized heads' behavior for a particular relation into the following groups (note that the same head can be in different groups for different relations of interest):
\begin{itemize}[topsep=0pt,itemsep=0pt,partopsep=0pt, parsep=0pt, itemindent=15pt, leftmargin=0pt]
    \item \textbf{irrelevant} - those heads do not attend the relation of interest and do not respond to the modifications;
    \item \textbf{attending and fully responsive} - those heads do attend the relation, the model's accuracy decreases when modified to $0.01$, and increases when modified to $0.99$;
    \item \textbf{attending and negatively responsive} - the heads do attend the relation, and respond negatively to modifying to $0.01$ but do not show improvement when modified to $0.99$; we interpret those heads as already at the peak of their ability to help the model with pronoun disambiguation;
    \item \textbf{attending and positively responsive} - similar to the previous category but with only the positive response to modifying to $0.99$; we hypothesize that the heads in this category do not attend the relation to a sufficient degree for some examples and the correctly disambiguated examples are also supported by other heads in the model;
    \item \textbf{attending and non-responsive} - those heads do attend the relation but are not responsive to any modification; our interpretation is that those heads are important for other tasks than pronoun disambiguation;
    \item \textbf{non-attending and positively responsive} - the heads in this category are not attending the relation but positively influence the model's accuracy when modified to $0.99$.
\end{itemize}

In the following sections, we identify the heads by \textit{a-l-h}, where \textit{a} marks the attention module and can take values from \{\textit{e}, \textit{c}, \textit{d}\} meaning the encoder\mbox{-,} cross\mbox{-,} and decoder-attention respectively, \textit{l} and \textit{h} mark the layer and head number. For example, \textit{d-2-11} identifies the head number 11 in the self-attention module in the second decoder layer.


\begin{figure*}[!ht]
\centering
    \includegraphics[width=0.95\linewidth]{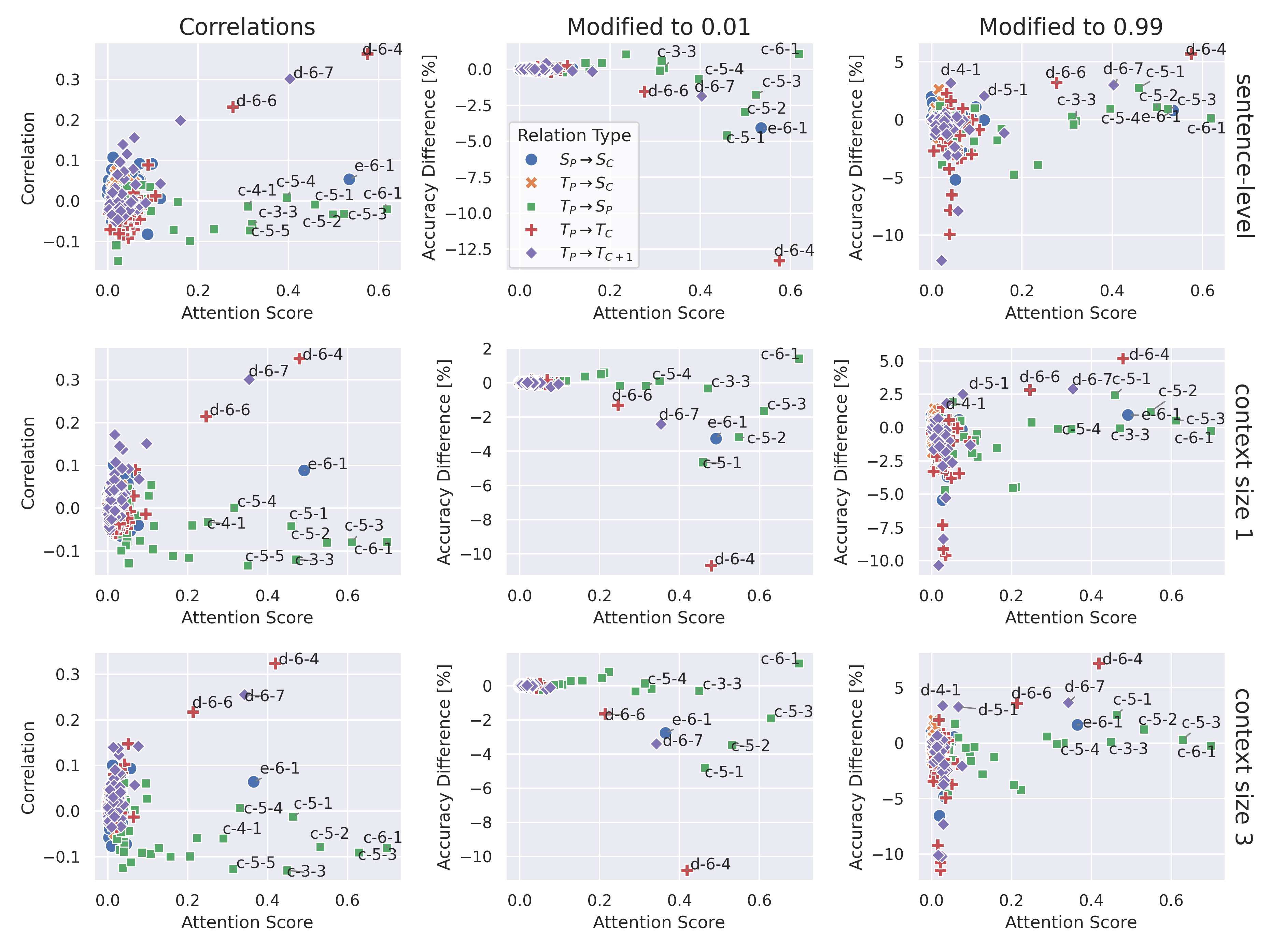}
    \caption{Results in terms of calculated metrics (correlations, difference in accuracy when modified to $0.01$, and modified to $0.99$; as columns) in relation to the averaged attention scores for the three models based on \textbf{OpusMT en-de} (sentence-level, context-aware-1, and context-aware-3; as rows).}
    \label{fig:opus-ende-relational}
\end{figure*}

\subsection{OpusMT en-de}
\label{sec:results-opus-mt}

We performed the analysis of the models based on OpusMT en-de (sentence-level, context-aware-1, and context-aware-3) on the ContraPro dataset. 
The sentence-level model was evaluated only on the examples where the antecedent was located in the current sentence. For the context-aware models, we employed the full dataset. 
We present the results for all three models in Figure~\ref{fig:opus-ende-relational}, where for each head we show the metrics introduced in Section~\ref{sec:methods} (correlations, accuracy when modified to $0.01$, and modified to $0.99$) in relation to the averaged attention scores. The expanded results 
can be found in Appendix~\ref{sec:expanded-results-opus-ende}.
\change{Additionally, to give a sense of the distribution of the results, we show histograms of the observed values of the metrics with annotated values corresponding to prominent heads in Appendix~\ref{sec:histograms}.}


Most heads do not pay - on average - a large attention to any of the relations of interest. 
Only a single encoder head (\textit{e-6-1}) was found to assign high attention scores to the $S_P \rightarrow S_C$ relation. It was negatively responsive for all models and positively responsive for the context-aware-3 model (increasing the accuracy by $1.6$ percentage points).
Head \textit{e-2-8} was non-attending and positively responsive (improvement of up to $2$ percentage points).


For cross-attention, we found that the $T_P \rightarrow S_P$ relation is attended by several heads (most prominently heads \textit{c-6-1}, \textit{c-5-3}, \textit{c-5-2}, \textit{c-5-1}, \textit{c-5-4}, and \textit{c-3-3}). 
Out of them only heads \textit{c-5-1} and \textit{c-5-2} (to a lesser extent) are responsive. 
While no head put large attention to the $T_P \rightarrow S_C$ relation, head \textit{c-5-8} was positively responsive. 

\begin{figure*}[!ht]
\centering
    \includegraphics[width=0.95\linewidth]{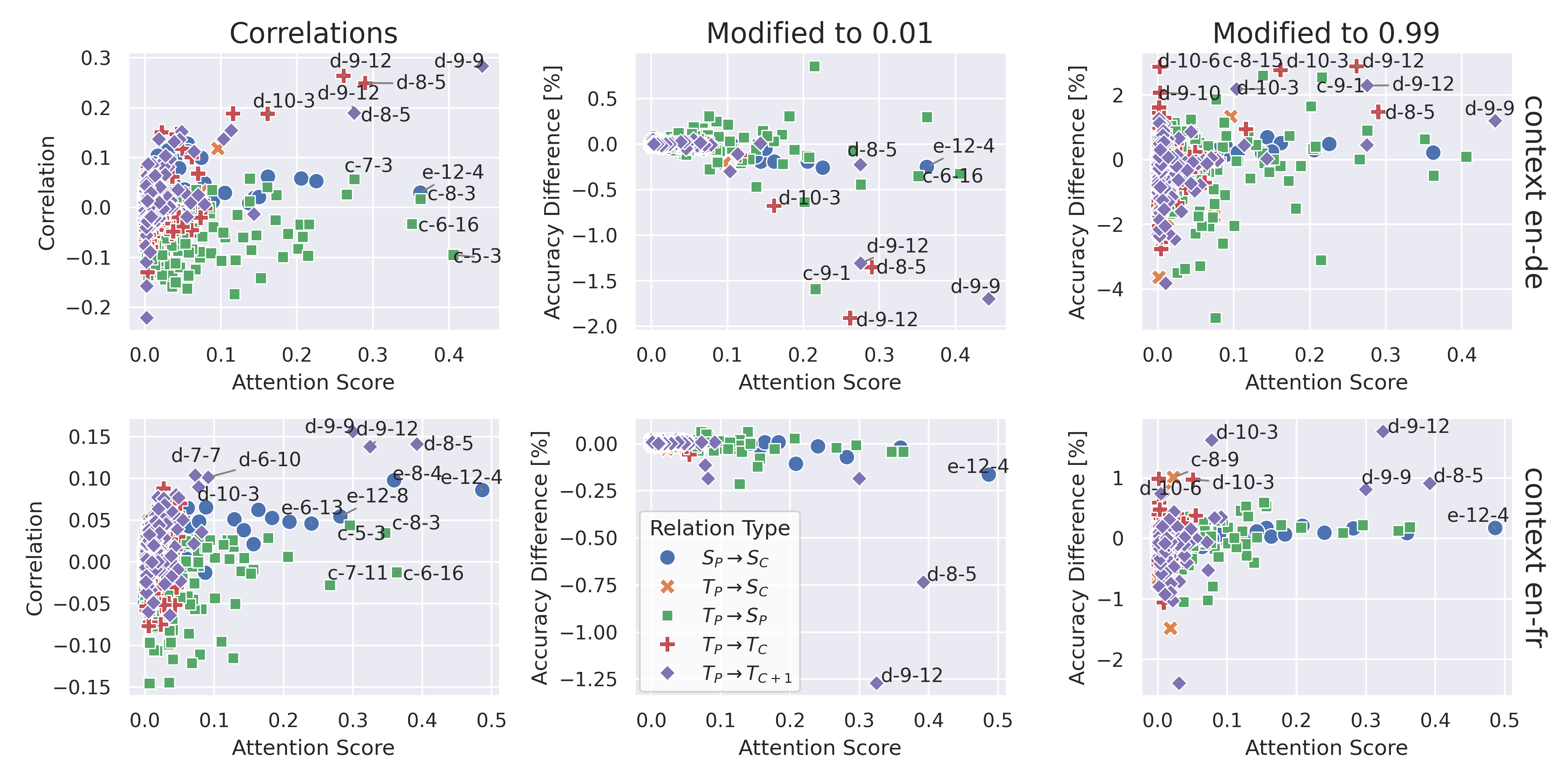}
    \caption{Results in terms of calculated metrics (correlations, difference in accuracy when modified to $0.01$, and modified to $0.99$; as columns) in relation to the averaged attention scores for the English-to-German and English-to-French directions (as rows) for the context-aware model based on \textbf{NLLB-200}.}
    \label{fig::nllb_600_combined}
\end{figure*}


The most responsive heads in the whole model are located in the decoder's attention. Two heads are attending and are responsive to the $T_P \rightarrow T_C$ relation.
The \textit{d-6-4} head shows the highest correlation (above 0.3 for all models) and responsiveness (accuracy difference of below $-10$ and above $5$ percentage points for modifying to $0.01$ and $0.99$ respectively). 
The \textit{d-6-6} head also shows relatively high correlation and responsiveness but not to the same extent.
For the $T_P \rightarrow T_{C+1}$ relation, the \textit{d-6-7} head is attending and responsive with high correlation. The negative responsiveness is higher for context-aware models compared to the sentence-level model.
This indicates that attending the context-informative token on the input can be beneficial but is not crucial to the sentence-level model. The context-aware models learn to rely on this relation more heavily. 
Additionally, heads \textit{d-4-1} and \textit{d-5-1} are non-attentive and only positively responsive.

\subsection{NLLB-200}
\label{sec:results-nllb}

The results of the analysis of the NLLB-200 context-aware models are presented in Figure~\ref{fig::nllb_600_combined}. The expanded results, including the sentence-level model, 
can be found in Appendix~\ref{sec:expanded-results-nllb-600M}.
\change{The histograms of the results are presented in Appendix~\ref{sec:histograms}.}

\subsubsection{English-to-German}


For the English-to-German direction, only a single encoder head - \textit{e-12-4} - was found to attend the $S_P \rightarrow S_C$ relation but was non-responsive. In the cross-attention modules, the $T_P \rightarrow S_P$ relation is attended by several heads in both models (most notably by \textit{c-7-3}, \textit{c-8-3}, \textit{c-6-16}, \textit{c-10-13}, and \textit{c-5-3} for context-aware model only) but showed low correlation and responsiveness. For the context-aware model, two heads visibly changed their behavior - head \textit{c-9-1} moderately attends the relation but does respond to modifying, and head \textit{c-8-15} is non-attentive and only positively responsive. The $T_P \rightarrow S_C$ relation was weakly attended by a single head (\textit{c-8-9}) which showed only the modest positive response.

Similar to OpusMT models, the most attentive and responsive heads are located in the decoder attention. Head \textit{d-10-3} is moderately attending the $T_P \rightarrow T_C$ relation, have relatively high correlation, and is highly responsive (improving the accuracy by almost $3$ percentage points). Surprisingly, when this head is modified to $0.99$ for the $T_P \rightarrow T_{C+1}$ relation it improves the accuracy of the context-aware model. The \textit{d-9-9} head attends $T_P \rightarrow T_{C+1}$ relation and exhibits the highest correlation among all heads of the model and responds negatively. 
Several heads are non-attentive but positively responsive. Most notably, head \textit{d-10-6} improves accuracy by $2$ and $3$ percentage points for sentence-level and context-aware models respectively, and head \textit{d-9-10} by approximately $2$ percentage points for both models.
Interestingly, head \textit{d-9-12} seems to pay attention to both relations and responds strongly to modifying to $0.01$. While it improves the accuracy for both relations when modified to $0.99$, the improvement is larger (by $3$ percentage points in both models) for the $T_P \rightarrow T_C$ relation than for the $T_P \rightarrow T_{C+1}$ relation with larger gains for context-aware model. Another head that pays attention to both relations is \textit{d-8-5} but it shows moderate correlation and responsiveness.







\subsubsection{English-to-French}

\begin{figure*}[!ht]
\center{}
    \includegraphics[width=0.95\linewidth]{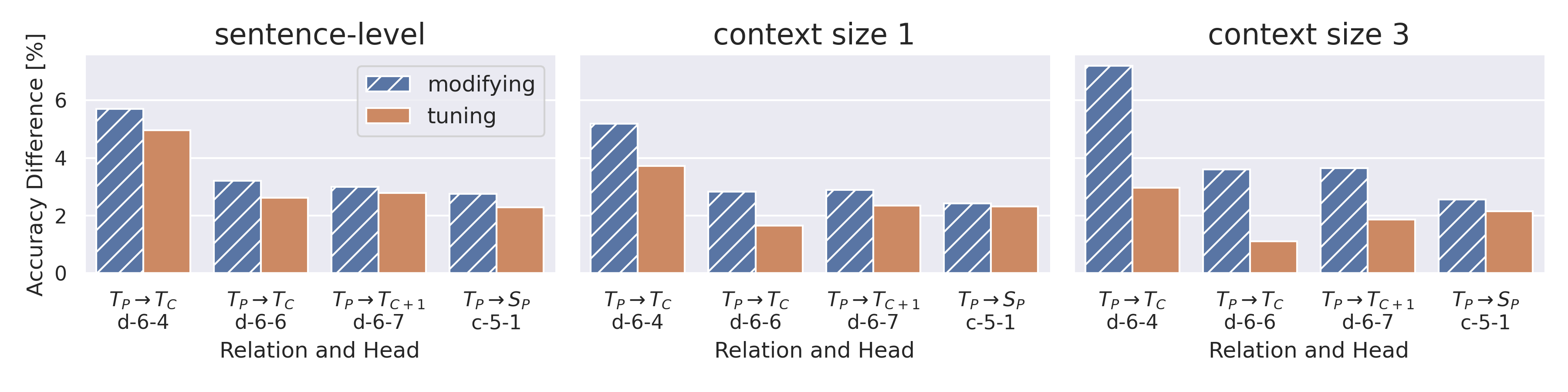}
    \caption{The difference in accuracy on ContraPro of the models based on OpusMT en-de (sentence-level, context-aware-1, and context-aware-3) for modifying and tuning selected heads.}
    \label{fig:head-tuning-results}
\end{figure*}

\change{For the English-to-French direction, the accuracy of the unmodified models on LCPT reaches $95\%$ for the sentence-level model\footnote{For the sentence-level model we only consider examples where the antecedent is in the current sentence.} and almost $92\%$ for the context-aware model, which is considerably higher than on ContraPro. This could explain the lower ranges of responses to modifying heads and correlation coefficients observed in the results.}

In contrast to the English-to-German direction, we observed a higher number encoder heads attending the $S_P \rightarrow S_C$ relation (e.g., \textit{e-12-4}, \textit{e-8-4}, \textit{e-6-13}, and \textit{e-12-8}). Nevertheless, none of them responded to modifying. Similarly, several heads (\textit{c-6-16}, \textit{c-8-3}, \textit{c-5-3}, \textit{c-7-11}) were attending but non-responsive to the $T_P \rightarrow S_P$ relation. For the $T_P \rightarrow S_C$ relation, head \textit{c-8-6} was non-attending and positively responsive. This behavior was exhibited by two decoder heads - \textit{d-10-3} and \textit{d-10-6} - for the $T_P \rightarrow T_C$ relation.
The $T_P \rightarrow T_{C+1}$ relation was attended by three heads - \textit{d-8-5} and \textit{d-9-12} heads also responding to modification, and \textit{d-9-9} head being non-responsive. Additionally, head \textit{d-10-3} was non-attending and positively responsive. Notice that this head is also responding to the $T_P \rightarrow T_C$ relation, albeit to a lesser extent.

Several heads show similar behavior in both language directions. Those include: \textit{e-12-4} in encoder-attention, \textit{c-8-3}, \textit{c-6-16}, \textit{c-5-3}, and \textit{c-8-9} in cross-attention, and \textit{d-9-12}, \textit{d-8-5}, and \textit{d-10-3} in decoder-attention. 

\subsection{Discussion}

Taking into account the results for all models we make the following observations.
\begin{itemize}[topsep=0pt,itemsep=0pt,partopsep=0pt, parsep=0pt, itemindent=15pt, leftmargin=0pt]
\item \textbf{Some heads appear to have a function identifiable through analysis.} This is confirmed by previous research \citep{clark-etal-2019-bert, voita-etal-2019-analyzing, tenney-etal-2019-bert, jo-myaeng-2020-roles, Olsson2022IncontextLA}, however we additionally demonstrate that adjustments to the functioning of heads 
(whether improved or diminished) leads to noticeable changes in models' performance (in terms of the accuracy of pronoun disambiguation).
\item \textbf{The decoder-attention (corresponding to the target-side context) has the highest impact on the pronoun disambiguation accuracy.} Note that we used the gold context (provided with the examples). In a real-world system, the context would come from the model's predictions. \change{It is interesting to note that the most relevant heads are located in higher layers. Our intuition is that in the decoder the output token is presumably decided in the layers closer to the output and only with this information can heads attempt to find the corresponding antecedent.}
\item \textbf{In the decoder-attention, the important context tokens are the tokens corresponding to both the antecedent being predicted (the $T_P \rightarrow T_C$ relation) and being passed to the model as input (the $T_P \rightarrow T_{C+1}$ relation)} with most heads specializing in attending one of the two and only some heads being able to utilize both.
\item \textbf{Attending to the relation by a head does not necessarily imply it has an impact on the pronoun disambiguation accuracy.} The most responsive heads already attend the relations of interest, but there exist heads that are not utilized by the model that could improve the performance if were attending the relations. 
\end{itemize}






Moreover, in the multi-lingual NLLB-200 model, we found several cross-lingual heads, exhibiting the same behavior for both tested language directions. 
Finally, we quantified the overlap of the improvement coming from modifying pairs of heads simultaneously to be lower than $30\%$ for all models
(see Appendix~\ref{sec:modifying-multiple-heads-expanded}).

\section{Tuning Heads}
\label{sec:head-tuning-results}

After identifying the most responsive heads, we fine-tuned them to assess to what extent the augmented behavior of the heads can be solidified into the models' parameters without compromising overall translation quality. To obtain the dataset containing the pronoun-antecedent pairs we applied the CTXPRO toolset \citep{wicks-post-2023-identifying} to the IWSLT 2017 en-de dataset (unrelated to the ContraPro dataset). The details can be found in Appendix~\ref{sec:head-tuning-details}.

We tuned selected heads of the OpusMT en-de models. The results are shown in Figure~\ref{fig:head-tuning-results}. The models exhibit higher accuracy for each tuned head without reducing the translation quality (see Appendix~\ref{sec:head-tuning-details} for the extended results). The improvement is the highest for the sentence-level model and reduces with increased context size. The low number of examples with antecedent distance of two or more in the training dataset could explain the reduction in improvement for the context-aware-3 model. Alternatively, the representations of the tokens that are passed to the tuned head are insufficient for the head to attend the relation. We leave further investigation of this phenomenon and the exploration of the most efficient methods to tune the heads for future research.

\section{Conclusions}
\label{sec:conclusions}


In this paper, we researched the influence of the attention heads in the Context-aware MT models on the task of pronoun disambiguation in English-to-German and English-to-French language directions. We measured and modified the attention scores corresponding to the relations that could influence the prediction of a pronoun: pronoun-antecedent on the source and target sides, pronoun-pronoun, and pronoun-antecedent between the target and source sides. 

We found that some heads do attend the relations of interest but not all of them influence the pronoun disambiguation capabilities of the models. We showed that some heads are \textit{underutilized} by the models - the models' performance could improve if they attended the relations. We confirmed that the target-side context is more impactful than the source-side context. Additionally, we note that the target-side context cue tokens can be useful to the model both when passed to the model (as the input of the decoder) as well as when they are predicted (as the output of the decoder). Lastly, we showed that the heads can be fine-tuned to attend the relations improving the models.


\section{Limitations}
\label{sec:limitations}

We only investigated the pre-trained models that we fine-tune for the Context-aware MT. The attention heads could behave differently if the models were trained from the random initialization. 
Additionally, we employed the contrastive datasets to analyze the models. The behavior of the models can differ from the generative setting. 
We also used the gold target context. In the real-world scenario, the models would base its predictions on the previously generated target context. 
Lastly, we only considered the pronoun disambiguation task and only two language directions: English-to-German and English-to-French.

\section{Acknowledgments}
The research presented in this paper was conducted as part of VOXReality project\footnote{\url{https://voxreality.eu/}}, which was funded by the European Union Horizon Europe program under grant agreement No 101070521.

\bibliography{anthology,custom}

\appendix

\section{Derivation of Modifying Heads Formula}
\label{sec:modifying-heads-derivation}

In this section, we will provide a short derivation of the Modifying Heads equation (eq.~\ref{eq:modified-heads}) (see Section~\ref{sec:modifying-heads}). Let us consider the $i$-th token in the left-hand side set $\mathcal{Y}$ of the relation of interest $\mathcal{Y} \rightarrow \mathcal{X}$ for the example $d$ from the dataset. For a selected layer $l$ and head $h$, the goal is to find the values of pre-softmax attention scores $\tilde{H}^{l,h,d}_{i,j}$ for all tokens $j$ from the right-hand side set $\mathcal{X}$ of the relation of interest such that the sum of the resulting attention scores (post-softmax) $\tilde{Z}^{l,h,d}_{i,j}$ would be equal to a desired value $C$. We also assume that the attention scores would be spread equally among the tokens from the right-hand side set $\mathcal{X}$. This can be formulated as:
\begin{equation} 
\label{eq:modified-sum-to-value}
\begin{aligned}
&C = \sum_{j' \in \mathcal{X}^d} \tilde{Z}^{l,h,d}_{i,j'} = |\mathcal{X}^d| \tilde{Z}^{l,h,d}_{i,j} \\
&\forall i \in \mathcal{Y}^d, j \in \mathcal{X}^d,
\end{aligned}
\end{equation}

The attention scores are calculated using the following equation:
\begin{equation} 
\label{eq:modified-softmax}
\begin{aligned}
&Z^{l,h,d}_{i,k} = \frac{\exp(H^{l,h,d}_{i,k})}{\sum_{k' \in X^d} \exp(H^{l,h,d}_{i,k'})} \\ 
&\forall i \in \mathcal{Y}^d, k \in X^d,
\end{aligned}
\end{equation}
where $X_d$ are the tokens from which the the right-hand side set $\mathcal{X}$ are selected. We expand eq.\ref{eq:modified-sum-to-value} using eq.\ref{eq:modified-softmax} and split the sum in the denominator into two parts - inside $\mathcal{X}^d$ and outside $k \in X^d \setminus \mathcal{X}^d$:
\begin{equation} 
\label{eq:modified-expanded}
\begin{aligned}
C = &|\mathcal{X}^d| \tilde{Z}^{l,h,d}_{i,j} \\
= &\frac{|\mathcal{X}^d| \exp(\tilde{H}^{l,h,d}_{i,j})}{|\mathcal{X}^d| \exp(\tilde{H}^{l,h,d}_{i,j}) + \sum_{k \in X^d \setminus \mathcal{X}^d} \exp(H^{l,h,d}_{i,k})} \\
&\forall i \in \mathcal{Y}^d, j \in \mathcal{X}^d.
\end{aligned}
\end{equation}
This equation can be reformulated as:
\begin{equation} 
\label{eq:modified-reformulated}
\begin{aligned}
&\exp(\tilde{H}^{l,h,d}_{i,j}) = \frac{C}{(1-C) |\mathcal{X}^d|} \sum_{k \in X^d \setminus \mathcal{X}^d} \exp(H^{l,h,d}_{i,k}) \\
&\forall i \in \mathcal{Y}^d, j \in \mathcal{X}^d,
\end{aligned}
\end{equation}
Finally, we apply the logarithm to both sides of eq.\ref{eq:modified-reformulated} and come to the equation for Modifying Heads from Section~\ref{sec:modifying-heads}:
\begin{equation} 
\begin{aligned}
&\tilde{H}^{l,h,d}_{i,j} = \log \bigl( \frac{C}{|\mathcal{X}^d| (1-C)} \sum_{k \in X^d \setminus \mathcal{X}^d} \exp ( H^{l,h,d}_{i,k} ) \bigr) \\ 
&\forall i \in \mathcal{Y}^d, j \in \mathcal{X}^d.
\end{aligned}
\end{equation}

\section{Disabling Heads Method}
\label{sec:disabling-heads}

In addition to the methods described in Section~\ref{sec:methods} we disabled heads of the models.
Disabling heads means that we assign equal attention score to all key $K$ tokens, thus they cannot function as before.
It can be done for all query $Q$ tokens or only for the selected ones. To minimize the disturbance to the model, we only disable heads for the Q tokens in the $\mathcal{Y}$ set. This allows the head to function normally for all other tokens but prevents it from attending to the contextually informative tokens from the contextually dependent tokens. Regardless, it still does not distinguish between relations of interest. For example, disabling a head in the cross-attention prevents it from attending both $T_P \rightarrow S_P$ and $T_P \rightarrow S_C$ relations reducing the granularity of the analysis. 

For an example $d$ from the dataset, disabling heads is defined as:
\begin{equation} \label{eq:disabled-heads}
\begin{aligned}
&\tilde{Z}^{l,h,d}_{i,k} = \frac{1}{|X^d|} \\ 
&\forall i \in \mathcal{Y}^d, k \in X^d,
\end{aligned}
\end{equation}
where $\tilde{Z}$ are the updated attention scores (after softmax is applied), $X^d$ is the whole set of attended tokens, including the attended tokens-of-interest $\mathcal{X}^d$.
Implementation-wise, we set pre-softmax attention scores $H$ to zero and rely on the softmax function to spread attention equally among the tokens.

\section{Details of the Datasets}
\label{sec:datasets-details}

In this section, we present the details of the datasets used in our investigation. The number of documents, sentences, and tokens in the subsets of the IWSLT 2017 \citep{cettolo-etal-2017-overview} dataset for English-to-German and English-to-French directions can be seen in Table~\ref{tab:iwslt-dataset-stats}.

\begin{table}[!ht]
    \centering
    \begin{tabular}{lrrrrrr}
    \hline
        \textbf{Dataset} & \textbf{Docs} & \textbf{Sent/Doc} & \textbf{Tok/Sent} \\ 
        \hline
        En-De Train & 1698 & 121.4 & 21.9 \\ 
        En-De Valid & 62 & 87.6 & 20.6 \\
        En-De Test & 12 & 90.0 & 20.8 \\ 
        \hline
        En-Fr Train  & 1914 & 121.6 & 22.0 \\ 
        En-Fr Valid  & 66 & 88.2 & 20.9 \\ 
        En-Fr Test  & 12 & 100.8 & 21.4  \\ 
    \hline
    \end{tabular}
    \caption{The details of the IWSLT 2017 datasets.}
    \label{tab:iwslt-dataset-stats}
\end{table}

The number of examples in the ContraPro and LCPT contrastive datasets for the different distances of the antecedent are presented in Table~\ref{tab:contrastive-dataset-stats}.

\begin{table}[!ht]
    \centering
    \begin{tabular}{lrr}
    \hline
        \textbf{Antecedent} & \multicolumn{2}{c}{\textbf{}}  \\ 
        \textbf{Distance} & \textbf{ContraPro} & \textbf{LCPT} \\ 
        \hline
        0 & 2400 & 5986 \\ 
        1 & 7075 & 4566 \\
        2 & 1510 & 1629 \\ 
        3  & 573 & 880 \\ 
        >3  & 442 & 939 \\ 
        \hline
        All & 12000 & 14000 \\
    \hline
    \end{tabular}
    \caption{The number of examples in the ContraPro and LCPT datasets with different distances of the antecedent.}
    \label{tab:contrastive-dataset-stats}
\end{table}

\section{Details of Context-aware Fine-tuning}
\label{sec:fine-tuning-hyperparameters}
We trained the models with Adafactor optimizer \citep{Shazeer2018AdafactorAL} on a single GPU (NVIDIA GeForce RTX 3090 24GB) for $10$ epochs and select the checkpoint with the highest BLEU \citep{papineni-etal-2002-bleu} on the ContraPro dataset \citep{muller-etal-2018-large} (translating the sentences with the provided target context). The models acquired the highest BLEU score after two epochs for OpusMT en-de with a context window of one, and only after one epoch for OpusMT en-de with a context window of three and NLLB-200 (with a context window of one).

The hyper-parameters are presented in Table~\ref{tab:hyper-parameters}. For OpisMT en-de with the context size of three, we decreased the batch size to $16$ and increased the gradient accumulation steps to $16$.
For NLLB-200, we decreased the batch size further to $8$ and increased the gradient accumulation steps to $32$. This was done to reduce the memory requirements during training. We trained for $10$ epochs and selected the checkpoint with the highest BLEU score on the ContraPro dataset. We tuned learning rate on OpusMT en-de model for context size of one and tried the following values: 2e-6, 5e-6, 1e-5, 2e-5, 5e-5. We chose the value of 1e-5 for learning rate as leading to the highest BLEU on the ContraPro dataset (when generating translations). We used this value for the fine-tuning of all models.

\begin{table}[!ht]
    \centering
    \begin{tabular}{lr}
    \hline
        \textbf{Hyper-parameter} & \textbf{Value}    \\ \hline
        Optimizer & Adafactor \\
        Learning Rate & 1e-5 \\
        LR Scheduler & Linear \\
        LR Warmup Ratio & 0 \\
        Weight Decay & 0.01 \\
        Max Gradient Norm & 1.0 \\
        Batch Size & 32\textsuperscript{*} \\
        Gradient Accumulation Steps & 8\textsuperscript{*} \\
        Num Epoch & 10 \\
        \hline
    \end{tabular}
    \caption{The hyper-parameters of the Context-aware fine-tuning of the tested models.\\
    \textsuperscript{*} For Opus MT en-de with the context size of three and NLLB-200, we decreased the batch size to $16$ and $8$, and increased the gradient accumulation steps to $16$ and $32$ respectively.}
    \label{tab:hyper-parameters}
\end{table}

\section{Modifying Multiple Heads}
\label{sec:modifying-multiple-heads-expanded}

For all investigated models, we found that several heads are improving the accuracy of the model when modified to attend the relations of interest. To asses to what extent the improvement of different heads is overlapping we selected several heads with highest improvement in accuracy when modified to $0.99$ and modified each pair of heads together to $0.99$ for the relations they responded. We experimented with two models: OpusMT en-de with context size of three (we selected 8 heads), and context-aware NLLB-200 model on English-to-German and English-to-French directions (we selected 7 and 5 heads respectively). We report the percentage overlap of the accuracy improvement of modifying both heads against the sum of improvements of modifying each head separately, calculated as:
\begin{equation} \label{eq:percentage-overlap}
\begin{aligned}
&O_\mathcal{M}=\frac{\sum_{(l,h) \in \mathcal{M}} (A_{l,h} - A_B) - (A_\mathcal{M} - A_B)}{\sum_{(l,h) \in \mathcal{M}} (A_{l,h} - A_B)},
\end{aligned}
\end{equation}
where $\mathcal{M}$ is the list of heads being simultaneously modified, $A_B$ is the base accuracy of the model, $A_\mathcal{M}$ is the accuracy observed when all heads are modified to $0.99$, and $A_{l,h}$ is the accuracy of the model with a single $h$-th head in $l$-th layer being modified to $0.99$.

\begin{figure}[!ht]
\centering
    \includegraphics[width=1\linewidth]{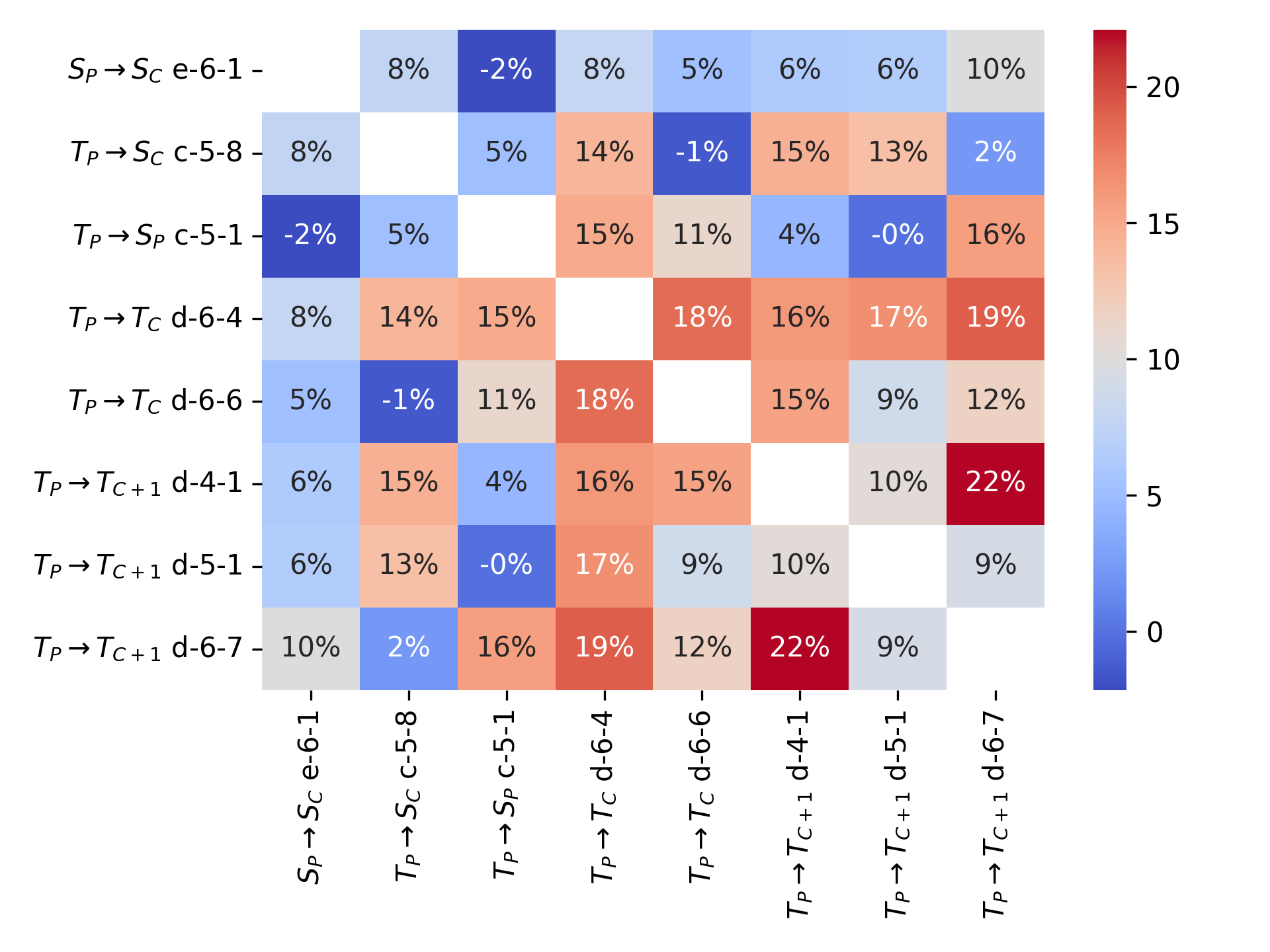}
    \caption{The percentage overlap of the improvement in accuracy on the ContraPro dataset for selected heads of the context-aware-1 model based on \textbf{OpusMT en-de}.}
    \label{fig:opsu-mt-multiple-heads-overlap}
\end{figure}

\begin{figure}[!ht]
\centering
    \includegraphics[width=1\linewidth]{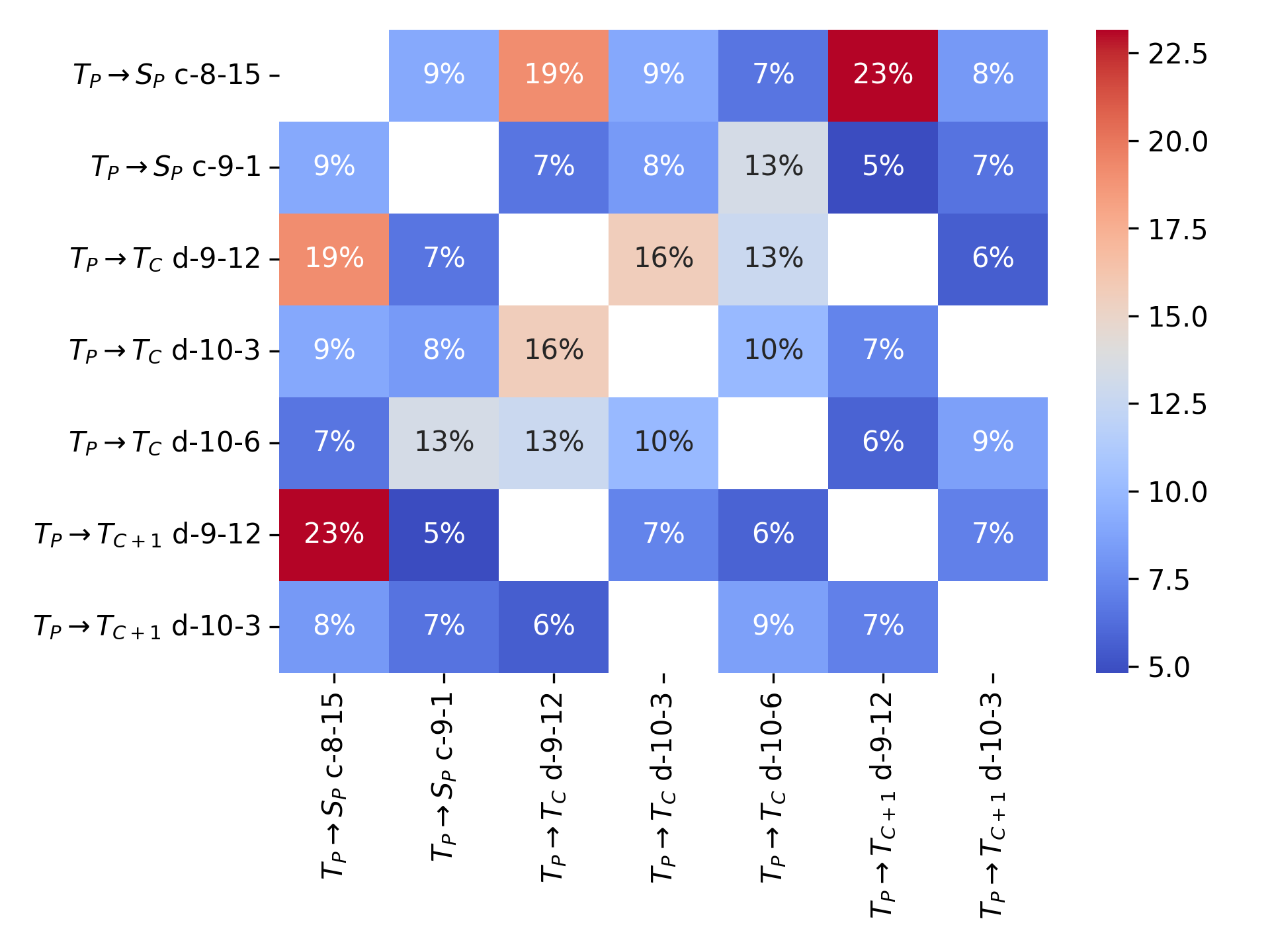}
    \caption{The percentage overlap of the improvement in accuracy on the \textbf{ContraPro} dataset (English-to-German) for selected heads of the context-aware model based on \textbf{NLLB-200}.}
    \label{fig:nllb-ende-multiple-heads-overlap}
\end{figure}

\begin{figure}[!ht]
\centering
    \includegraphics[width=1\linewidth]{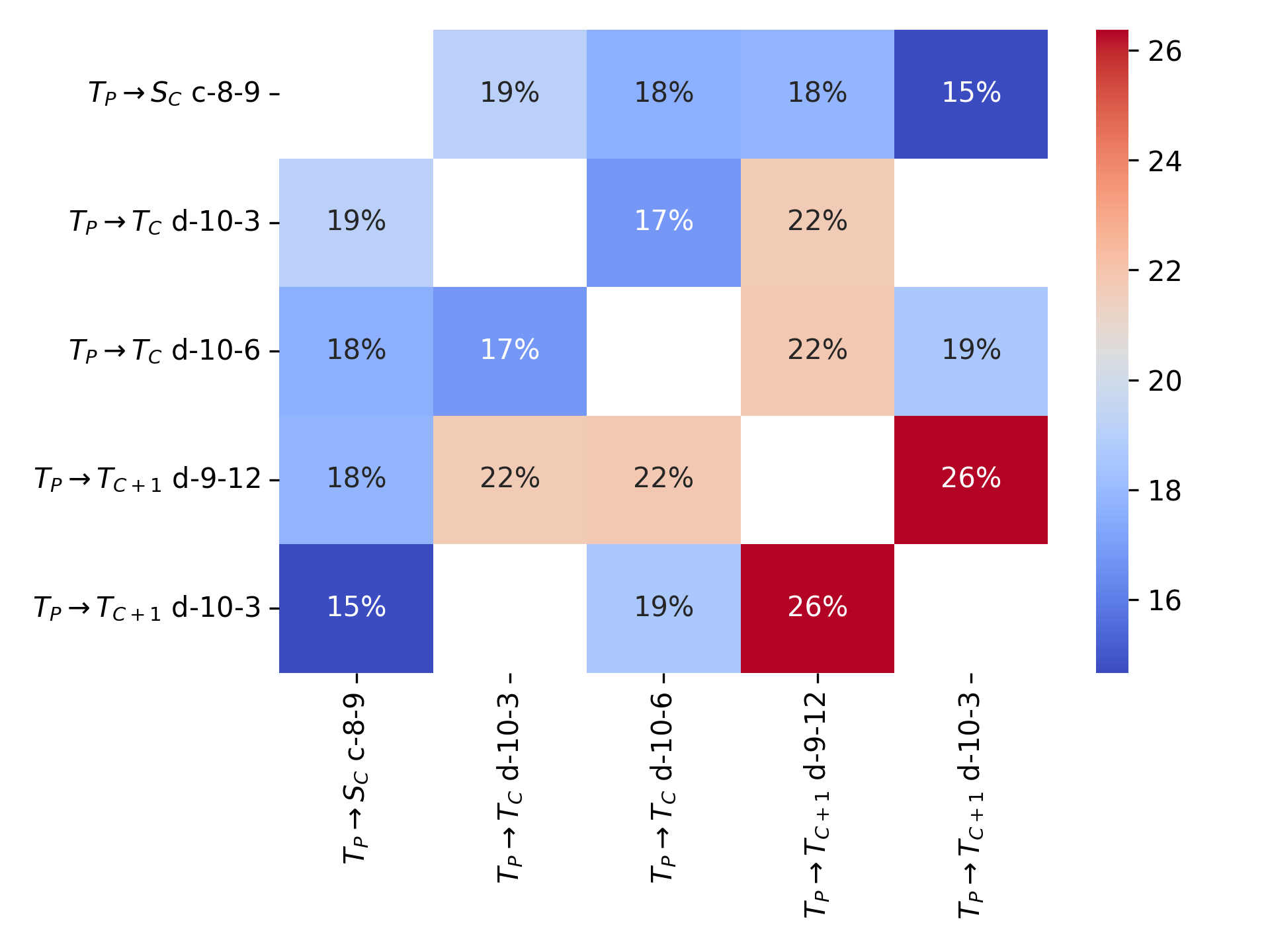}
    \caption{The percentage overlap of the improvement in accuracy on the \textbf{LCPT} dataset (English-to-French) for selected heads of the context-aware model based on \textbf{NLLB-200}.}
    \label{fig:nllb-enfr-multiple-heads-overlap}
\end{figure}

We present the results for the OpusMT en-de model in Figure~\ref{fig:opsu-mt-multiple-heads-overlap}. The corresponding results for NLLB-200 model can be found in in Figures~\ref{fig:nllb-ende-multiple-heads-overlap} and \ref{fig:nllb-enfr-multiple-heads-overlap} for English-to-German and English-to-French directions respectively. We do not report the result for the cases where the same head was selected based on two different relations (e.g., head \textit{d-9-12} for the $T_P \rightarrow T_{C}$ and $T_P \rightarrow T_{C+1}$ relations).
It can be seen that some overlap is present for most of the heads but it is typically lower than $20\%$ for English-to-German direction in both models and slightly higher (from $15\%$ to $26\%$) for English-to-French direction in NLLB-200. This means that heads perform complementary (to an extent) functions in the models, and the redundancy of the function they perform is limited. Furthermore, improving heads in tandem could lead to the model with a better performance.



\section{Details of Head Tuning}
\label{sec:head-tuning-details}

We trained the selected heads of a model to generate the attention scores matching the scores resulting from modifying to $0.99$. We froze all parameters of the model apart from the parameters responsible for the Q and K transformation in a targeted layer and updated only the parameters corresponding to the targeted head. Although we experimented with the calculation of the loss based on post-softmax attention scores $Z$, we observed a slight decrease in BLEU on the IWSLT 2017 testset. Therefore, we decided to use the pre-softmax attention scores $H$ to calculate the loss. We formulate the target pre-softmax attention scores for a token $i \in \mathcal{Y}$ as:
\begin{equation} 
\label{eq:heads-tuning-target}
\begin{aligned}
&\hat{H}^{l,h,d}_{i,j} = \begin{cases}
    \tilde{H}^{l,h,d}_{i,j}, \text{if } j \in \mathcal{X},\\
    \bar{H}^{l,h,d}_{i,j}, \text{if } j \notin \mathcal{X},\\
\end{cases}
\end{aligned}
\end{equation}
where $\tilde{H}$ is calculated according to eq.~\ref{eq:modified-heads} (see Section~\ref{sec:modifying-heads}, and $\bar{H}$ is the attention score obtained from the unmodified model. In order to obtain the values of $\bar{H}$, at the start of training, we make a copy of the model parameters and perform a forward pass twice, first for the trained model and second for the unmodified model. We calculate the MSE loss between the model's and target pre-softmax scores and backpropagate it to the model's parameters.

We obtain training examples by applying the CTXPRO \citep{wicks-post-2023-identifying} toolset to the training subset of the IWSLT 2017 en-de dataset. The toolset applies predefined rules to find context-related phenomena in the dataset. We extracted the examples marked by the following rules: 
\begin{itemize}[topsep=0pt,itemsep=0pt,partopsep=0pt, parsep=0pt, itemindent=15pt, leftmargin=0pt]
    \item \verb|ACC.NEUT.SING|, 
    \item \verb|DAT.FEM.SING|, 
    \item \verb|NOM.FEM.SING|, 
    \item \verb|NOM.NEUT.SING|, 
    \item \verb|DAT.NEUT.SING|, 
    \item \verb|DAT.MASC.SING|, 
    \item \verb|NOM.MASC.SING|, 
    \item \verb|ACC.FEM.SING|, 
    \item \verb|ACC.MASC.SING|.
\end{itemize}
The resulting number of examples with different antecedent distances are presented in Table~\ref{tab:ctxpro-dataset-stats}. The hyper-parameters used in fine-tuning can be found in Table~\ref{tab:head-tuning-hyper-parameters}.

\begin{table}[!ht]
    \centering
    \begin{tabular}{lrr}
    \hline
        \textbf{Antecedent} & \multicolumn{1}{c}{\textbf{}}  \\ 
        \textbf{Distance} & \textbf{CTXPRO} \\ 
        \hline
        0 & 3816  \\ 
        1 & 2883  \\
        2 & 845  \\ 
        3  & 355  \\ 
        >3  & 448 \\ 
        \hline
        All & 8347 \\
    \hline
    \end{tabular}
    \caption{The number of examples in the CTXPRO dataset.}
    \label{tab:ctxpro-dataset-stats}
\end{table}


\begin{table*}[!ht]
\noindent
\centering
\begin{tabular}{
  L{\dimexpr.19\linewidth-2\tabcolsep-1.3333\arrayrulewidth}
  L{\dimexpr.15\linewidth-2\tabcolsep-1.3333\arrayrulewidth}
  L{\dimexpr.1\linewidth-2\tabcolsep-1.3333\arrayrulewidth}
  G{\dimexpr.138\linewidth-2\tabcolsep-1.3333\arrayrulewidth}
  G{\dimexpr.138\linewidth-2\tabcolsep-1.3333\arrayrulewidth}
  G{\dimexpr.138\linewidth-2\tabcolsep-1.3333\arrayrulewidth}
  G{\dimexpr.138\linewidth-2\tabcolsep-1.3333\arrayrulewidth}
  }
\hline
        \textbf{Model} & \textbf{Relation} & \textbf{Tuned Head} & \textbf{ContraPro Accuracy} & \textbf{Modified Accuracy} & \textbf{ContraPro BLEU}  & \textbf{IWSLT BLEU} \\
\hline
        
        Sentence-level &  - &   - & $81.46\%$ & - & $30.24$ & $32.42$  \\
         &  $T_P \rightarrow T_{C}$ &   \textit{d-6-4} & $86.42\%$ & $87.17\%$ & $30.67$ & $32.39$  \\ 
         &  $T_P \rightarrow T_{C}$ &   \textit{d-6-6} & $84.08\%$ & $84.67\%$ & $30.40$ & $32.37$ \\
         &  $T_P \rightarrow T_{C+1}$ &   \textit{d-6-7} & $84.25\%$ & $84.46\%$ & $30.40$ & $32.38$ \\
         &  $T_P \rightarrow S_{P}$ &   \textit{c-5-1} & $83.75\%$ & $84.21\%$ & $30.52$ & $32.31$ \\
\hline
         Context-aware-1 &  - &  - & $78.35\%$ & - & $31.00$ & $34.40$  \\
         &  $T_P \rightarrow T_{C}$ &   \textit{d-6-4} & $82.08\%$ & $83.53\%$ & $31.17$ & $34.50$  \\

         &  $T_P \rightarrow T_{C}$ &   \textit{d-6-6} & $80.01\%$ & $81.18\%$ & $31.06$ & $34.55$ \\
         &  $T_P \rightarrow T_{C+1}$ &   \textit{d-6-7} & $80.69\%$ & $81.24\%$ & $31.02$ & $34.47$ \\
         &  $T_P \rightarrow S_{P}$ &   \textit{c-5-1} & $80.68\%$ & $80.78\%$ & $31.15$ & $34.47$ \\
\hline
         Context-aware-3 &  - & - & $79.08\%$ & - & $31.35$ & $34.59$  \\
         &  $T_P \rightarrow T_{C}$ &   \textit{d-6-4} & $82.04\%$ & $86.28\%$ & $31.62$ & $34.58$  \\

         &  $T_P \rightarrow T_{C}$ &   \textit{d-6-6} & $80.18\%$ & $82.68\%$ & $31.41$ & $34.56$ \\
         &  $T_P \rightarrow T_{C+1}$ &   \textit{d-6-7} & $80.94\%$ & $82.72\%$ & $31.40$ & $34.58$ \\
         &  $T_P \rightarrow S_{P}$ &   \textit{c-5-1} & $81.23\%$ & $81.64\%$ & $31.52$ & $34.59$ \\
         
\hline
    \end{tabular}
    \caption{The results of fine-tuning selected heads to attend the specified relation in terms of the accuracy and BLEU on ContraPro (the \textit{Tuned Accuracy} and \textit{ContraPro BLEU} columns respectively), BLEU on IWSLT 2017 en-de dataset (the \textit{IWSLT BLEU} column) compared to the accuracy of the model with modifying the head to $0.99$ (the \textit{Modified Accuracy} column). The first row for each model shows the results of the unmodified and not fine-tuned model. The Accuracy and BLEU on the ContraPro dataset for the sentence-level models are calculated only on the examples where the antecedent is in the current sentence.}
\end{table*}

\begin{table}[!ht]
    \centering
    \begin{tabular}{lr}
    \hline
        \textbf{Hyper-parameter} & \textbf{Value}    \\ \hline
        Optimizer & Adafactor \\
        Learning Rate & 1e-3 \\
        LR Scheduler & Linear \\
        LR Warmup Ratio & 0 \\
        Weight Decay & 0.01 \\
        Max Gradient Norm & 1.0 \\
        Batch Size & 12 \\
        Gradient Accumulation Steps & 8 \\
        Num Epoch & 10 \\
        \hline
    \end{tabular}
    \caption{The hyper-parameters of the fine-tuning of selected heads.}
    \label{tab:head-tuning-hyper-parameters}
\end{table}

\section{Expanded Results of the OpusMT en-de Models}
\label{sec:expanded-results-opus-ende}

\begin{figure*}[!ht]
\center{}
    \begin{subfigure}{0.49\linewidth}
        \includegraphics[width=1\linewidth]{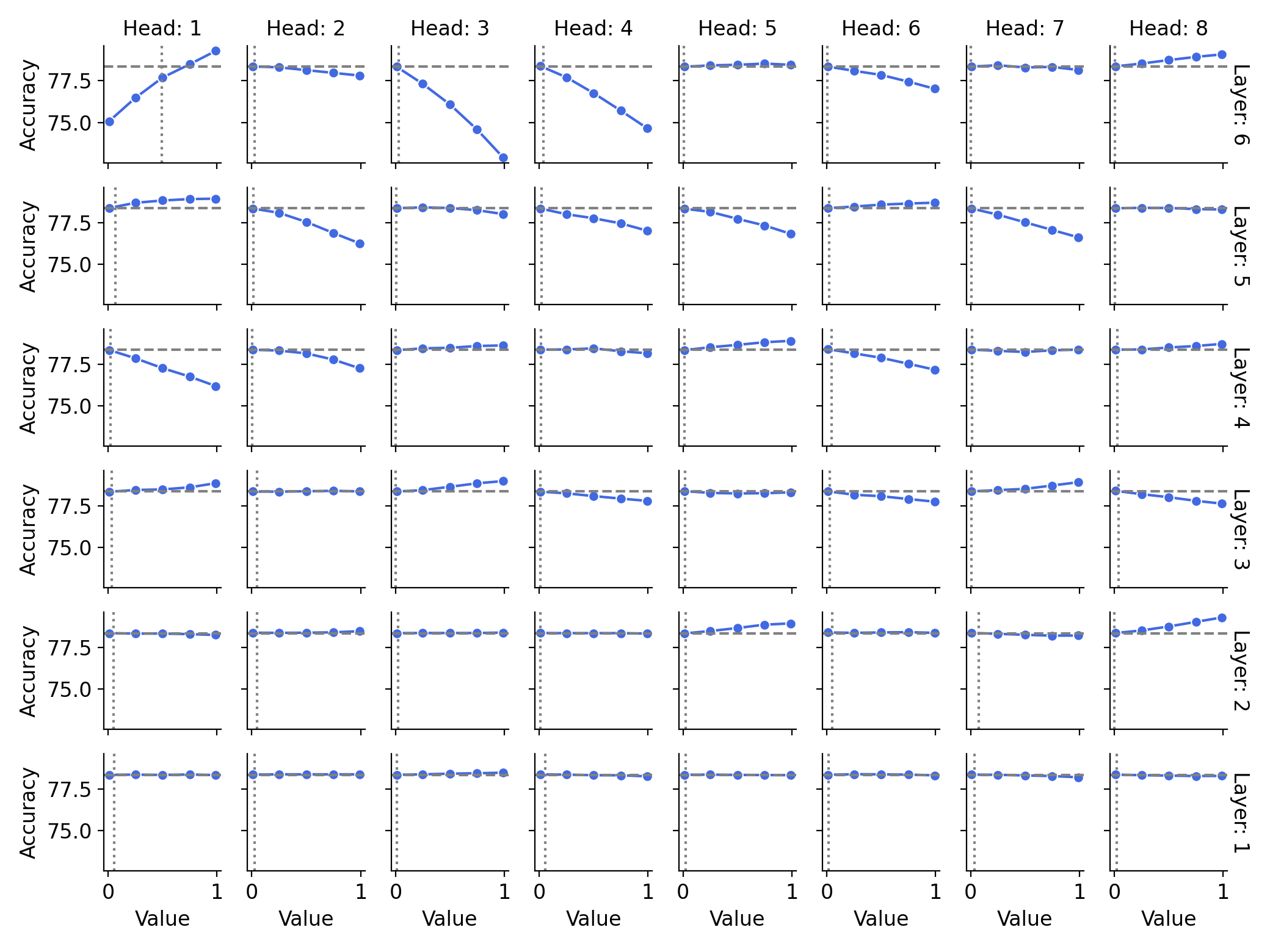}
        \caption{$S_P \rightarrow S_C$}
        \label{fig:opus-ende-ctx-1-modified-enc}
    \end{subfigure}
    
    %
    %
    \begin{subfigure}{0.49\linewidth}
        \includegraphics[width=1\linewidth]{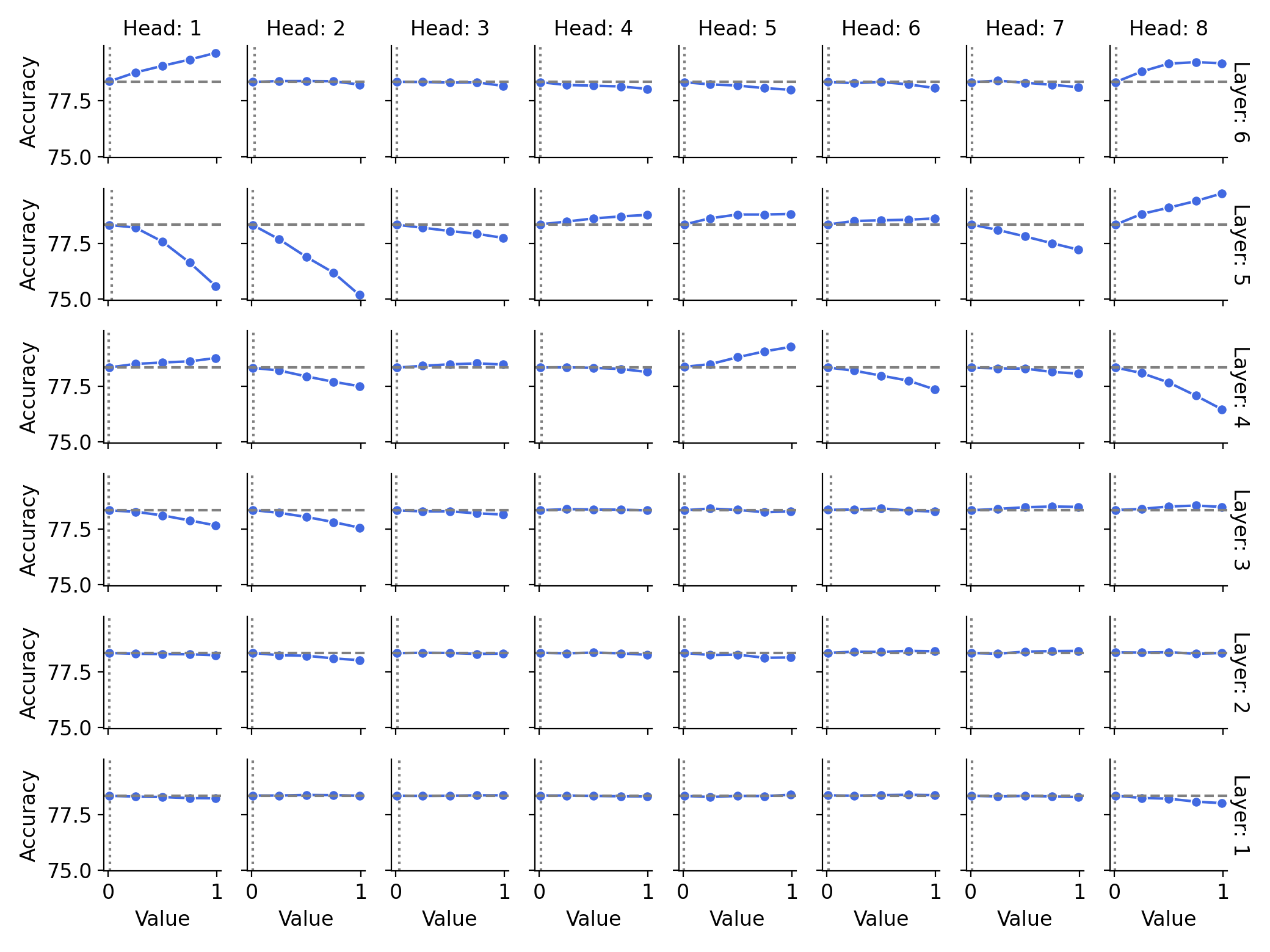}
        \caption{$T_P \rightarrow S_C$}
        \label{fig:opus-ende-ctx-1-modified-context-cross}
    \end{subfigure}
    %
    \begin{subfigure}{0.49\linewidth}
        \includegraphics[width=1\linewidth]{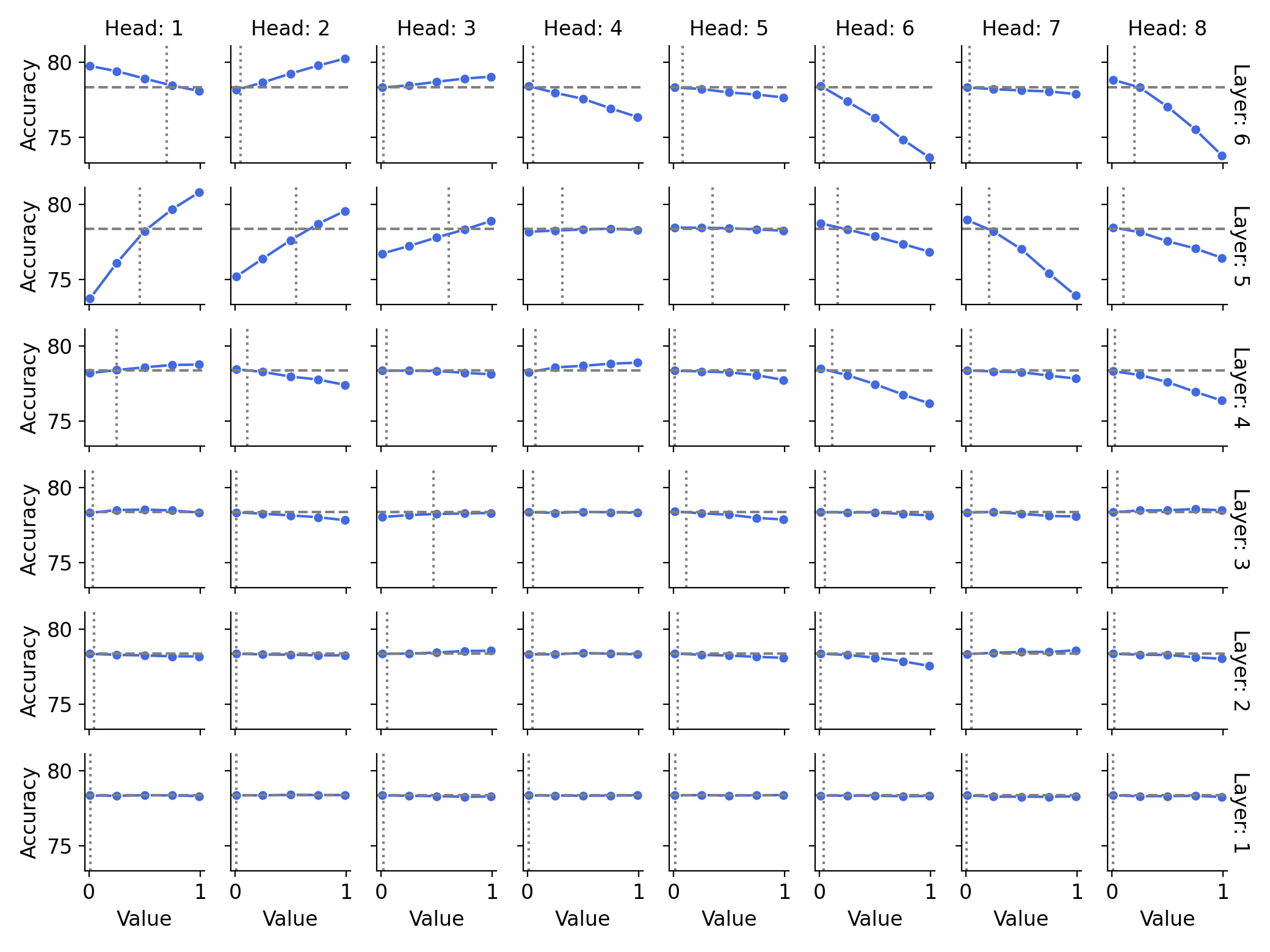}
        \caption{$T_P \rightarrow S_P$}
        \label{fig:opus-ende-ctx-1-modified-phrase-cross}
    \end{subfigure}
    \begin{subfigure}{0.49\linewidth}
        \includegraphics[width=1\linewidth]{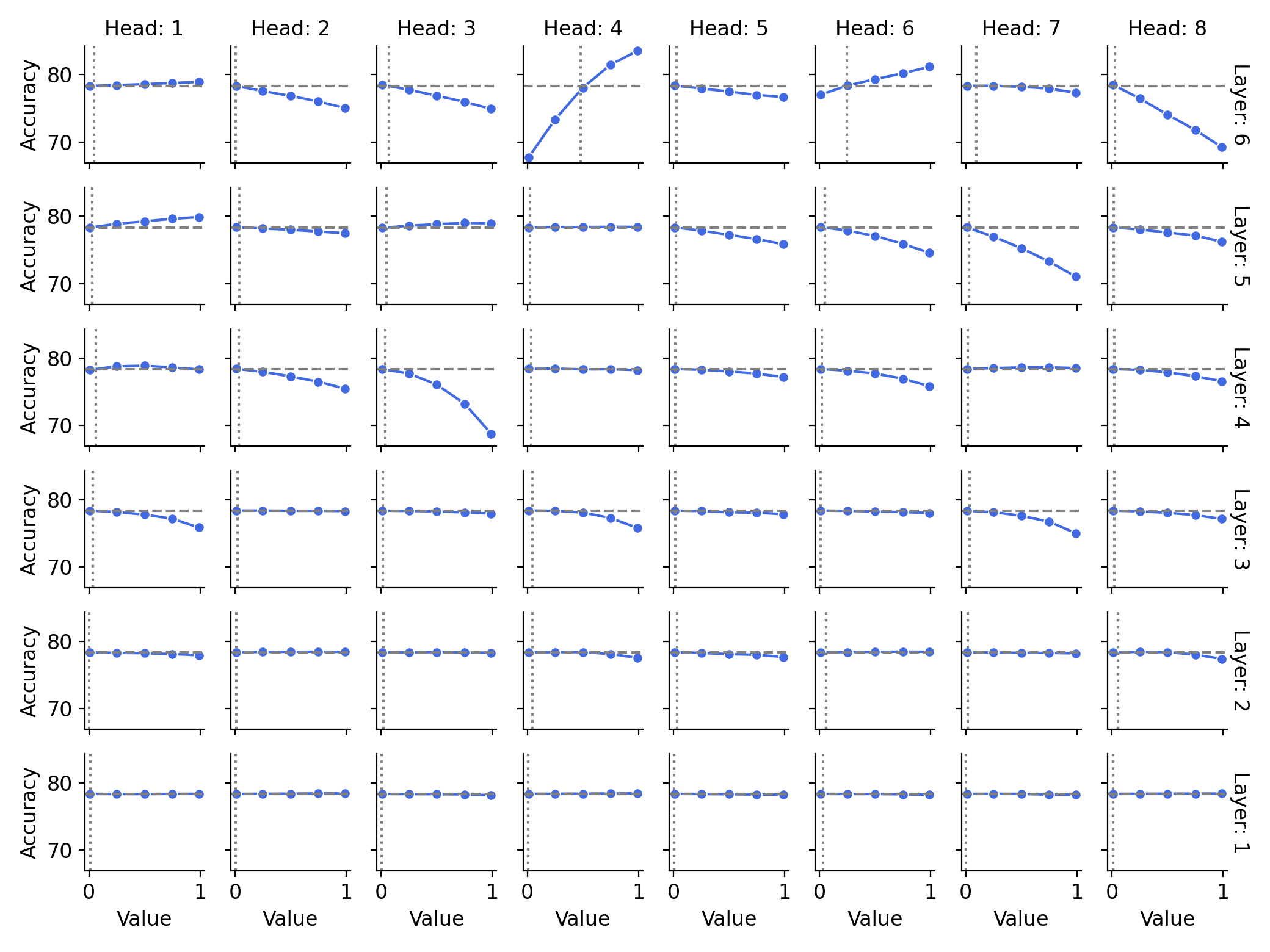}
        \caption{$T_P \rightarrow T_C$}
        \label{fig:opus-ende-ctx-1-modified-decoder}
    \end{subfigure}
    \begin{subfigure}{0.49\linewidth}
        \includegraphics[width=1\linewidth]{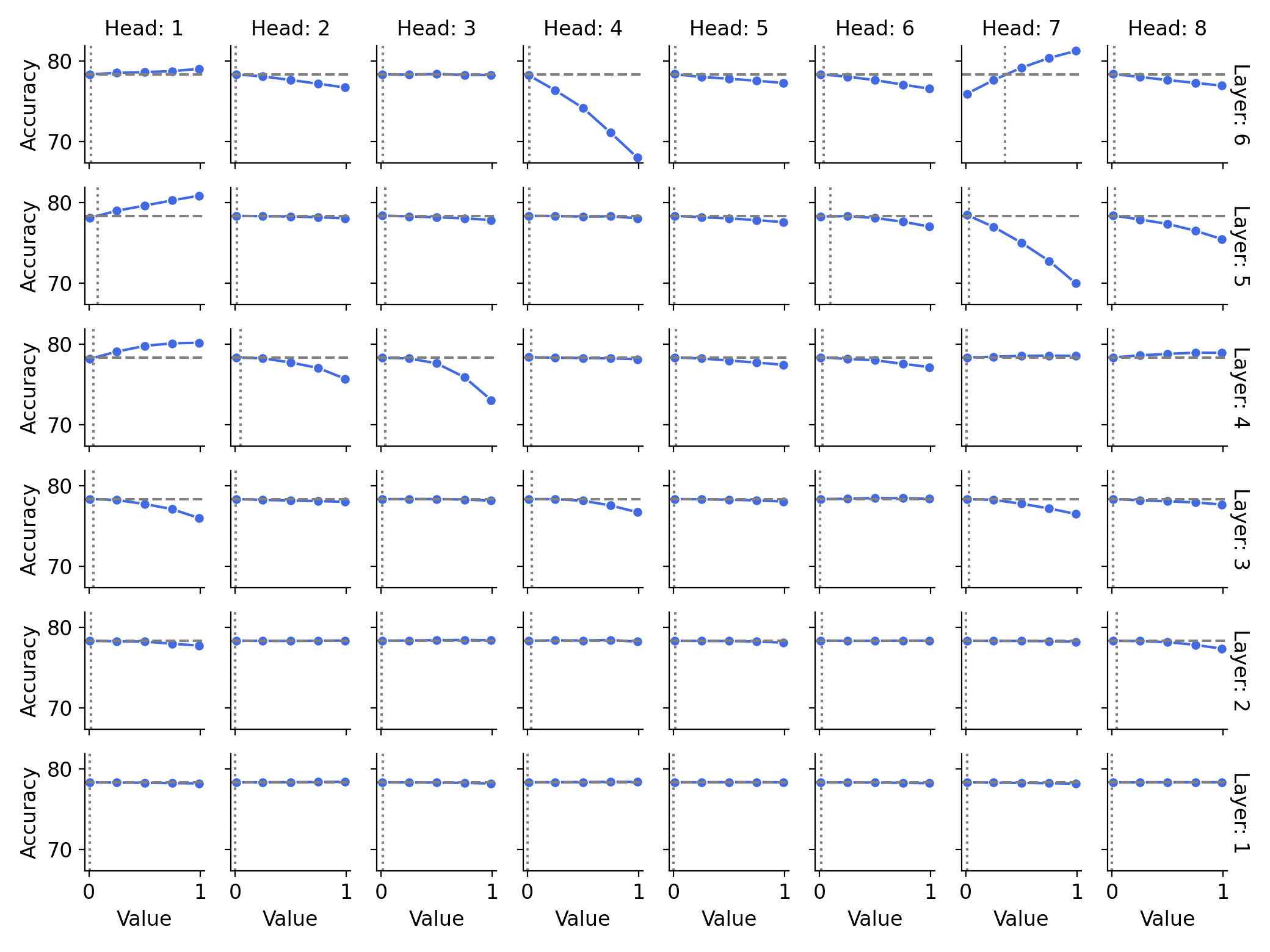}
        \caption{$T_P \rightarrow T_{C+1}$}
        \label{fig:opus-ende-ctx-1-modified-decoder-after}
    \end{subfigure}
    
    \caption{The accuracy of the \textbf{OpusMT en-de context-aware-1} model on the ContraPro dataset when modifying each of the attention heads to different values (out of $[0.01, 0.25, 0.5, 0.75, 0.99]$) for the investigated relations of interest.}
    \label{fig:opus-ende-ctx-1-modified}
\end{figure*}

In this section, we present the expanded results for the models based on OpusMT en-de (sentence-level, context-aware-1, and context-aware-3). The accuracy of the context-aware model with the context size of one on the ContraPro contrastive dataset when modifying heads to different values (from the set $[0.01, 0.25, 0.5, 0.75, 0.99]$) for five relations of interest can be seen in Figure~\ref{fig:opus-ende-ctx-1-modified}. The results show that the changes in accuracy are monotonic. 

\begin{figure*}[t]
  \includegraphics[width=\linewidth]{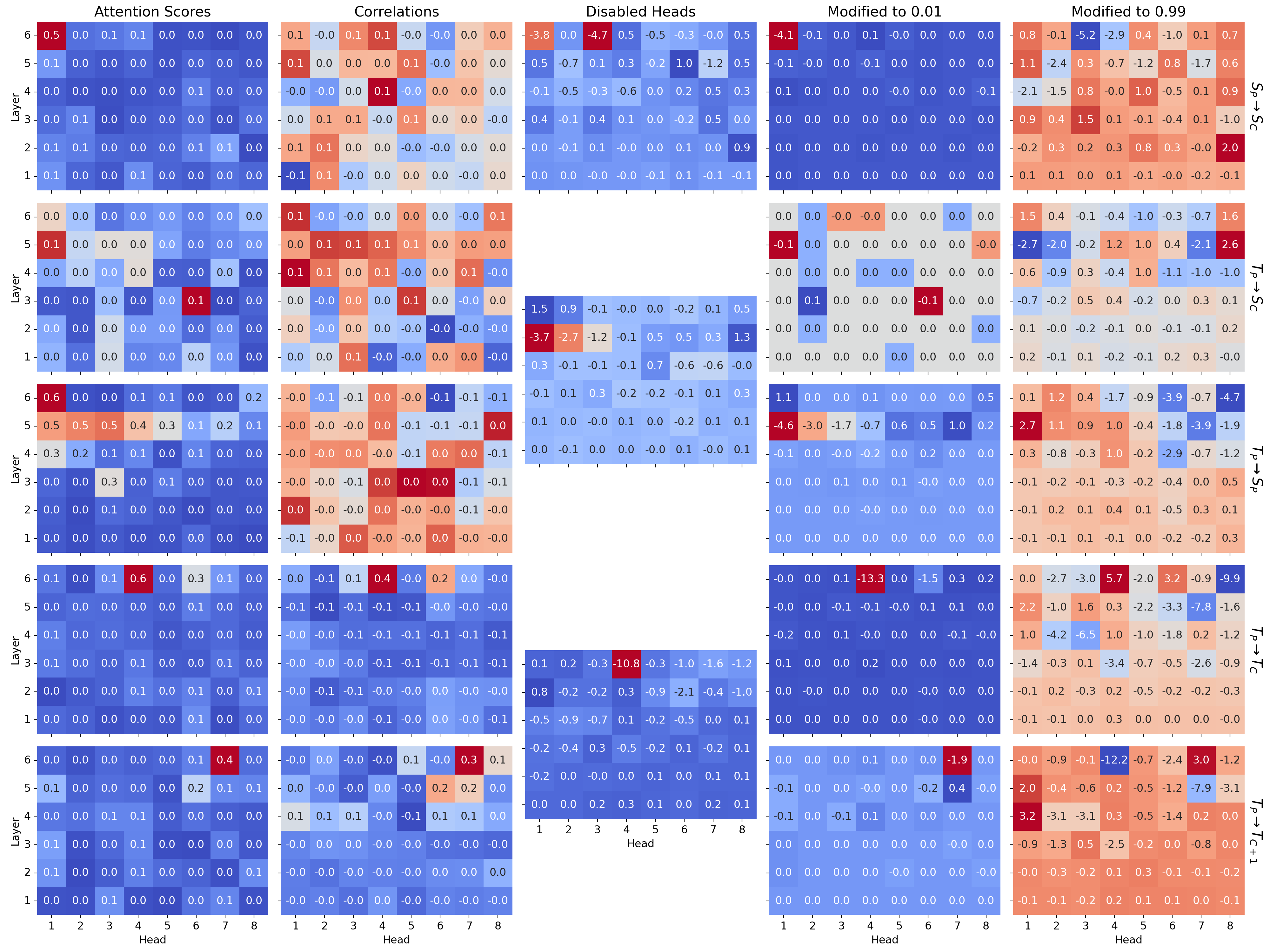}
  \caption{Measured average attention score, correlation between attention scores and accuracy on ContraPro dataset, and the difference in accuracy for: Disabling Heads, Modifying Heads to $0.01$, and Modifying Heads to $0.99$ (as columns) for each head of the \textbf{sentence-level OpusMT en-de} model for all relations of interest (as rows).}
  \label{fig:opus-ende-sentence-combined}
\end{figure*}

\begin{figure*}[t]
  \includegraphics[width=\linewidth]{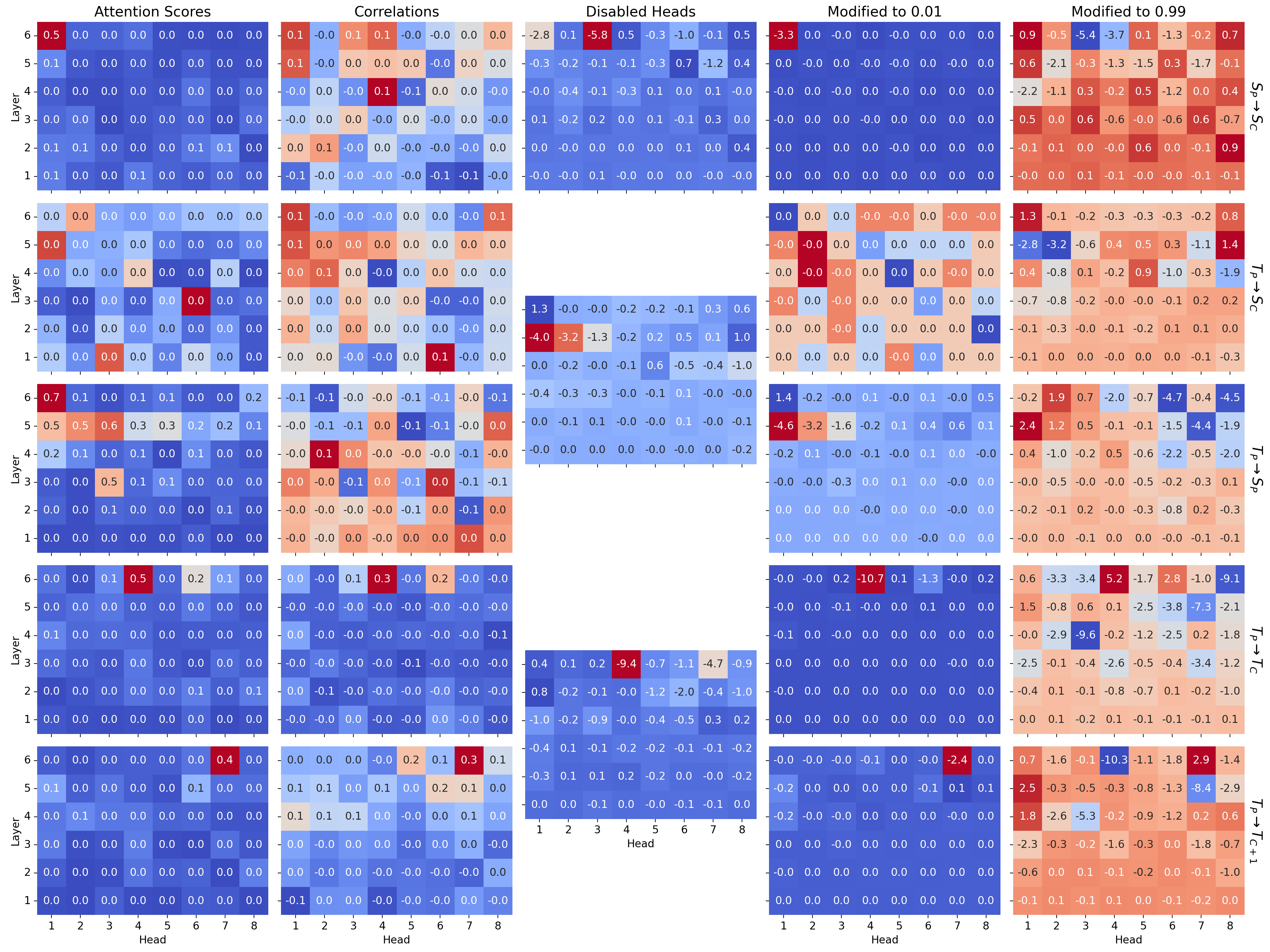}
  \caption{Measured average attention score, correlation between attention scores and accuracy on ContraPro dataset, and the difference in accuracy for: Disabling Heads, Modifying Heads to $0.01$, and Modifying Heads to $0.99$ (as columns) for each head of the \textbf{context-aware OpusMT en-de} model with the \textbf{context size of one} for all relations of interest (as rows).}
  \label{fig:opus-ende-ctx-1-combined}
\end{figure*}

\begin{figure*}[t]
  \includegraphics[width=\linewidth]{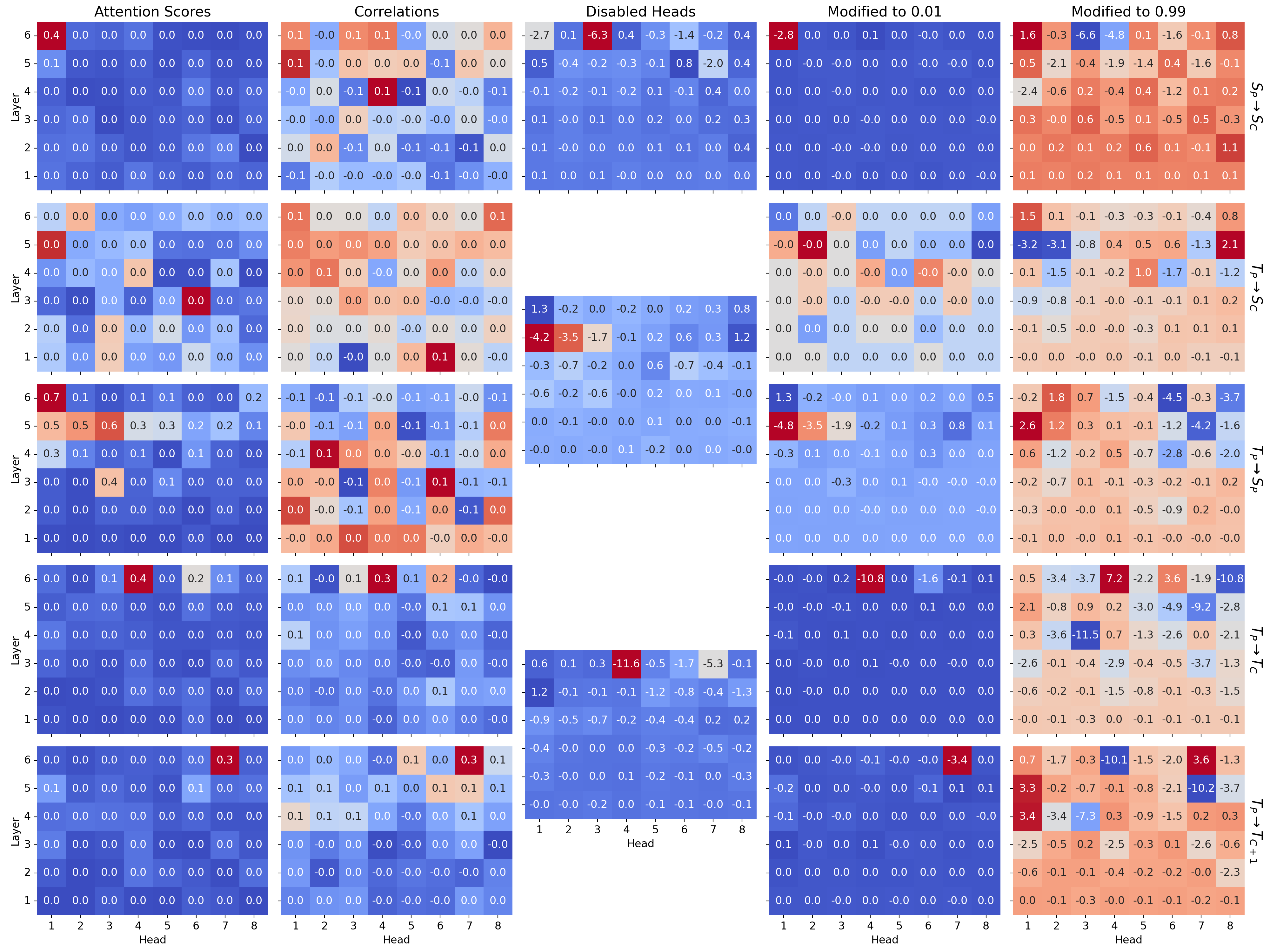}
  \caption{Measured average attention score, correlation between attention scores and accuracy on ContraPro dataset, and the difference in accuracy for: Disabling Heads, Modifying Heads to $0.01$, and Modifying Heads to $0.99$ (as columns) for each head of the \textbf{context-aware OpusMT en-de} model with the \textbf{context size of three} for all relations of interest (as rows).}
  \label{fig:opus-ende-ctx-3-combined}
\end{figure*}

The raw results for all measured quantities (average attention scores, correlation between scores and being correct on the contrastive dataset examples, the difference in accuracy when disabling heads, and modifying head to $0.01$ and $0.99$) for all relations of interest are presented in Figures~\ref{fig:opus-ende-sentence-combined}, \ref{fig:opus-ende-ctx-1-combined}, and \ref{fig:opus-ende-ctx-3-combined} for the three models (sentence-level, context-aware-1, and context-aware-3) respectively.

\section{Expanded Results of the NLLB-200 Models}
\label{sec:expanded-results-nllb-600M}

\begin{figure*}[!ht]
\centering
    \includegraphics[width=0.9\linewidth]{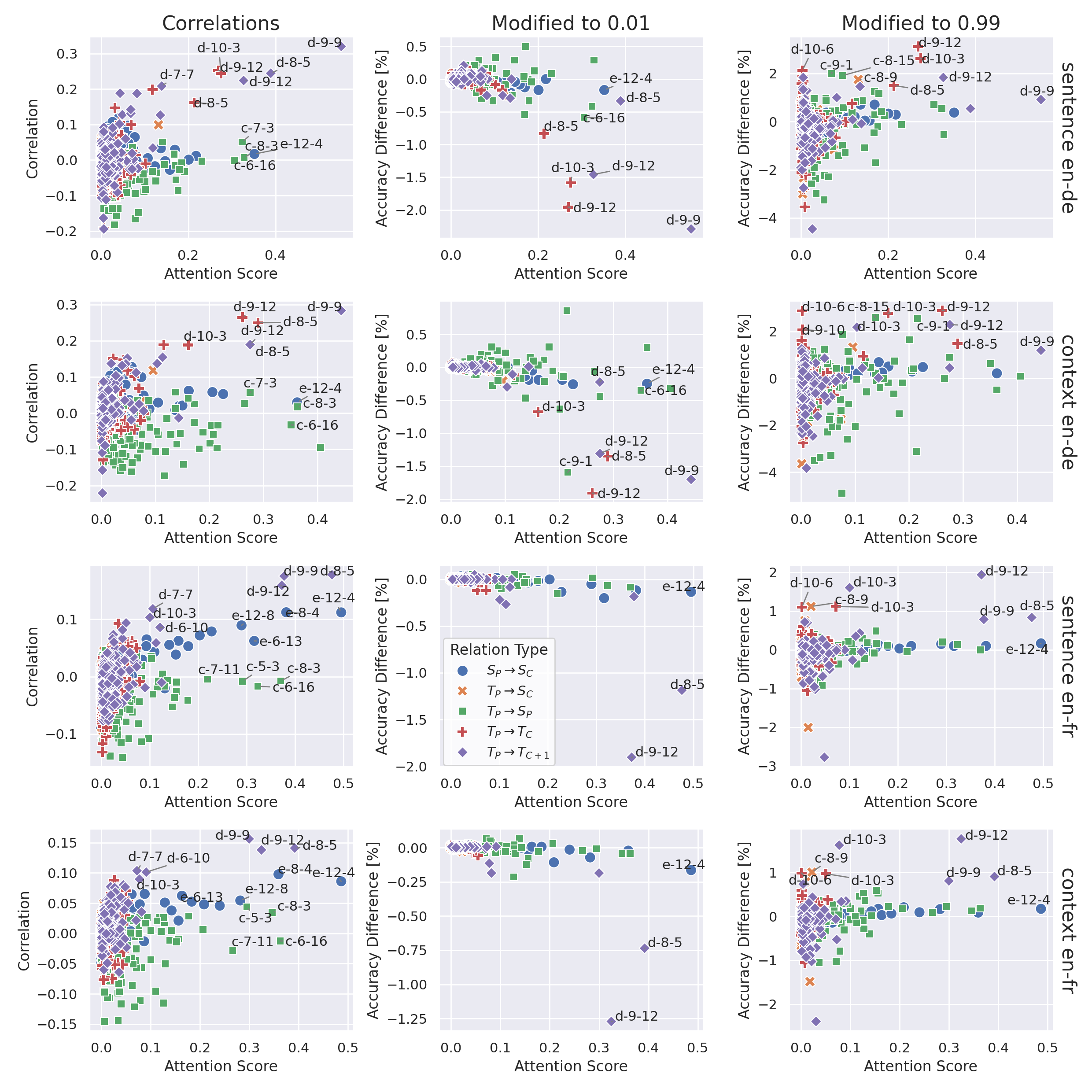}
    \caption{Results in terms of calculated metrics (correlations, difference in accuracy when modified to $0.01$, and modified to $0.99$; as columns) in relation to the averaged attention scores for the English-to-German and English-to-French directions for both sentence-level and context-aware models (as rows) based on \textbf{NLLB-200}.}
    \label{fig::nllb_600_combined_all}
\end{figure*}

Here, we show the expanded results for the two multi-lingual models (sentence-level and context-aware) based on NLLB-200. In Section~\ref{sec:results-nllb} we presented the measured metrics introduced in Section~\ref{sec:methods} (correlations, accuracy when modified to $0.01$, and modified to $0.99$) in relation to the averaged attention scores only for the context-aware model. The results for both models and both language directions can be seen in Figure~\ref{fig::nllb_600_combined_all}.

\begin{figure*}[t]
  \includegraphics[height=\linewidth, angle=270]{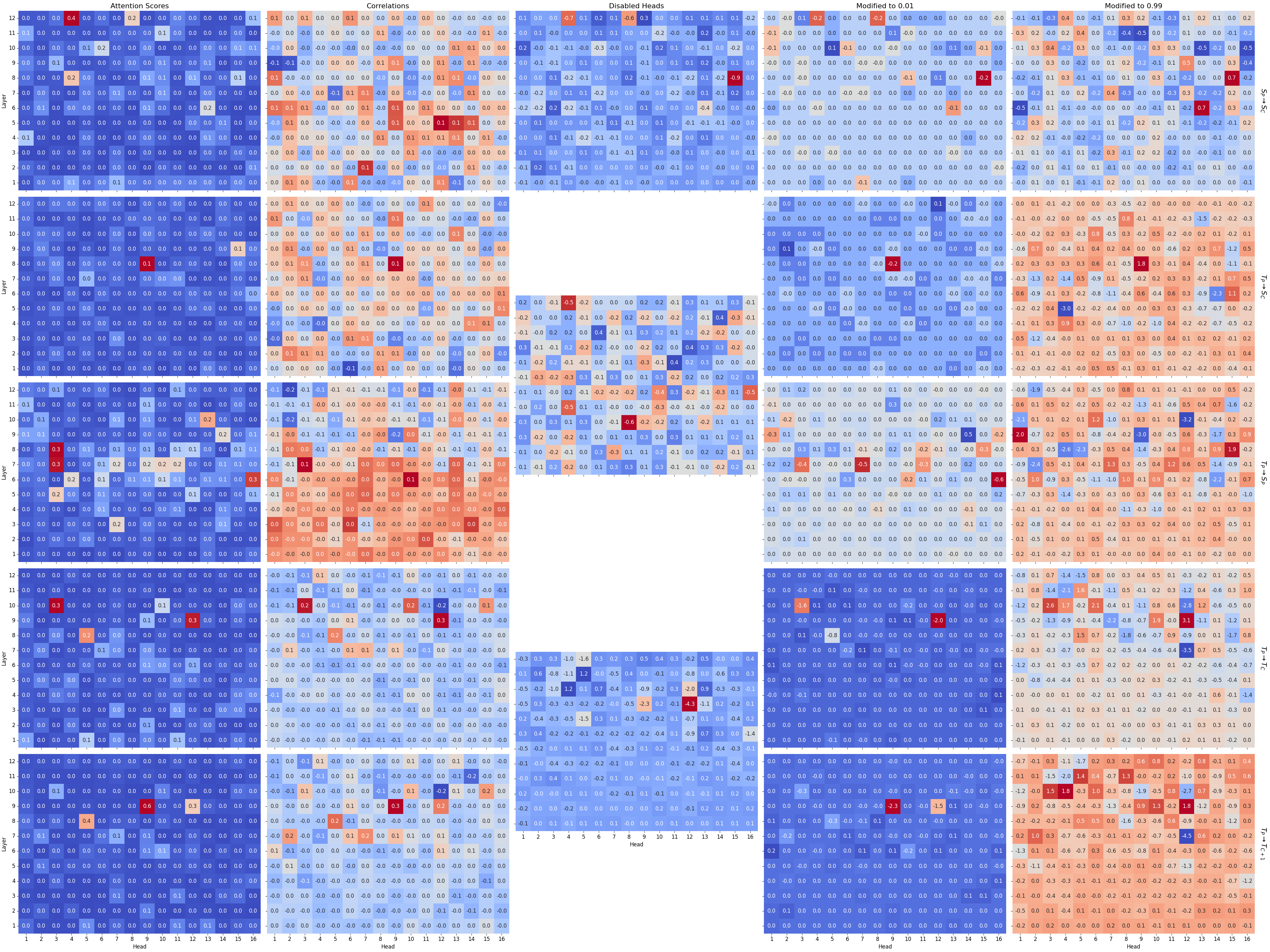}
  \caption{Measured average attention score, correlation between attention scores and accuracy on \textbf{ContraPro} (English-to-German) dataset, and the difference in accuracy for: Disabling Heads, Modifying Heads to $0.01$, and Modifying Heads to $0.99$ (as columns) for each head of the \textbf{sentence-level NLLB-200} model for all relations of interest (as rows).}
  \label{fig:nllb-600M-ende-sentence-combined}
\end{figure*}

\begin{figure*}[t]
  \includegraphics[height=\linewidth, angle=270]{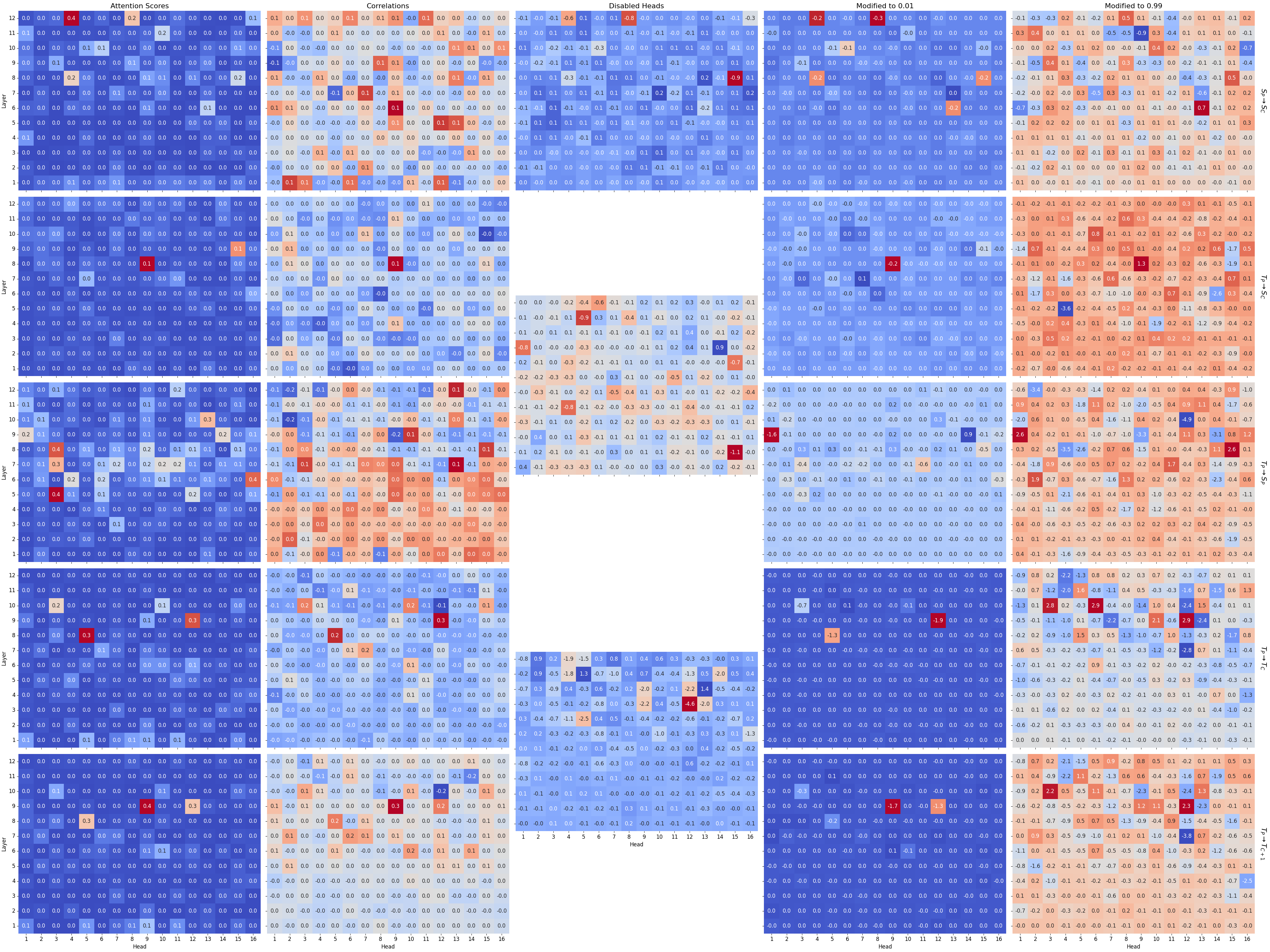}
  \caption{Measured average attention score, correlation between attention scores and accuracy on \textbf{ContraPro} (English-to-German) dataset, and the  difference in for: Disabling Heads, Modifying Heads to $0.01$, and Modifying Heads to $0.99$ (as columns) for each head of the \textbf{context-aware NLLB-200} model for all relations of interest (as rows).}
  \label{fig:nllb-600M-ende-ctx-1-combined}
\end{figure*}

\begin{figure*}[t]
  \includegraphics[height=\linewidth, angle=270]{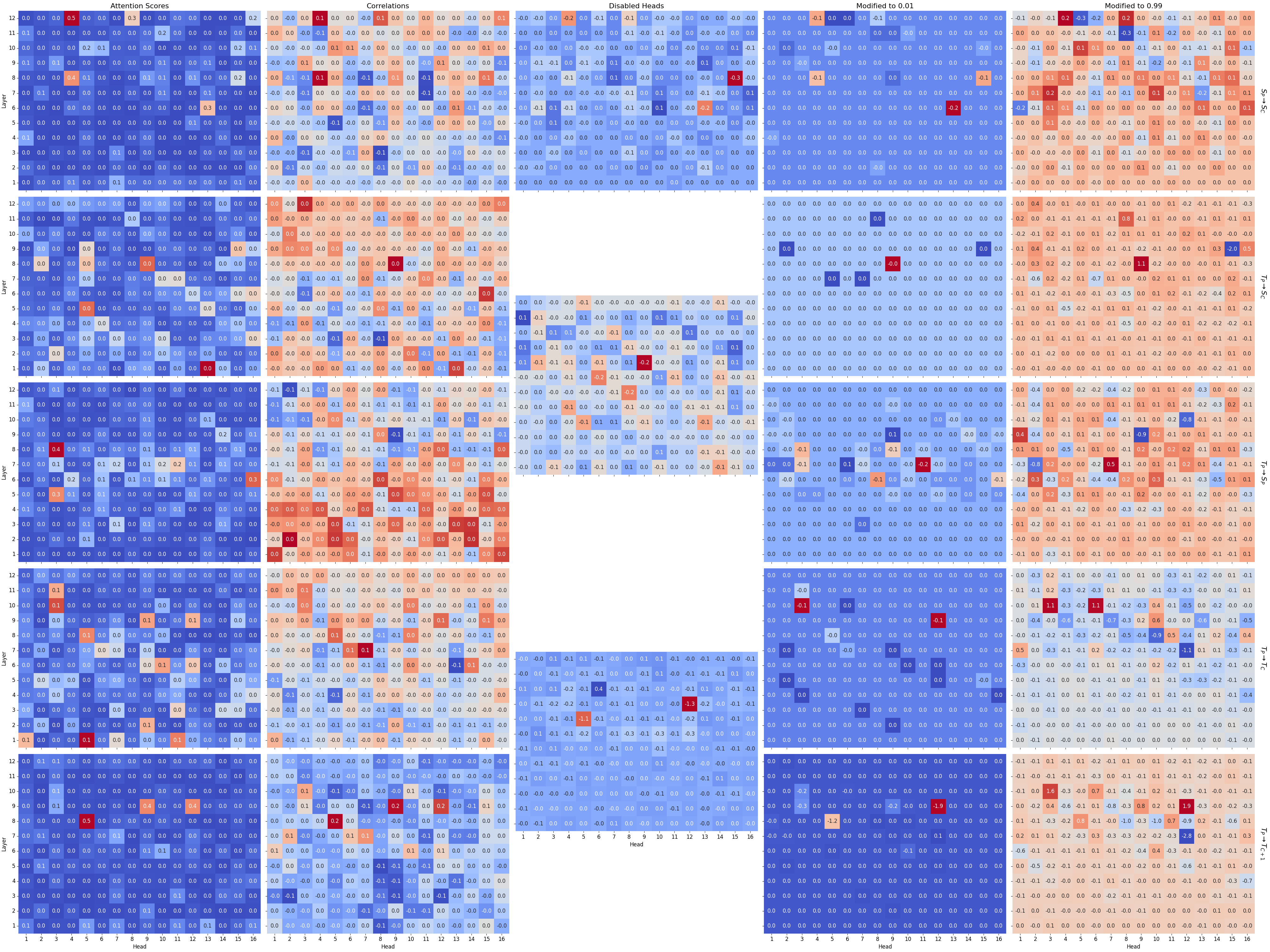}
  \caption{Measured average attention score, correlation between attention scores and accuracy on \textbf{LCPT} (English-to-French) dataset, and the difference in accuracy for: Disabling Heads, Modifying Heads to $0.01$, and Modifying Heads to $0.99$ (as columns) for each head of the \textbf{sentence-level NLLB-200} model for all relations of interest (as rows).}
  \label{fig:nllb-600M-enfr-sentence-combined}
\end{figure*}

\begin{figure*}[t]
  \includegraphics[height=\linewidth, angle=270]{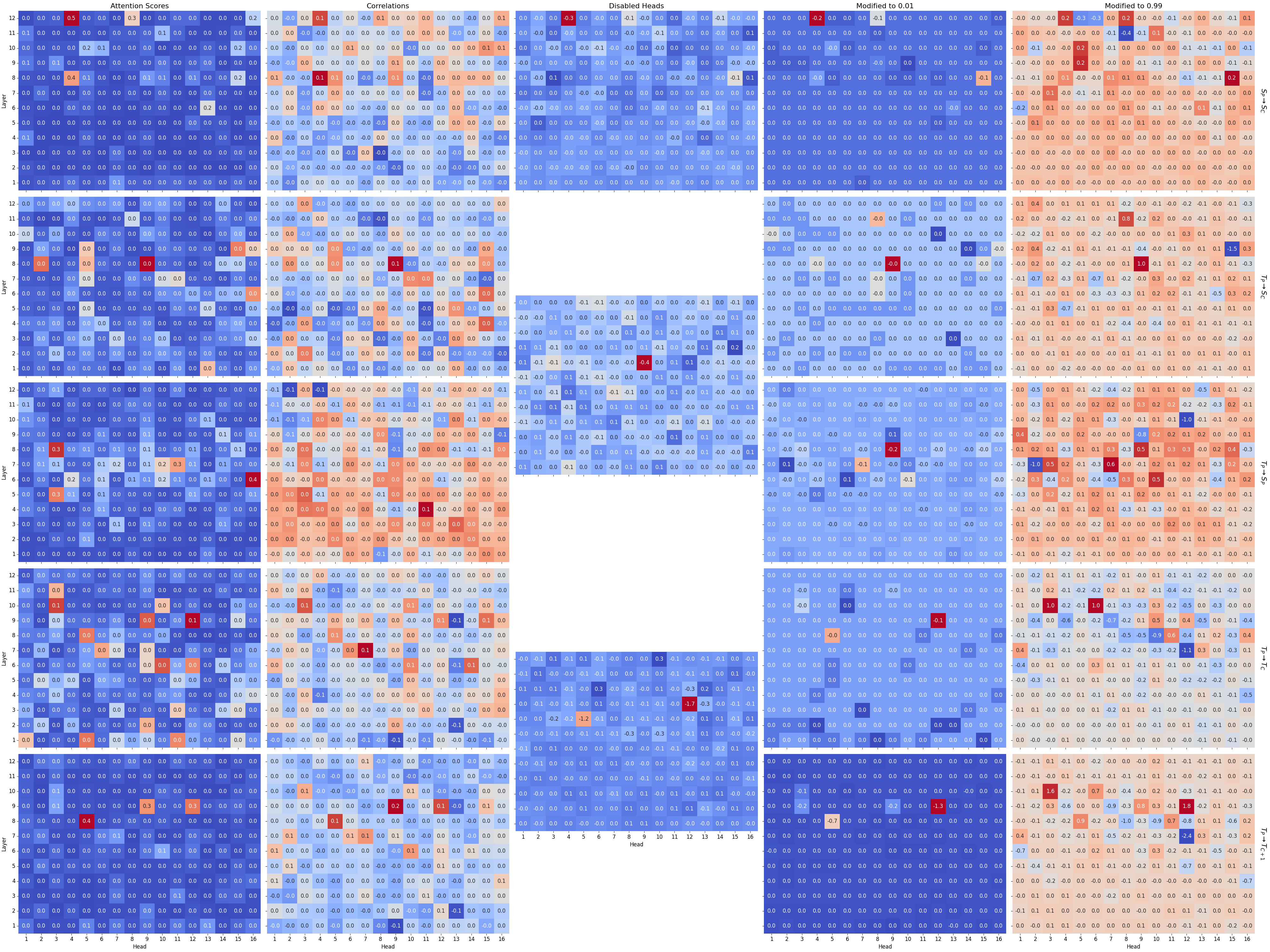}
  \caption{Measured average attention score, correlation between attention scores and accuracy on \textbf{LCPT} (English-to-French) dataset, and the difference in accuracy for: Disabling Heads, Modifying Heads to $0.01$, and Modifying Heads to $0.99$ (as columns) for each head of the \textbf{context-aware NLLB-200} model for all relations of interest (as rows).}
  \label{fig:nllb-600M-enfr-ctx-1-combined}
\end{figure*}

We show the raw results for all relations of interest in: Figure~\ref{fig:nllb-600M-ende-sentence-combined} for the sentence-level model on ContraPro (English-to-German) dataset, Figure~\ref{fig:nllb-600M-ende-ctx-1-combined} for the context-aware model on ContraPro dataset, Figure~\ref{fig:nllb-600M-enfr-sentence-combined} for the sentence-level model on LCPT (English-to-French) dataset, and Figure~\ref{fig:nllb-600M-enfr-ctx-1-combined} for the context-aware model on LCPT dataset.

\section{Histograms of the Results}
\label{sec:histograms}

\begin{figure*}[!ht]
\center{}
    \begin{subfigure}{0.41\linewidth}
        \includegraphics[width=1\linewidth]{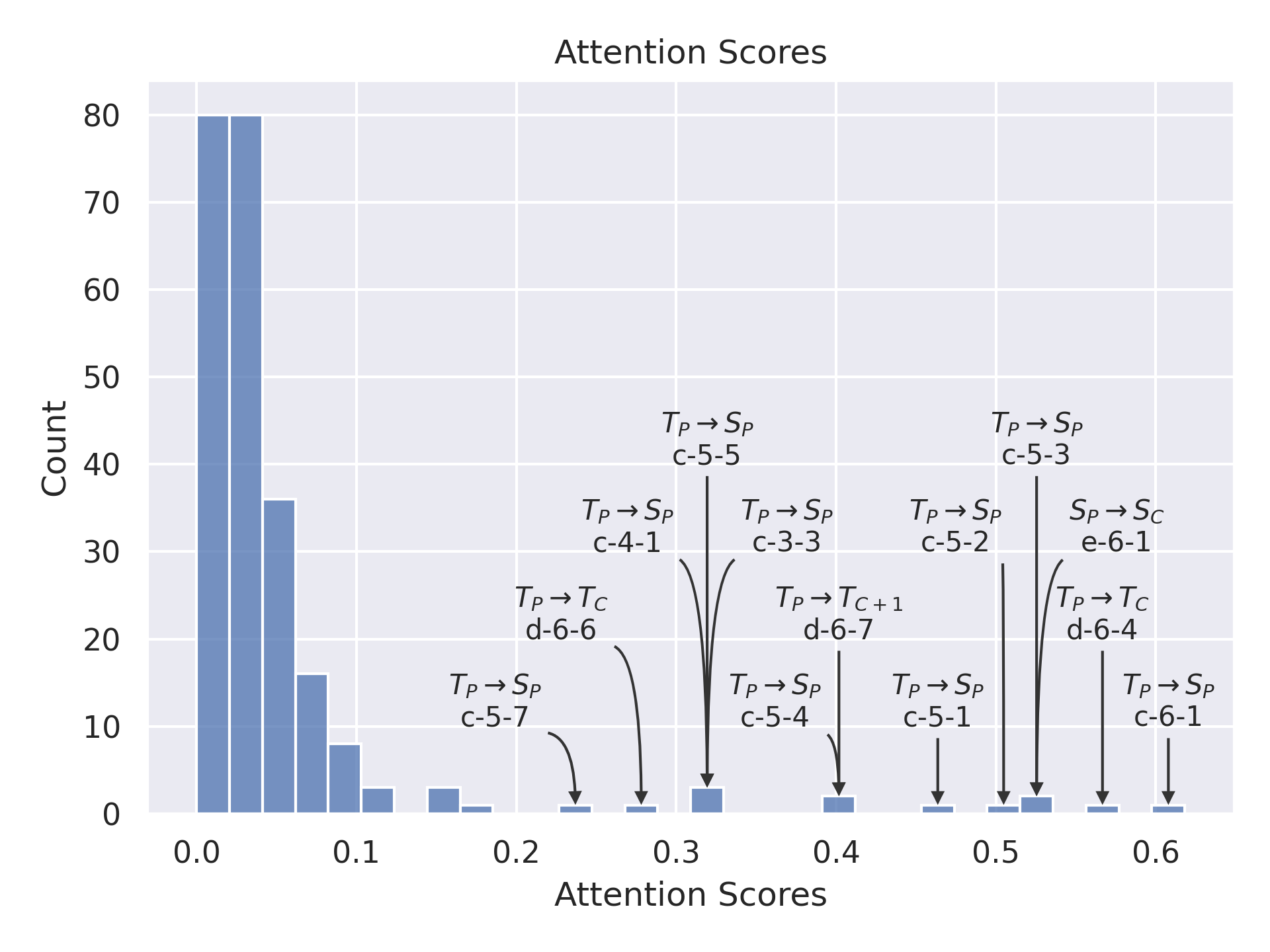}
    \end{subfigure}
    \begin{subfigure}{0.41\linewidth}
        \includegraphics[width=1\linewidth]{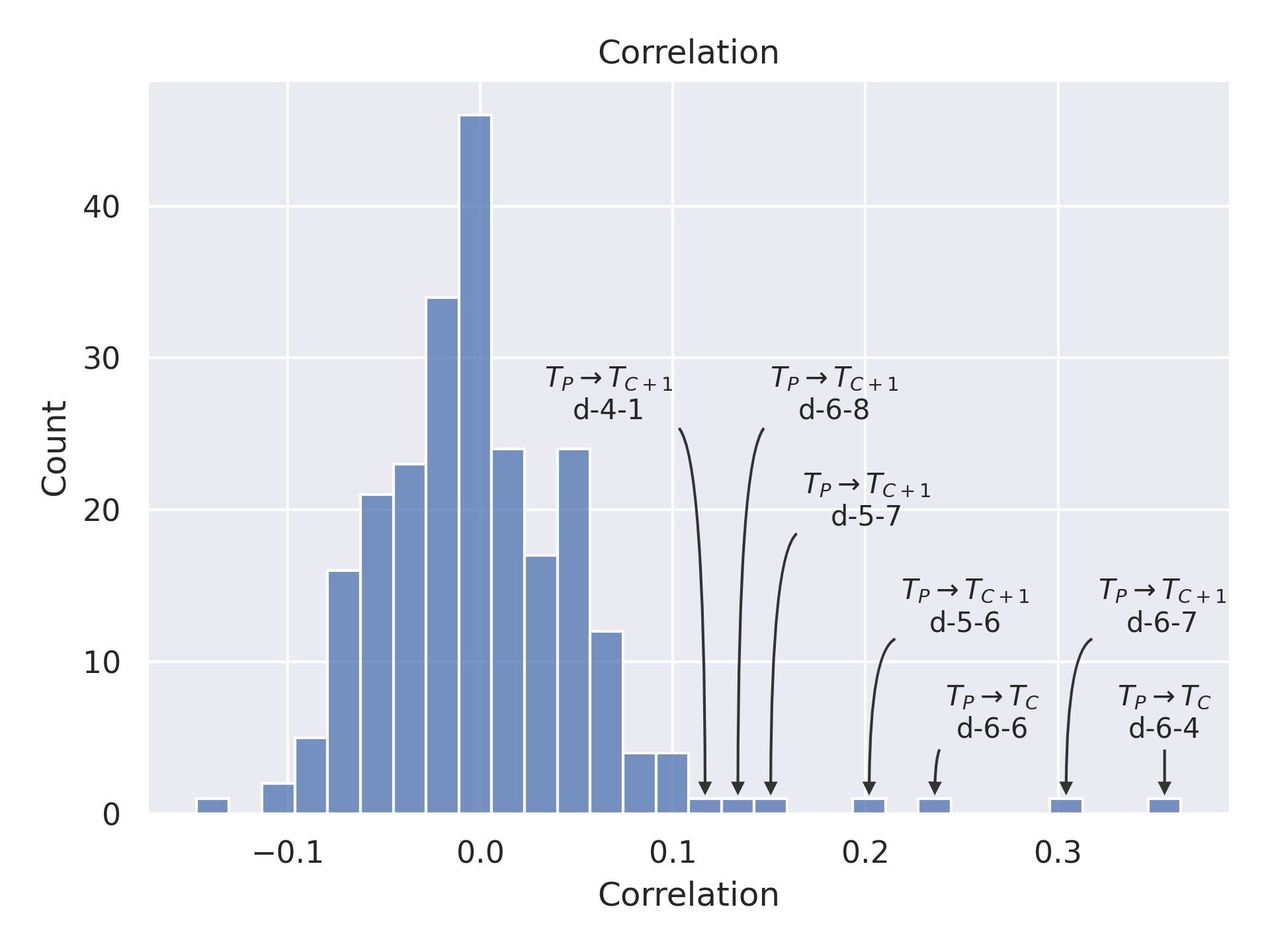}
    \end{subfigure}
    \begin{subfigure}{0.41\linewidth}
        \includegraphics[width=1\linewidth]{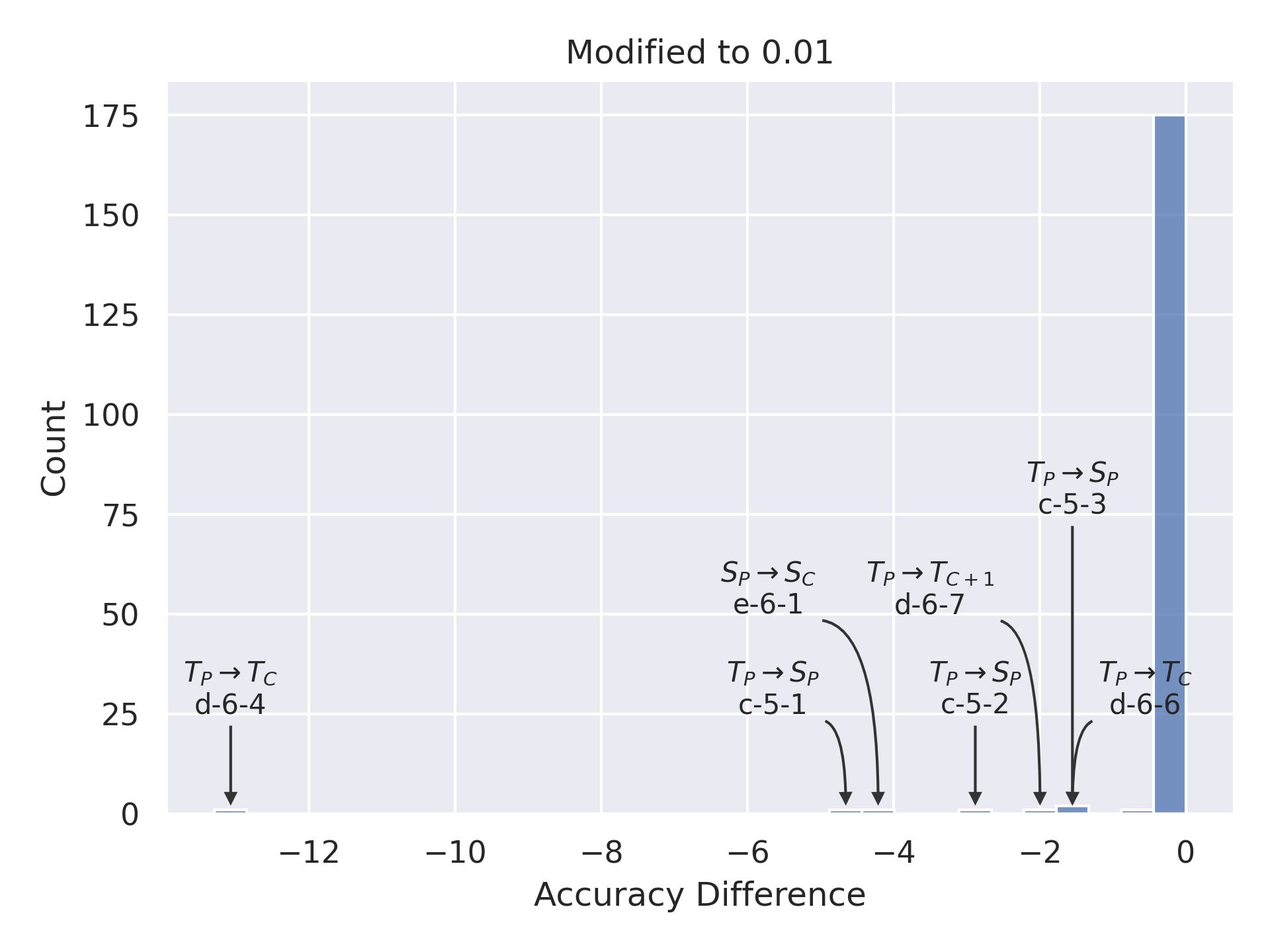}
    \end{subfigure}
    \begin{subfigure}{0.41\linewidth}
        \includegraphics[width=1\linewidth]{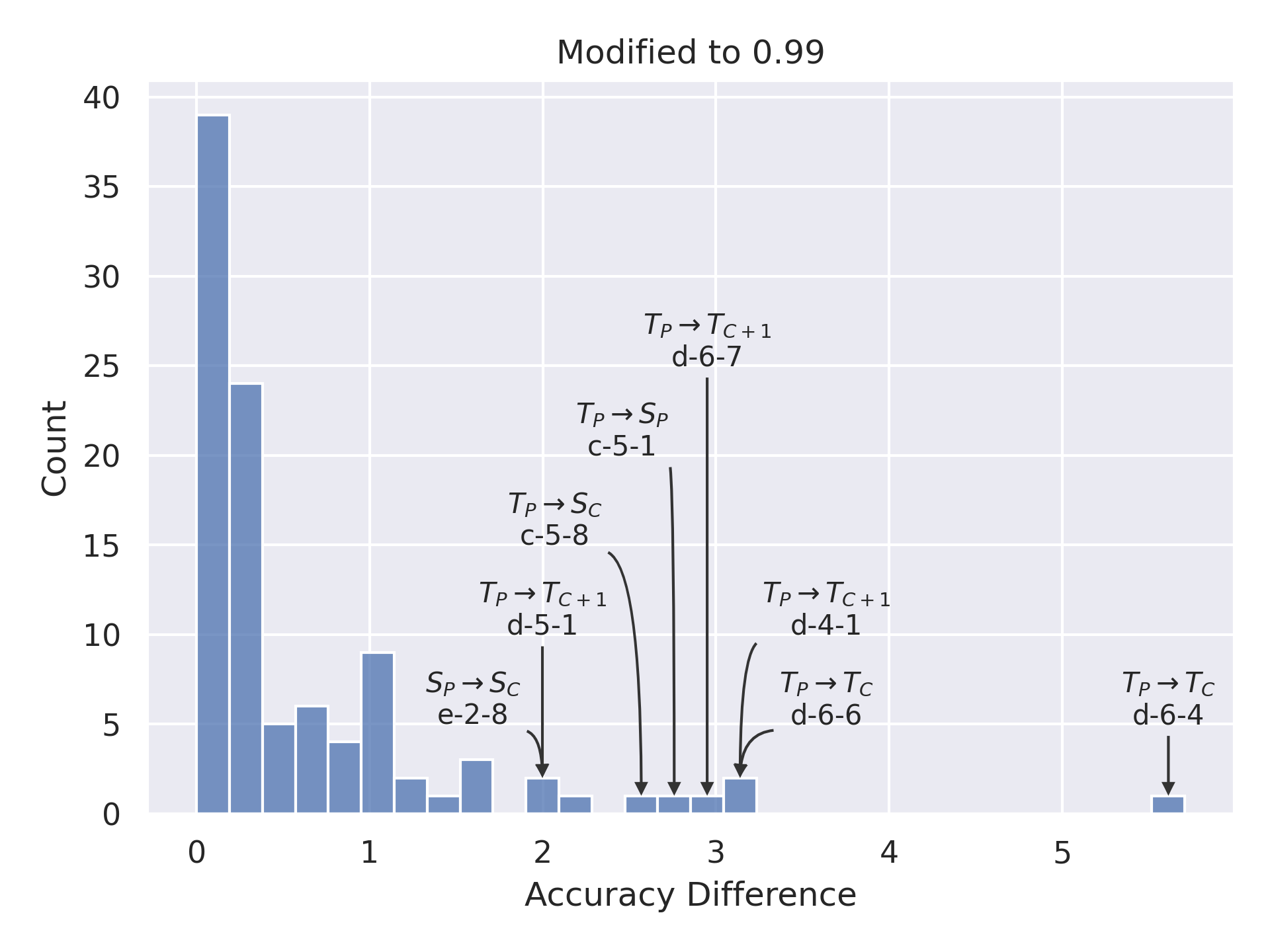}
    \end{subfigure}
    \caption{Histograms of average attention scores, correlations between attention scores and accuracy on ContraPro dataset, and the differences in accuracy for: Modifying Heads to $0.01$ (excluding positive values) and Modifying Heads to $0.99$ (excluding negative values) of each head of the \textbf{sentence-level OpusMT en-de} model with the values of noticeable heads annotated with arrows.}
    \label{fig:hist-opus-mt-sentence}
\end{figure*}

\begin{figure*}[!ht]
\center{}
    \begin{subfigure}{0.41\linewidth}
        \includegraphics[width=1\linewidth]{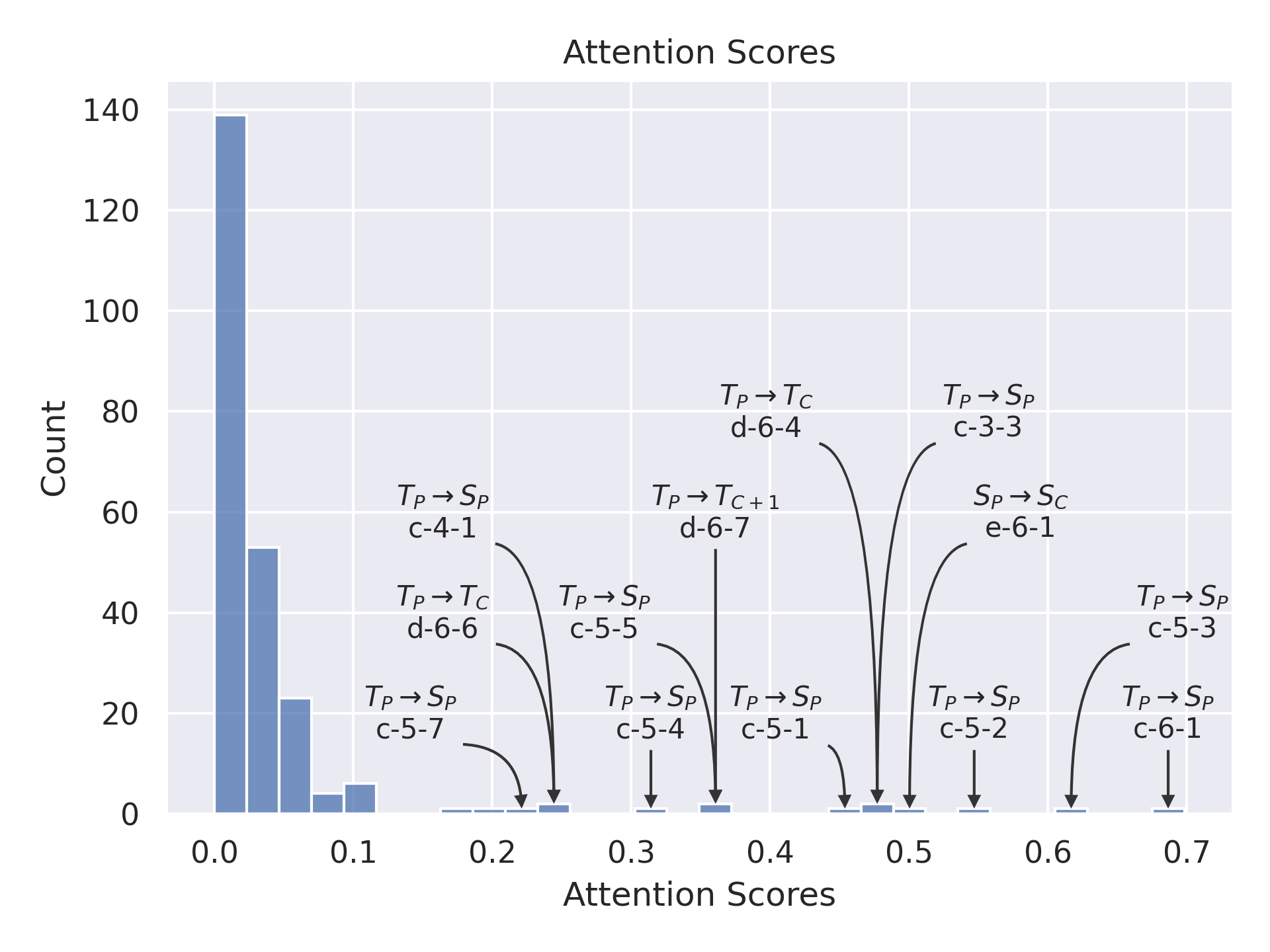}
    \end{subfigure}
    \begin{subfigure}{0.41\linewidth}
        \includegraphics[width=1\linewidth]{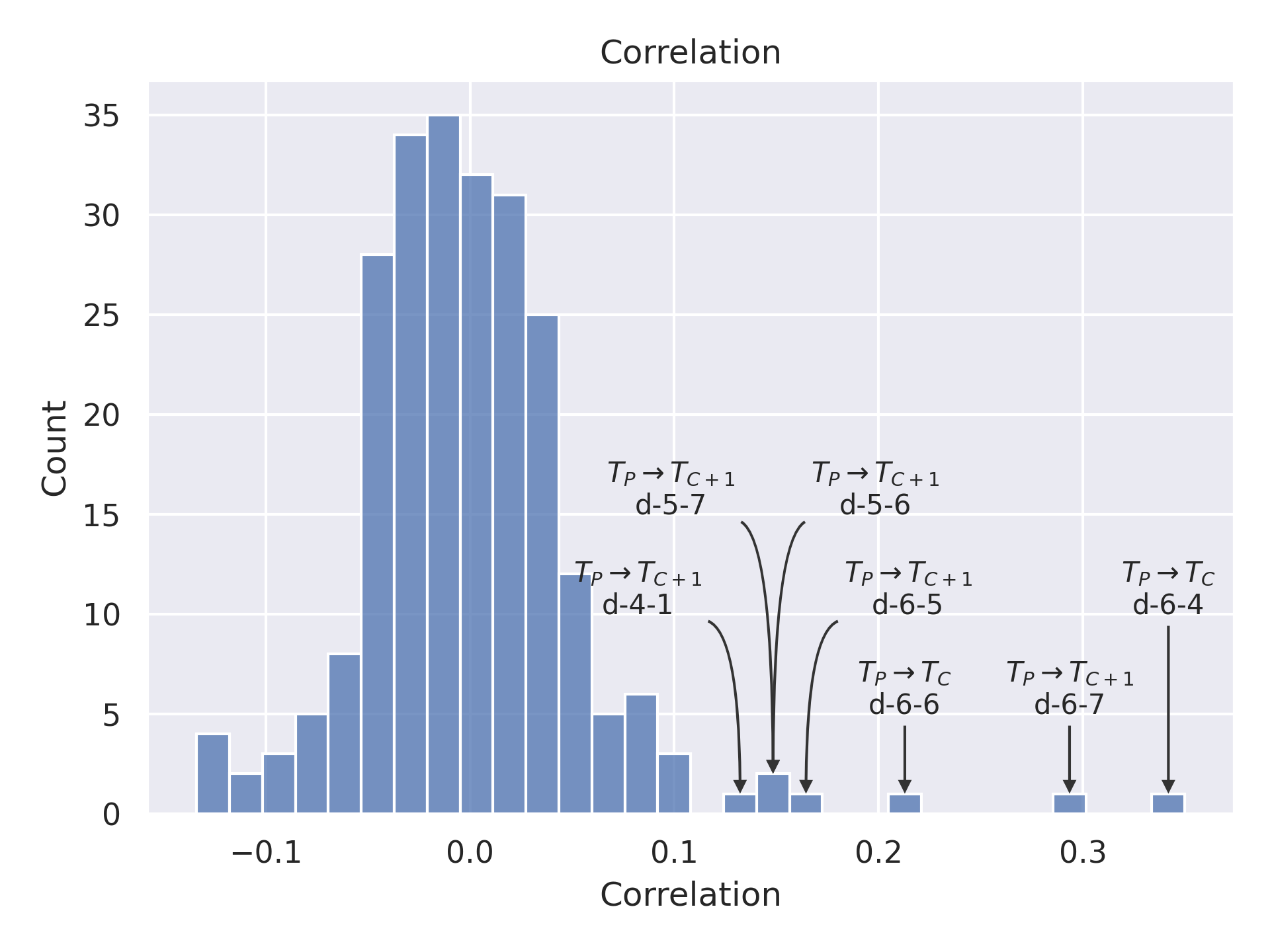}
    \end{subfigure}
    \begin{subfigure}{0.41\linewidth}
        \includegraphics[width=1\linewidth]{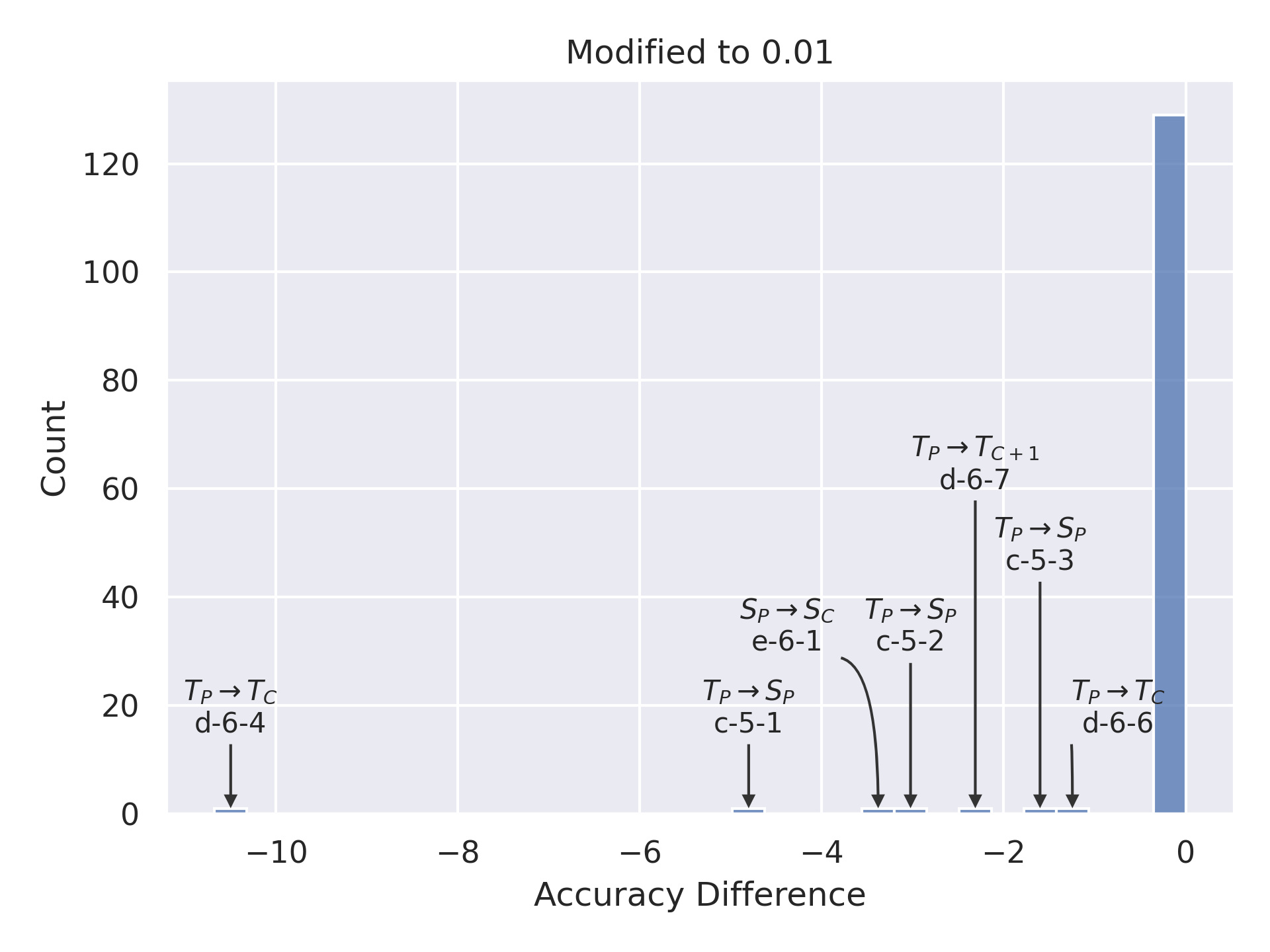}
    \end{subfigure}
    \begin{subfigure}{0.41\linewidth}
        \includegraphics[width=1\linewidth]{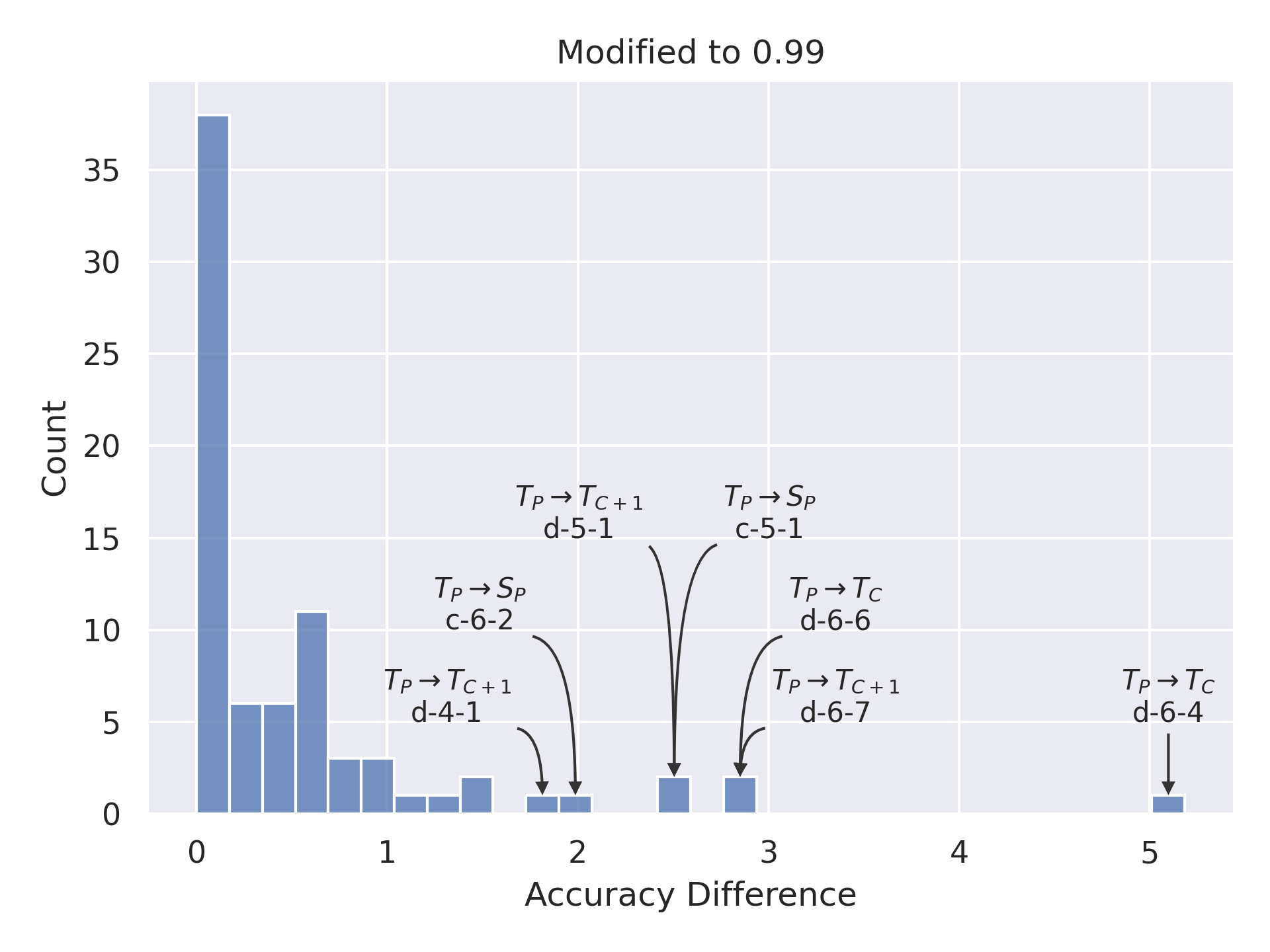}
    \end{subfigure}
    \caption{Histograms of average attention scores, correlations between attention scores and accuracy on ContraPro dataset, and the differences in accuracy for: Modifying Heads to $0.01$ (excluding positive values) and Modifying Heads to $0.99$ (excluding negative values) of each head of the \textbf{context-aware OpusMT en-de} model with the \textbf{context size of one} with the values of noticeable heads annotated with arrows.}
    \label{fig:hist-opus-mt-1}
\end{figure*}

\begin{figure*}[!ht]
\center{}
    \begin{subfigure}{0.41\linewidth}
        \includegraphics[width=1\linewidth]{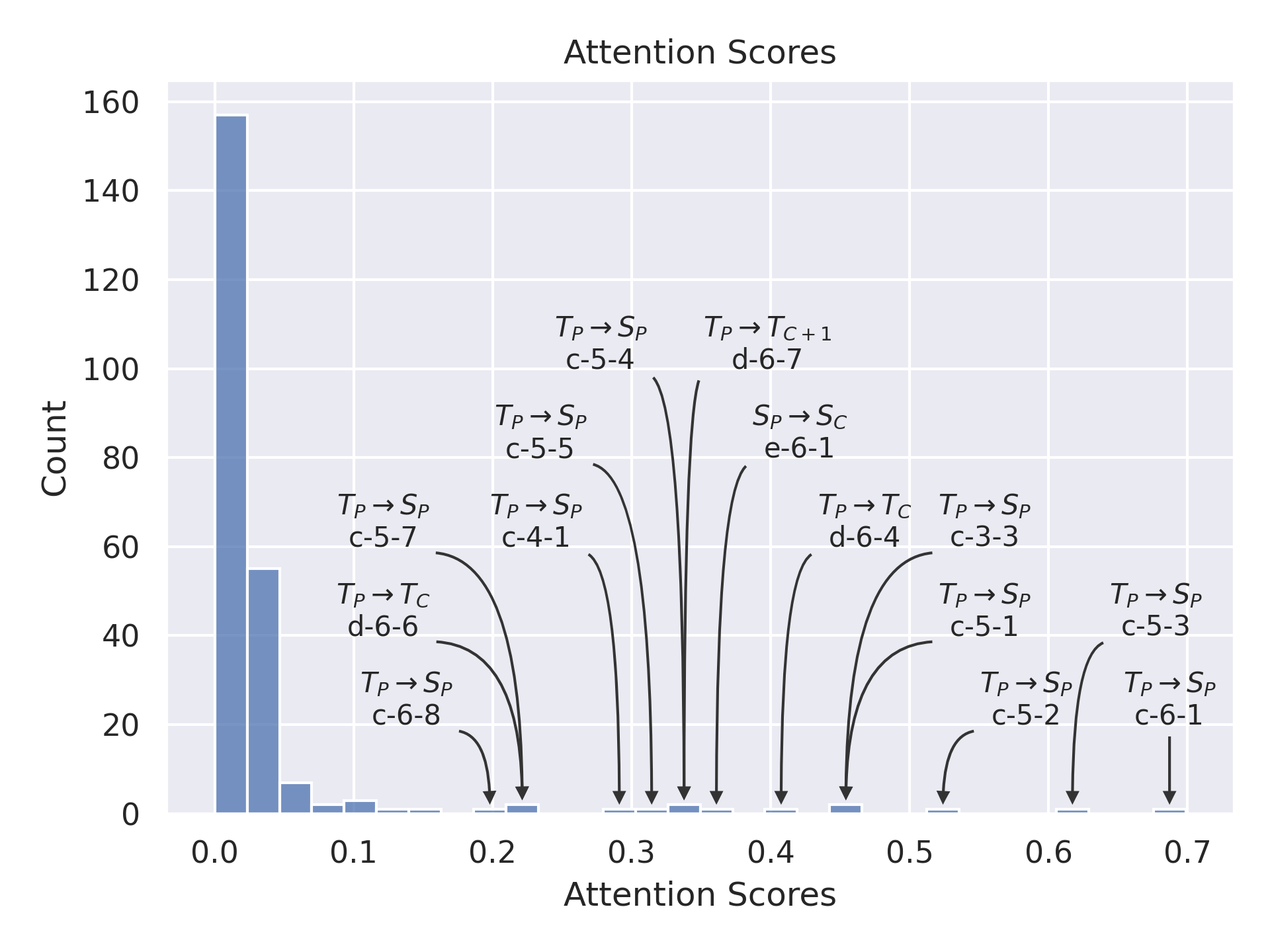}
    \end{subfigure}
    \begin{subfigure}{0.41\linewidth}
        \includegraphics[width=1\linewidth]{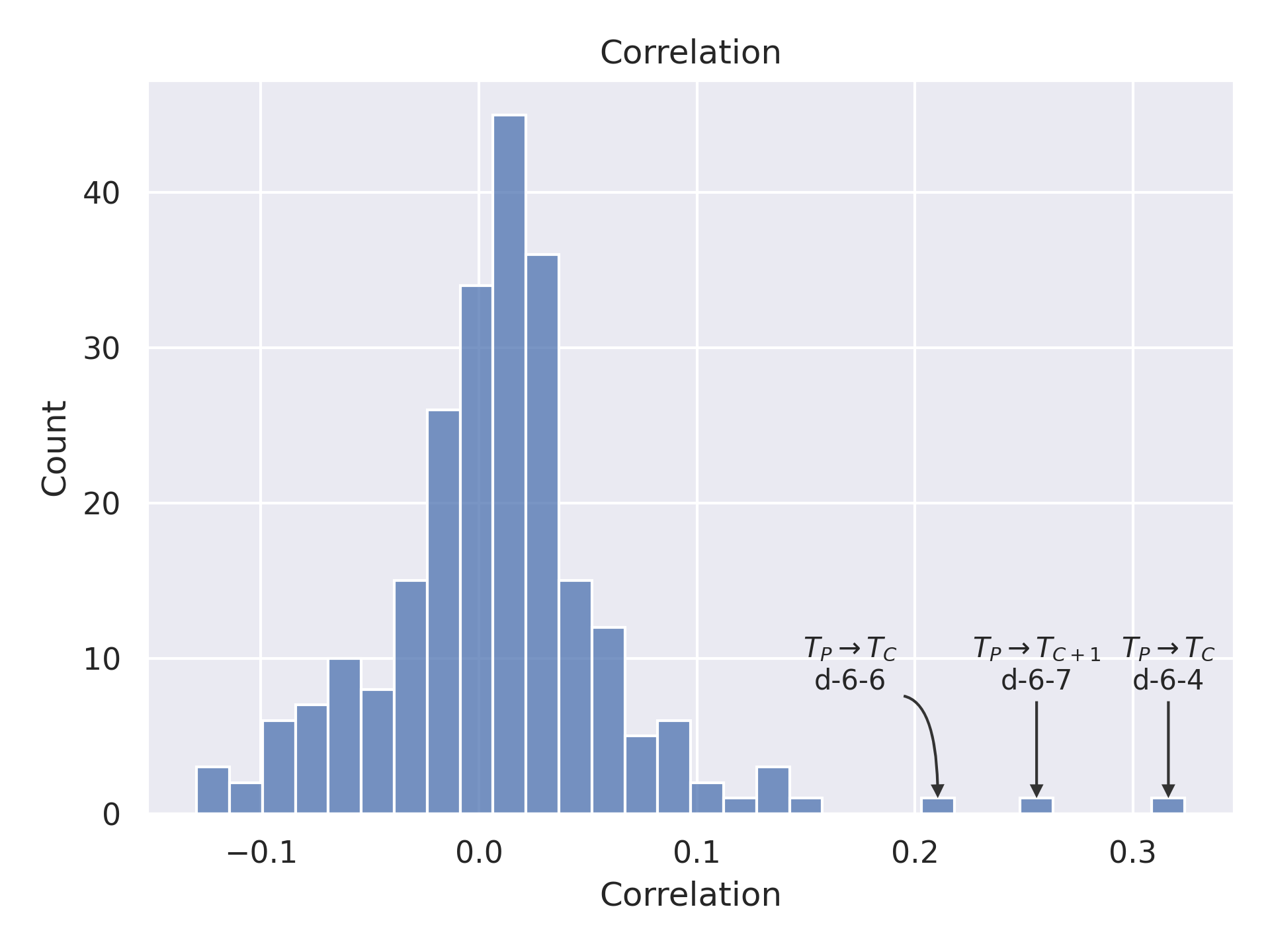}
    \end{subfigure}
    \begin{subfigure}{0.41\linewidth}
        \includegraphics[width=1\linewidth]{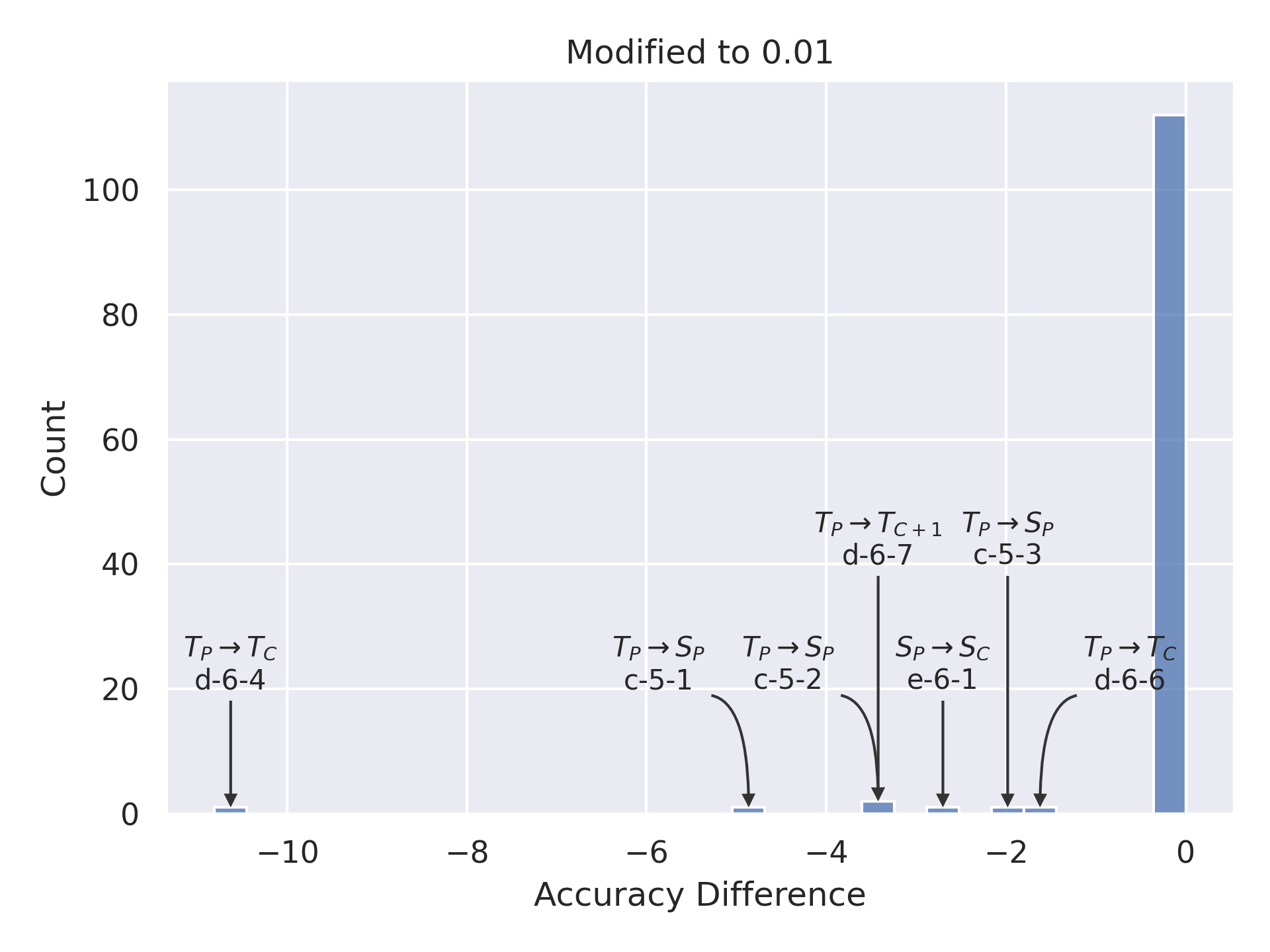}
    \end{subfigure}
    \begin{subfigure}{0.41\linewidth}
        \includegraphics[width=1\linewidth]{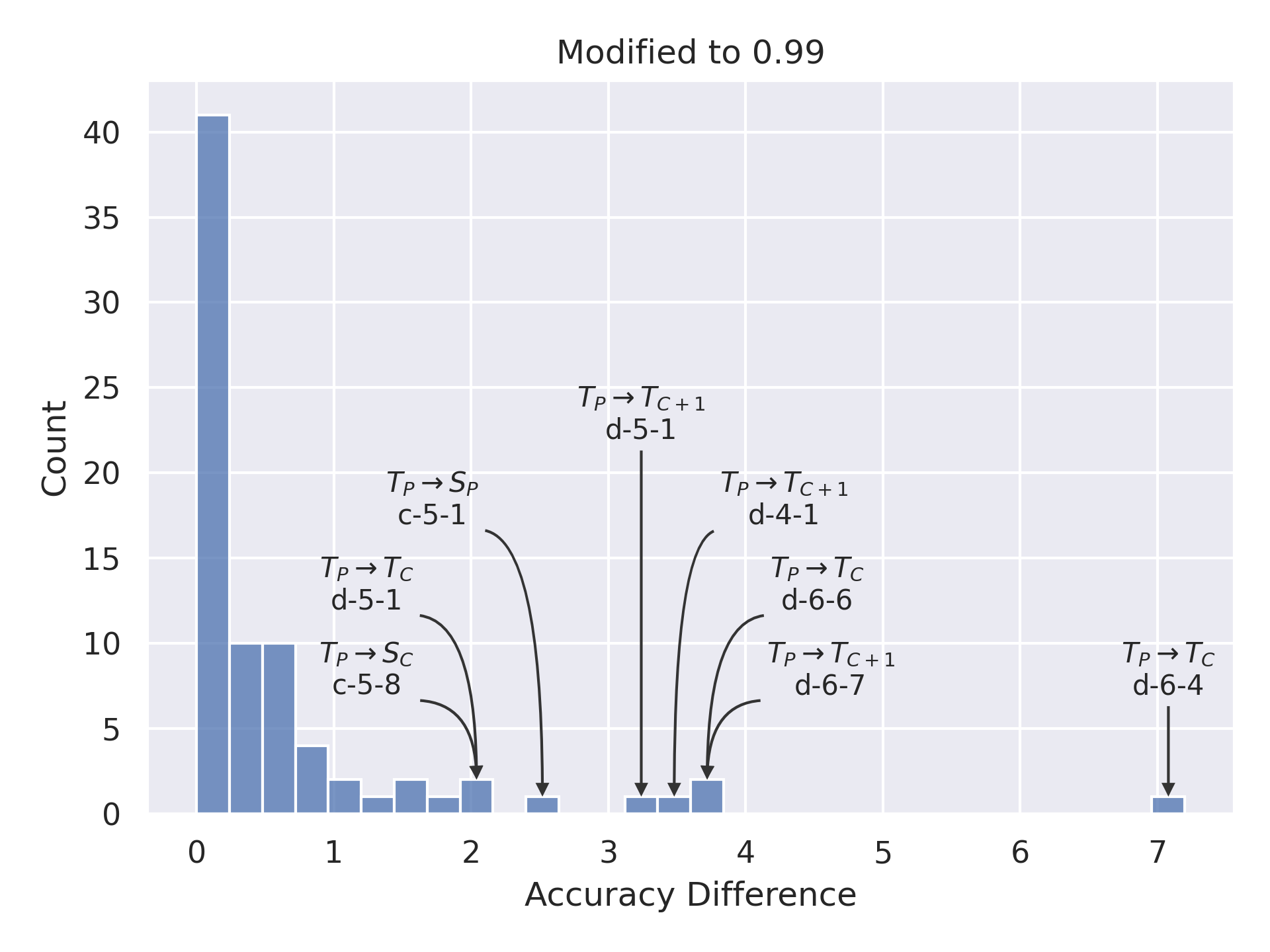}
    \end{subfigure}
    \caption{Histograms of average attention scores, correlations between attention scores and accuracy on ContraPro dataset, and the differences in accuracy for: Modifying Heads to $0.01$ (excluding positive values) and Modifying Heads to $0.99$ (excluding negative values) of each head of the \textbf{context-aware OpusMT en-de} model with the \textbf{context size of three} with the values of noticeable heads annotated with arrows.}
    \label{fig:hist-opus-mt-3}
\end{figure*}

\begin{figure*}[!ht]
\center{}
    \begin{subfigure}{0.41\linewidth}
        \includegraphics[width=1\linewidth]{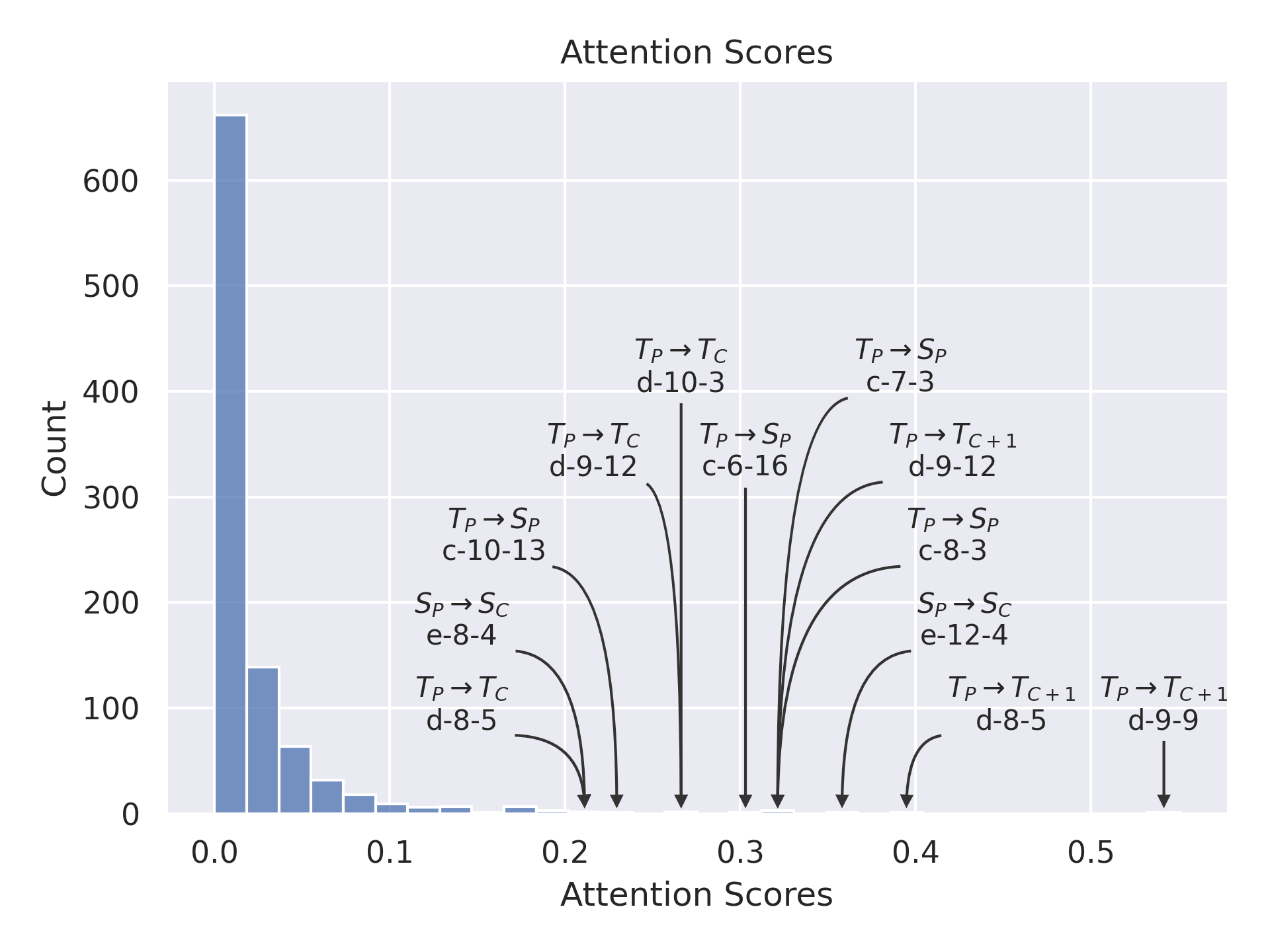}
    \end{subfigure}
    \begin{subfigure}{0.41\linewidth}
        \includegraphics[width=1\linewidth]{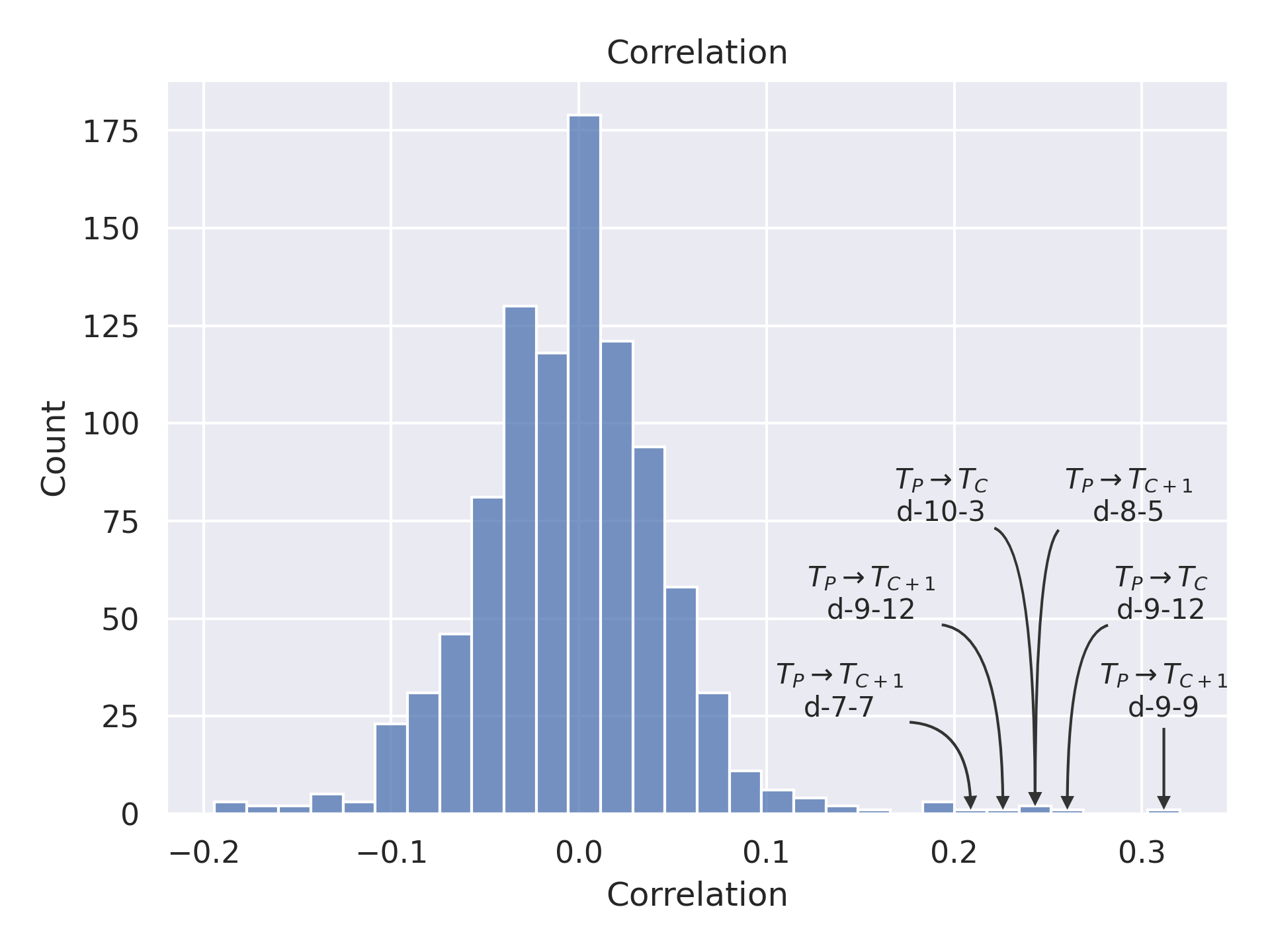}
    \end{subfigure}
    \begin{subfigure}{0.41\linewidth}
        \includegraphics[width=1\linewidth]{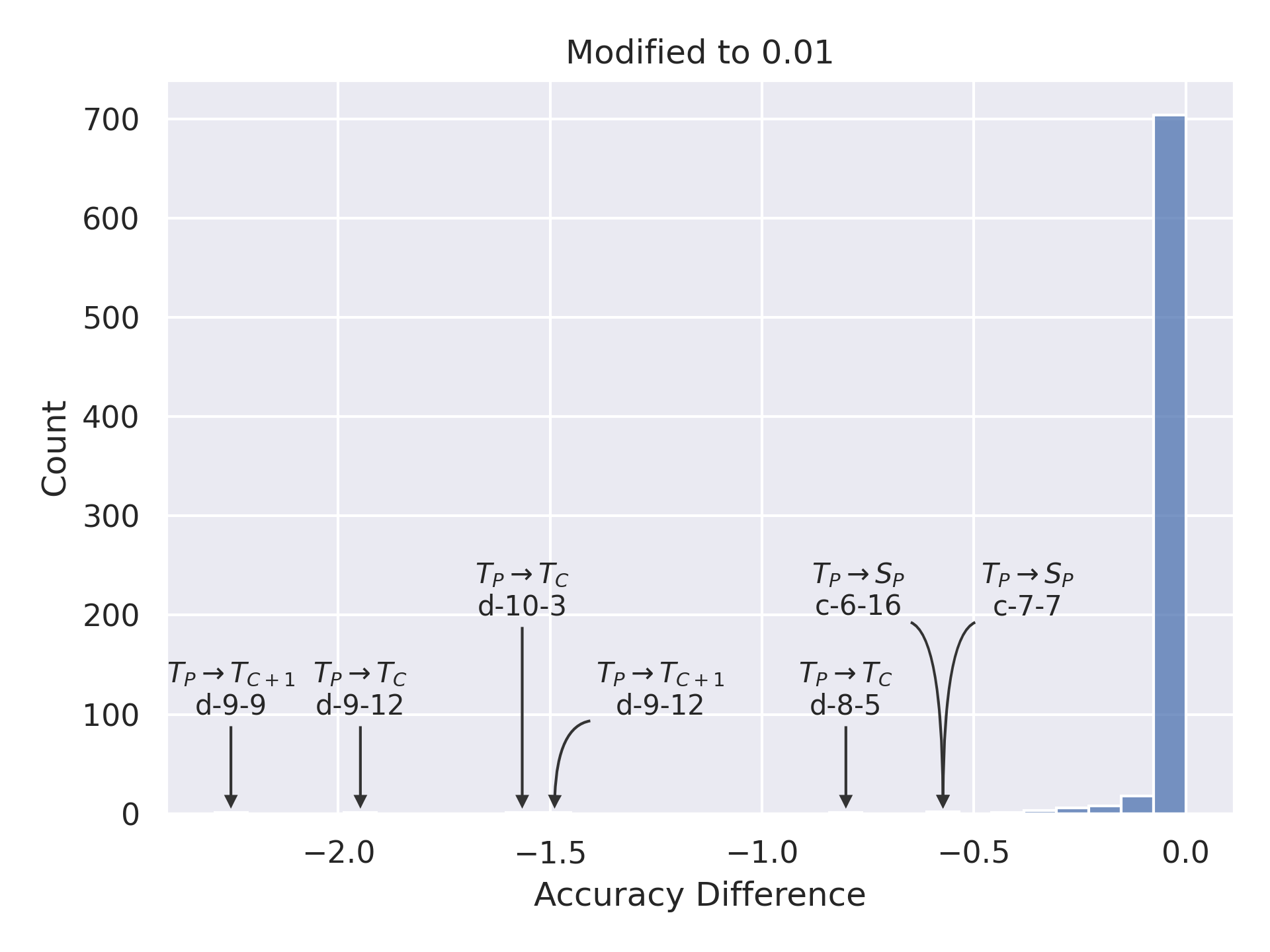}
    \end{subfigure}
    \begin{subfigure}{0.41\linewidth}
        \includegraphics[width=1\linewidth]{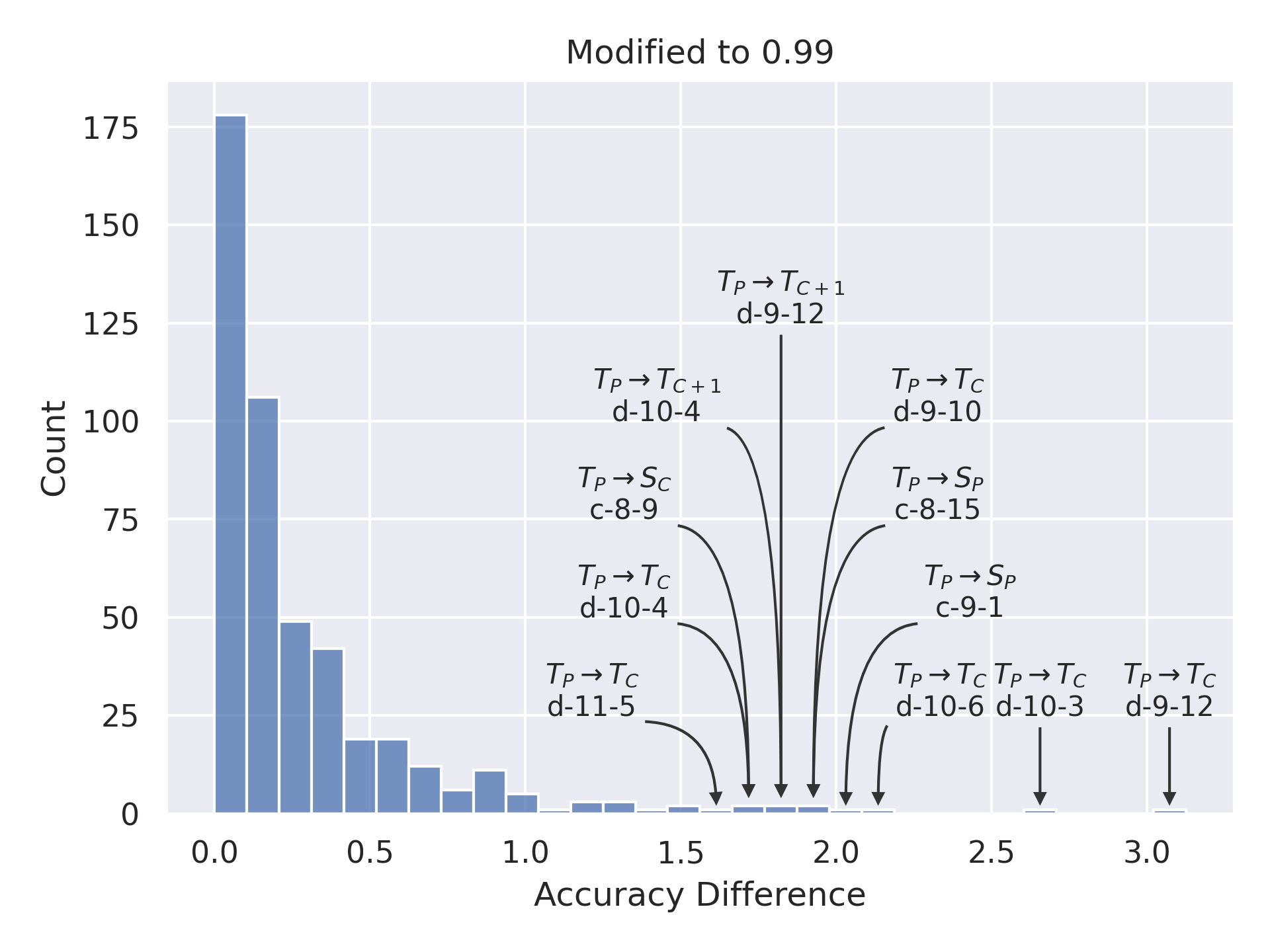}
    \end{subfigure}
    \caption{Histograms of average attention scores, correlations between attention scores and accuracy on \textbf{ContraPro} (English-to-German) dataset, and the differences in accuracy for: Modifying Heads to $0.01$ (excluding positive values) and Modifying Heads to $0.99$ (excluding negative values) of each head of the \textbf{sentence-level NLLB-200} model with the values of noticeable heads annotated with arrows.}
    \label{fig:hist-nllb-ende-sentence}
\end{figure*}

\begin{figure*}[!ht]
\center{}
    \begin{subfigure}{0.41\linewidth}
        \includegraphics[width=1\linewidth]{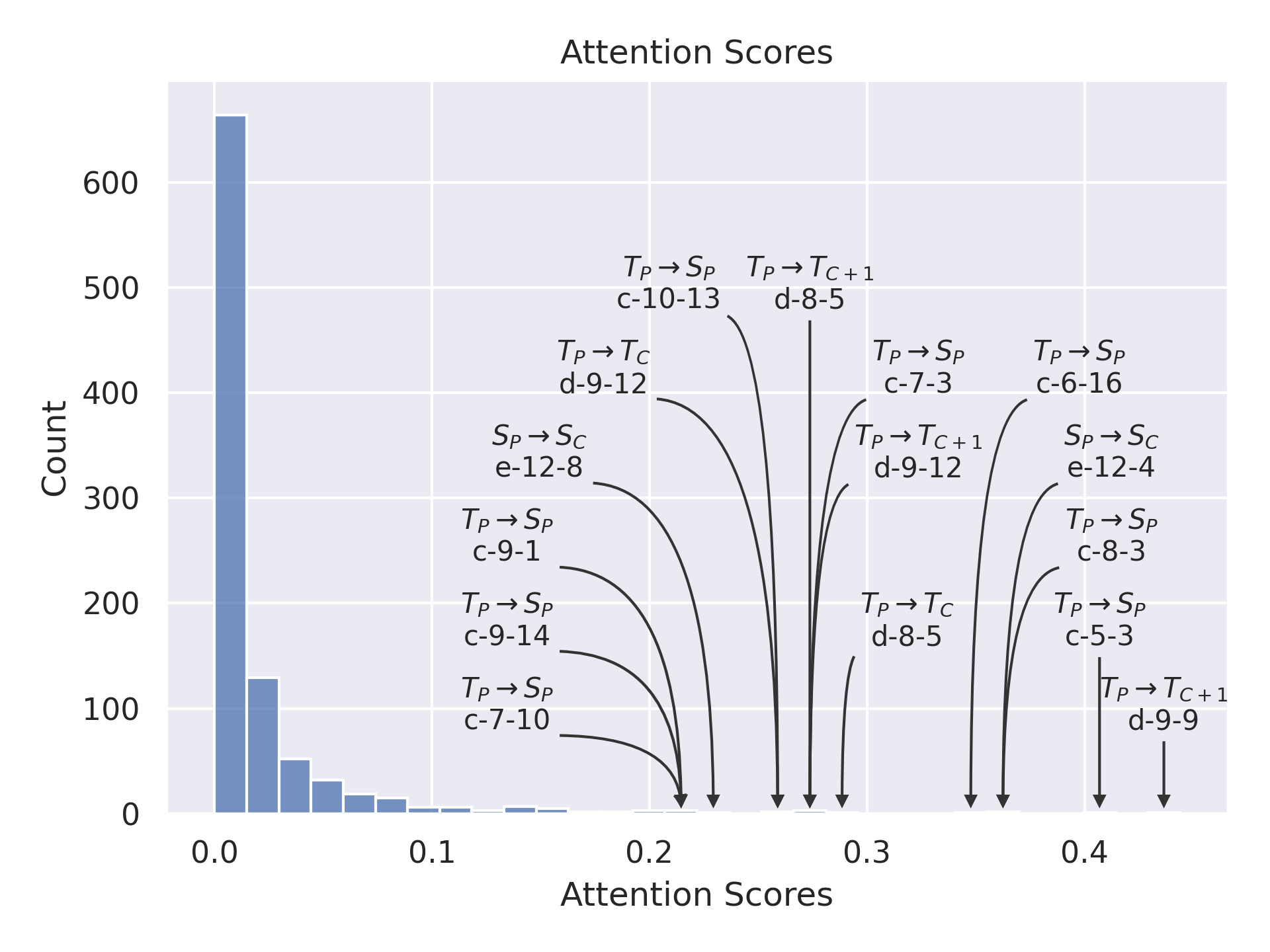}
    \end{subfigure}
    \begin{subfigure}{0.41\linewidth}
        \includegraphics[width=1\linewidth]{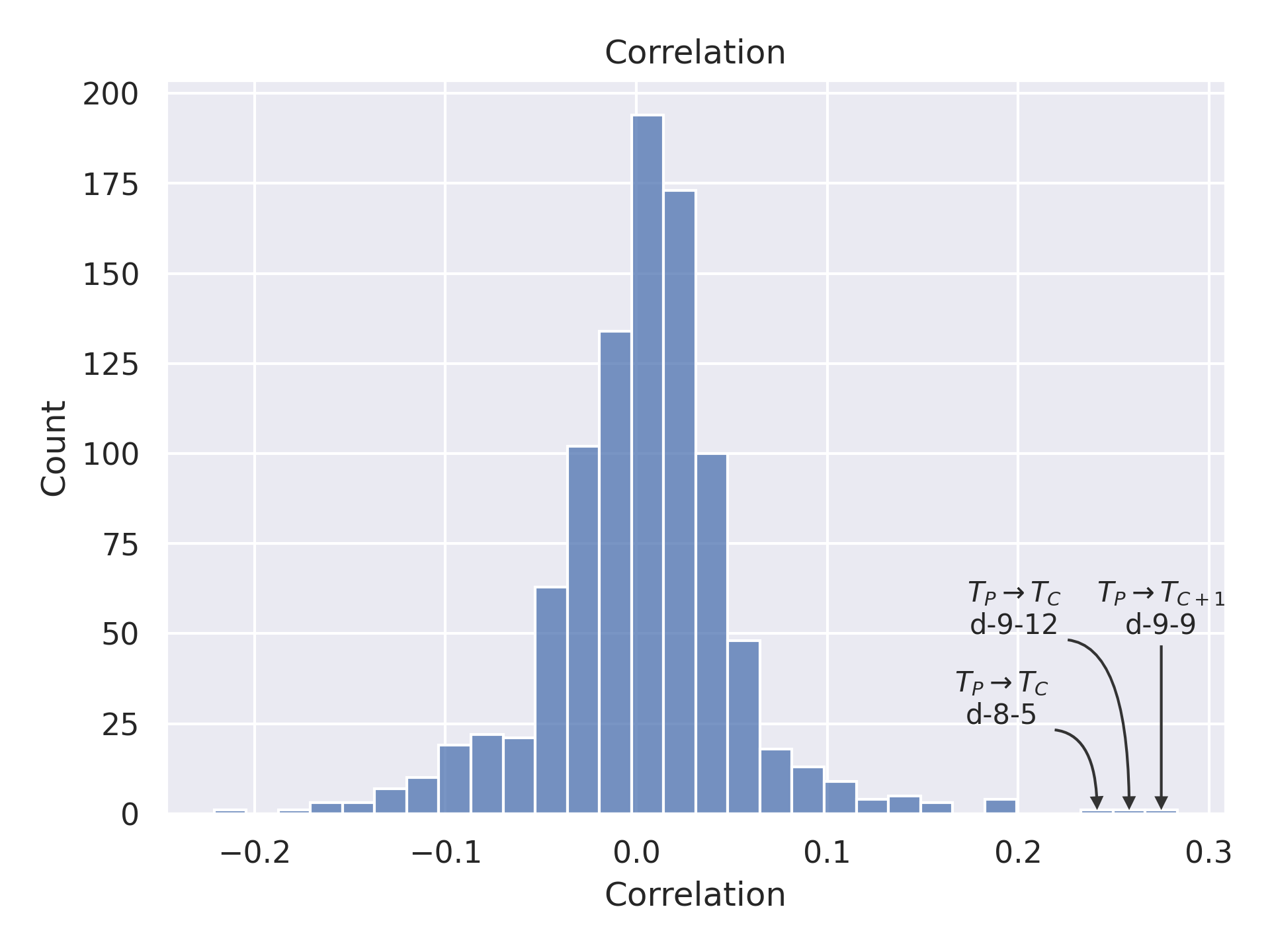}
    \end{subfigure}
    \begin{subfigure}{0.41\linewidth}
        \includegraphics[width=1\linewidth]{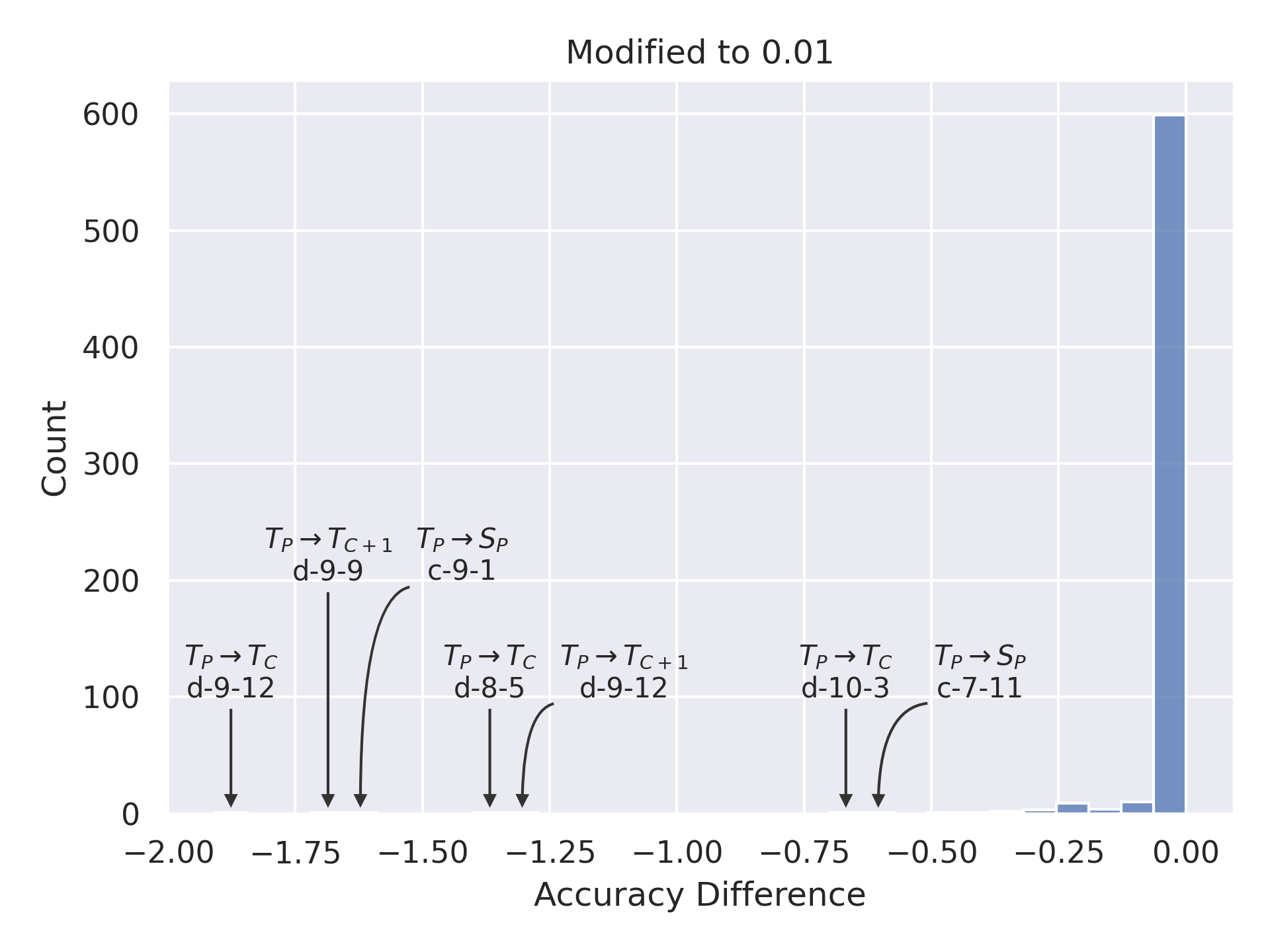}
    \end{subfigure}
    \begin{subfigure}{0.41\linewidth}
        \includegraphics[width=1\linewidth]{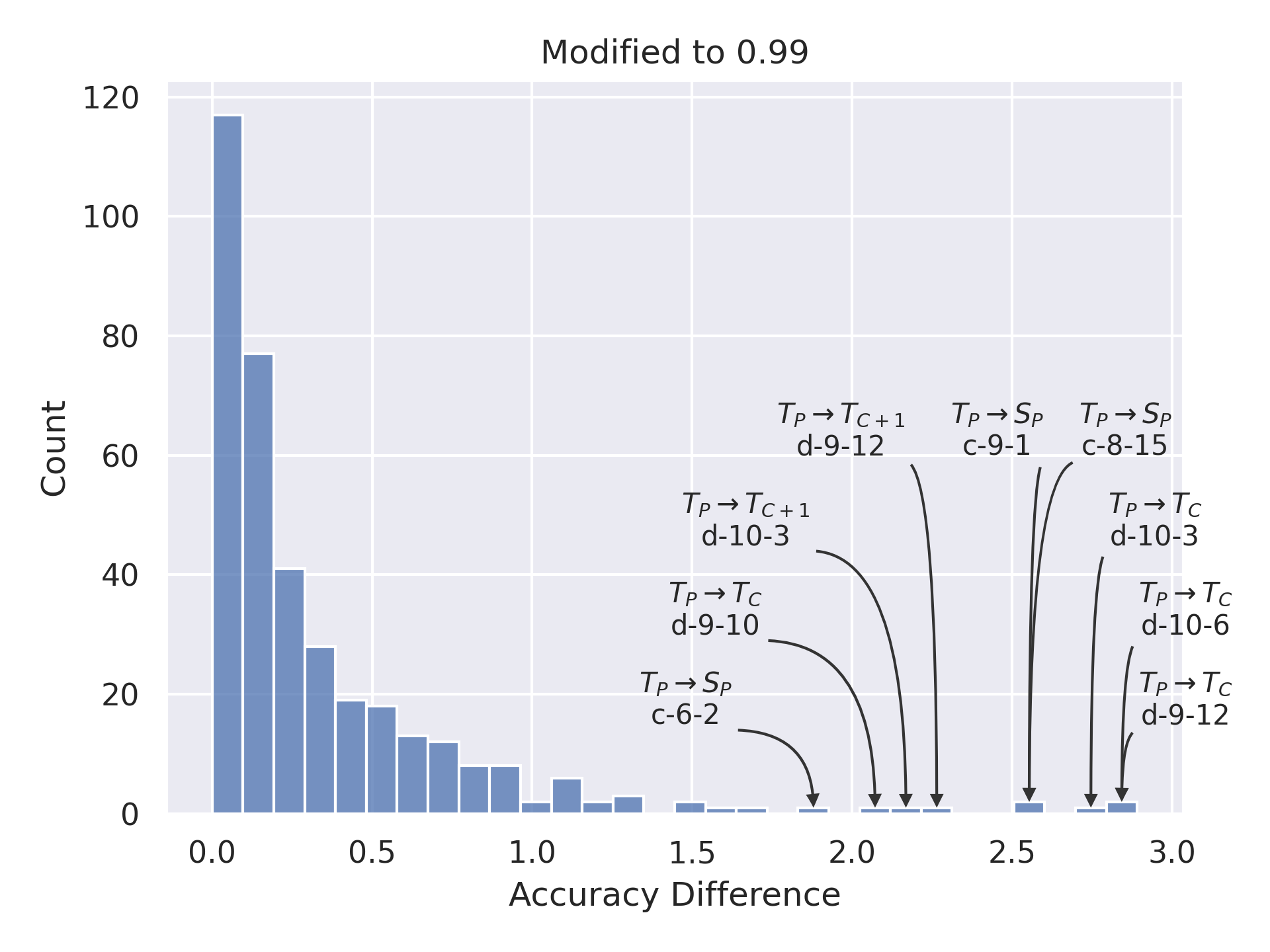}
    \end{subfigure}
    \caption{Histograms of average attention scores, correlations between attention scores and accuracy on \textbf{ContraPro} (English-to-German) dataset, and the differences in accuracy for: Modifying Heads to $0.01$ (excluding positive values) and Modifying Heads to $0.99$ (excluding negative values) of each head of the \textbf{context-aware NLLB-200} model with the values of noticeable heads annotated with arrows.}
    \label{fig:hist-nllb-ende-1}
\end{figure*}

\begin{figure*}[!ht]
\center{}
    \begin{subfigure}{0.41\linewidth}
        \includegraphics[width=1\linewidth]{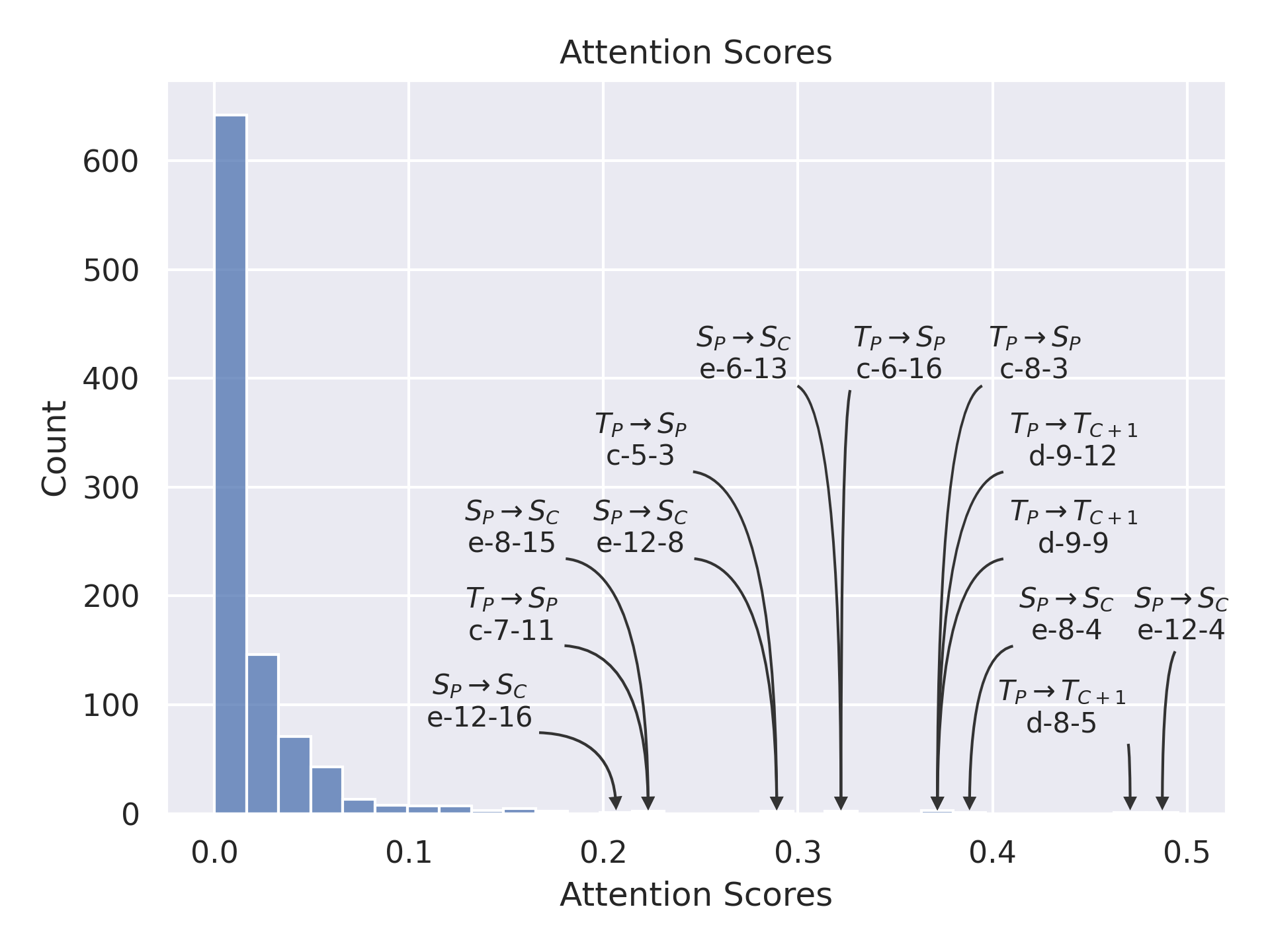}
    \end{subfigure}
    \begin{subfigure}{0.41\linewidth}
        \includegraphics[width=1\linewidth]{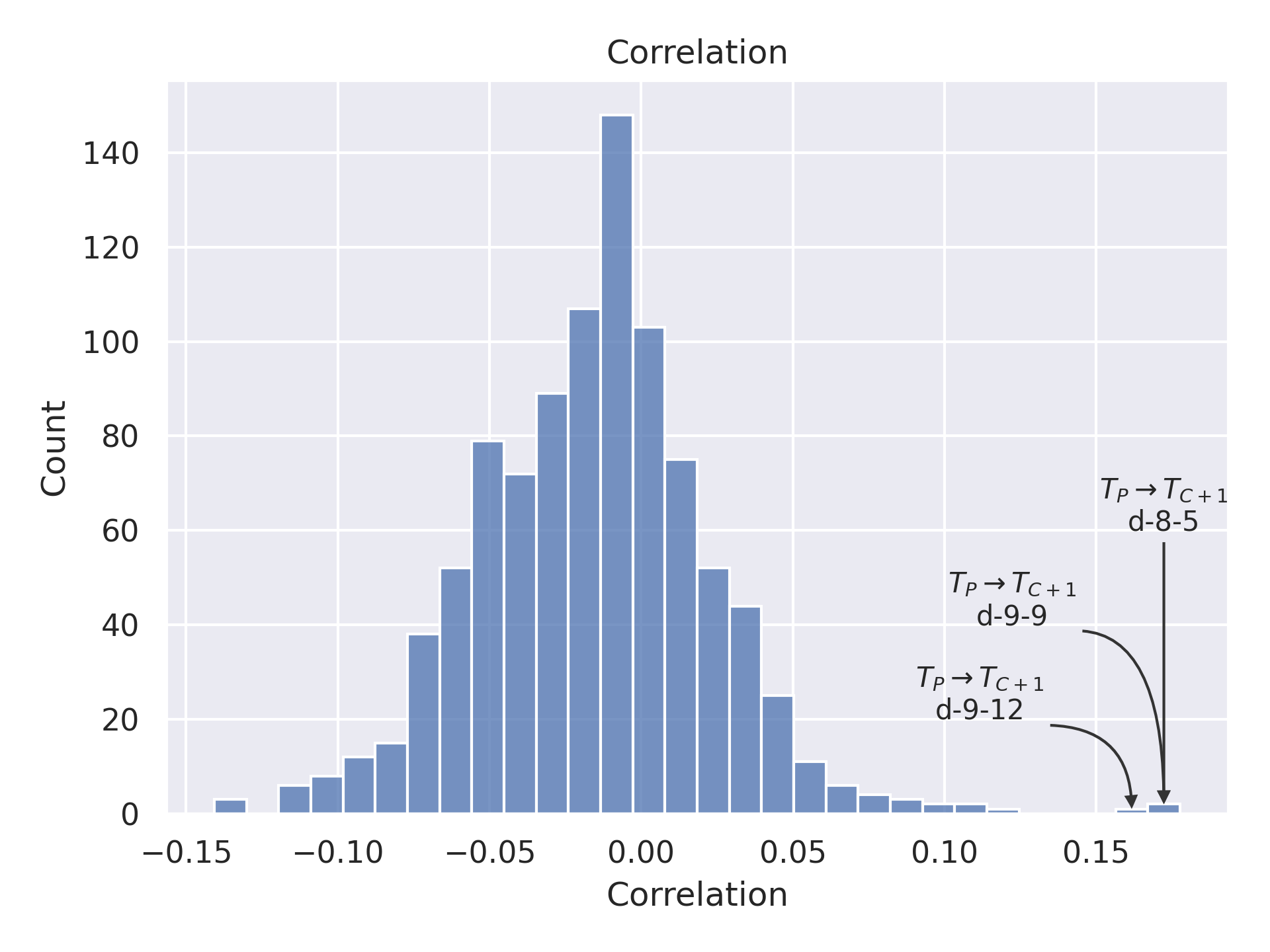}
    \end{subfigure}
    \begin{subfigure}{0.41\linewidth}
        \includegraphics[width=1\linewidth]{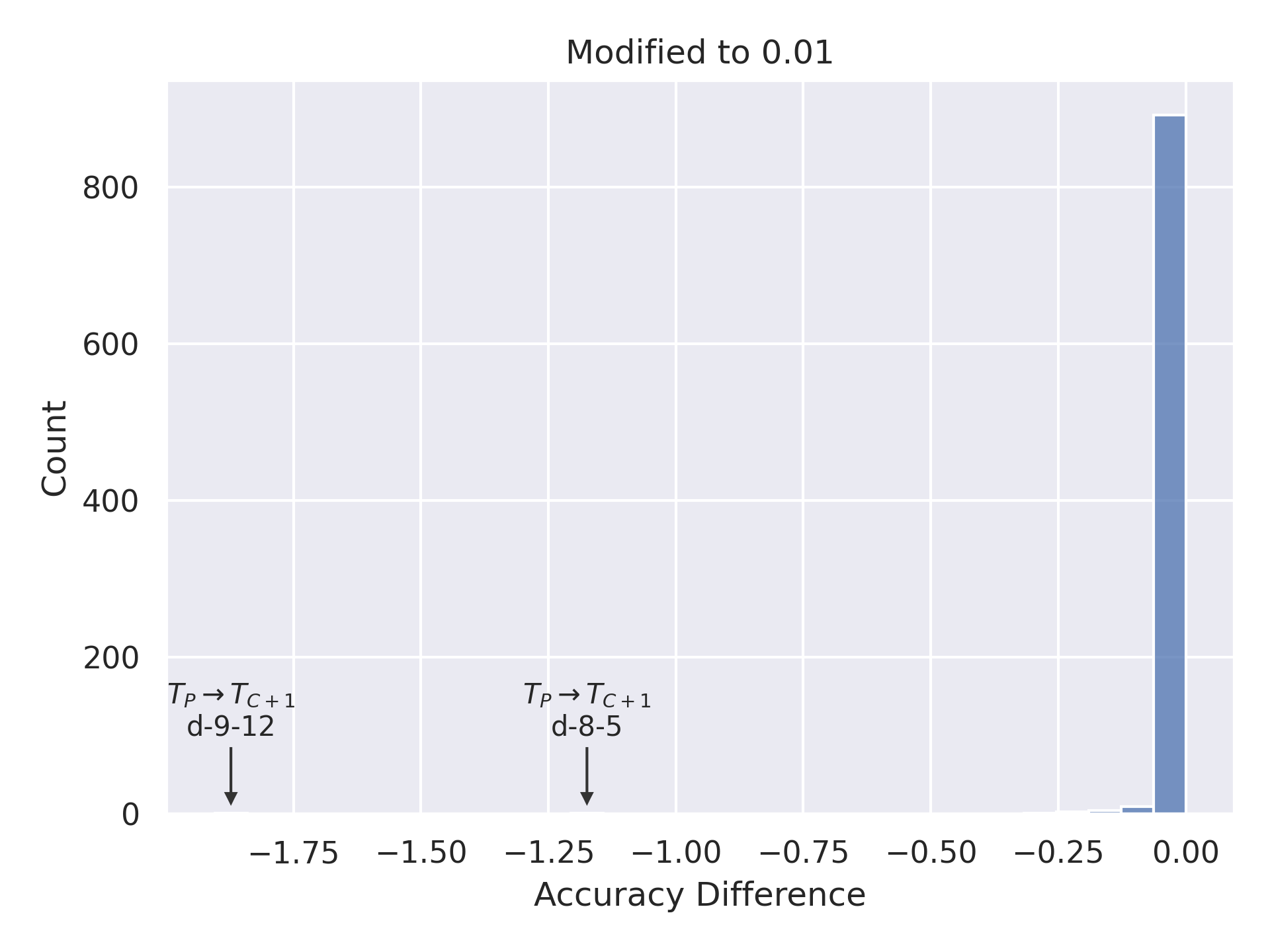}
    \end{subfigure}
    \begin{subfigure}{0.41\linewidth}
        \includegraphics[width=1\linewidth]{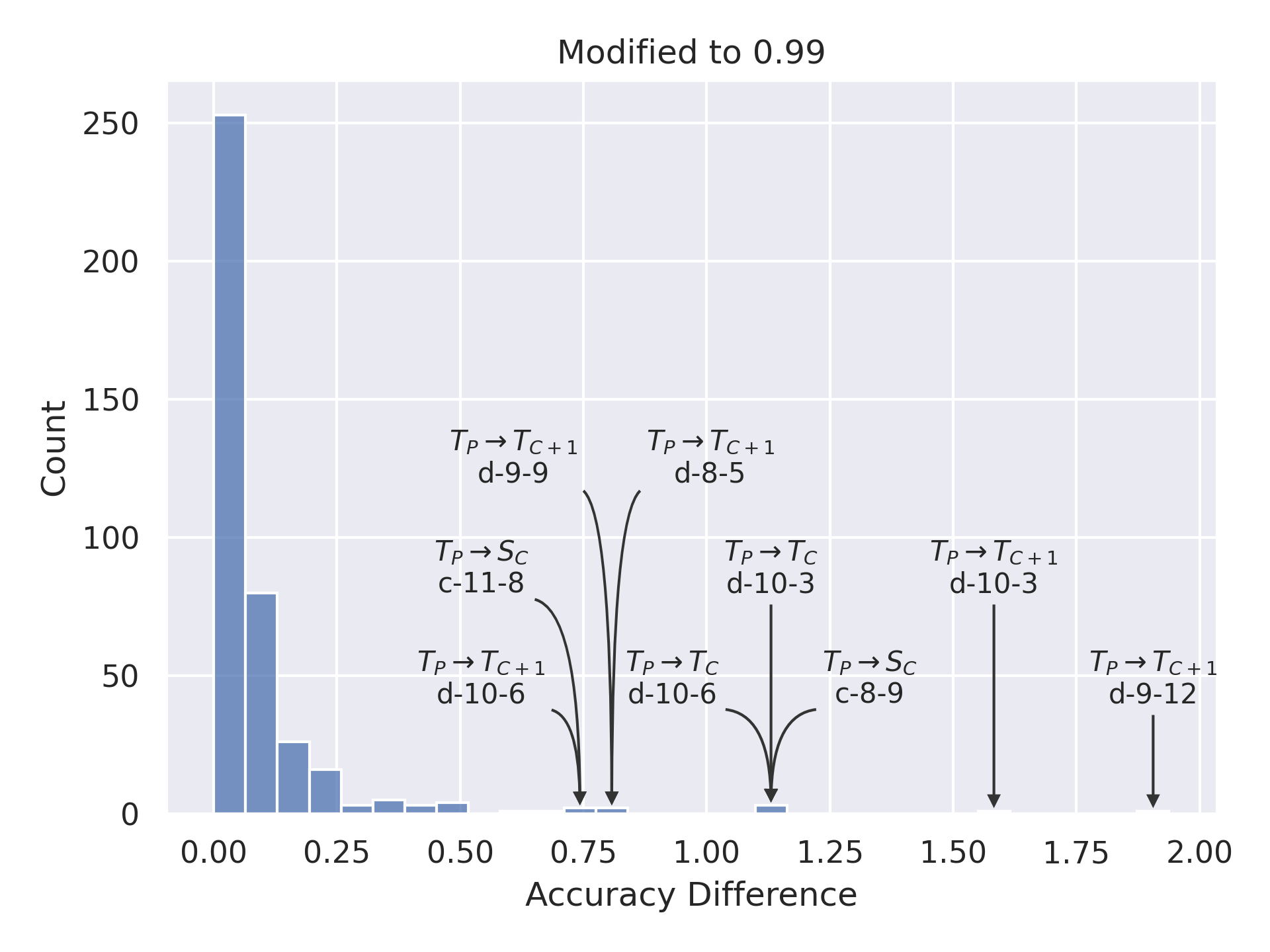}
    \end{subfigure}
    \caption{Histograms of average attention scores, correlations between attention scores and accuracy on \textbf{LCPT} (English-to-French) dataset, and the differences in accuracy for: Modifying Heads to $0.01$ (excluding positive values) and Modifying Heads to $0.99$ (excluding negative values) of each head of the \textbf{sentence-level NLLB-200} model with the values of noticeable heads annotated with arrows.}
    \label{fig:hist-nllb-enfr-sentence}
\end{figure*}

\begin{figure*}[!ht]
\center{}
    \begin{subfigure}{0.41\linewidth}
        \includegraphics[width=1\linewidth]{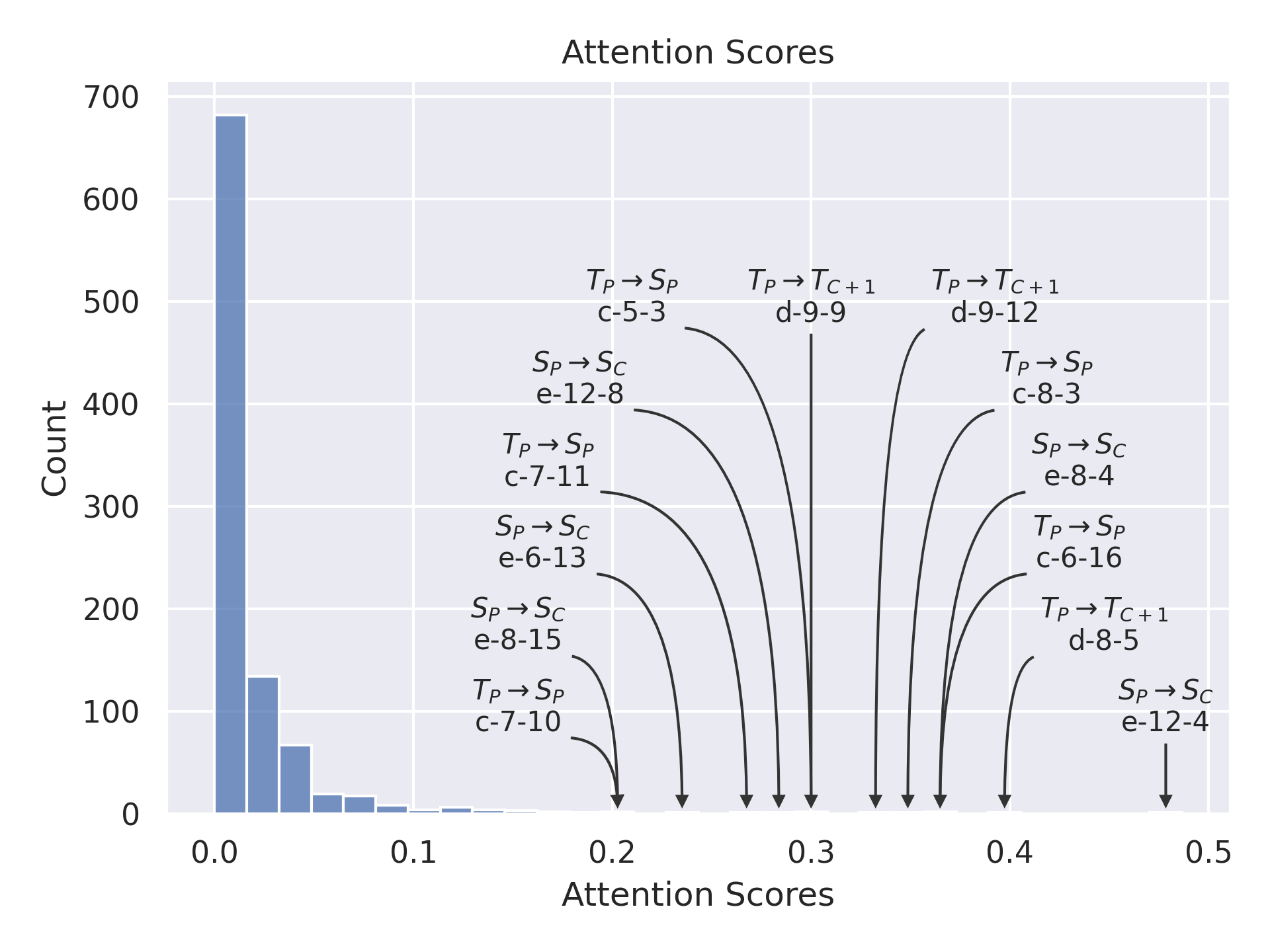}
    \end{subfigure}
    \begin{subfigure}{0.41\linewidth}
        \includegraphics[width=1\linewidth]{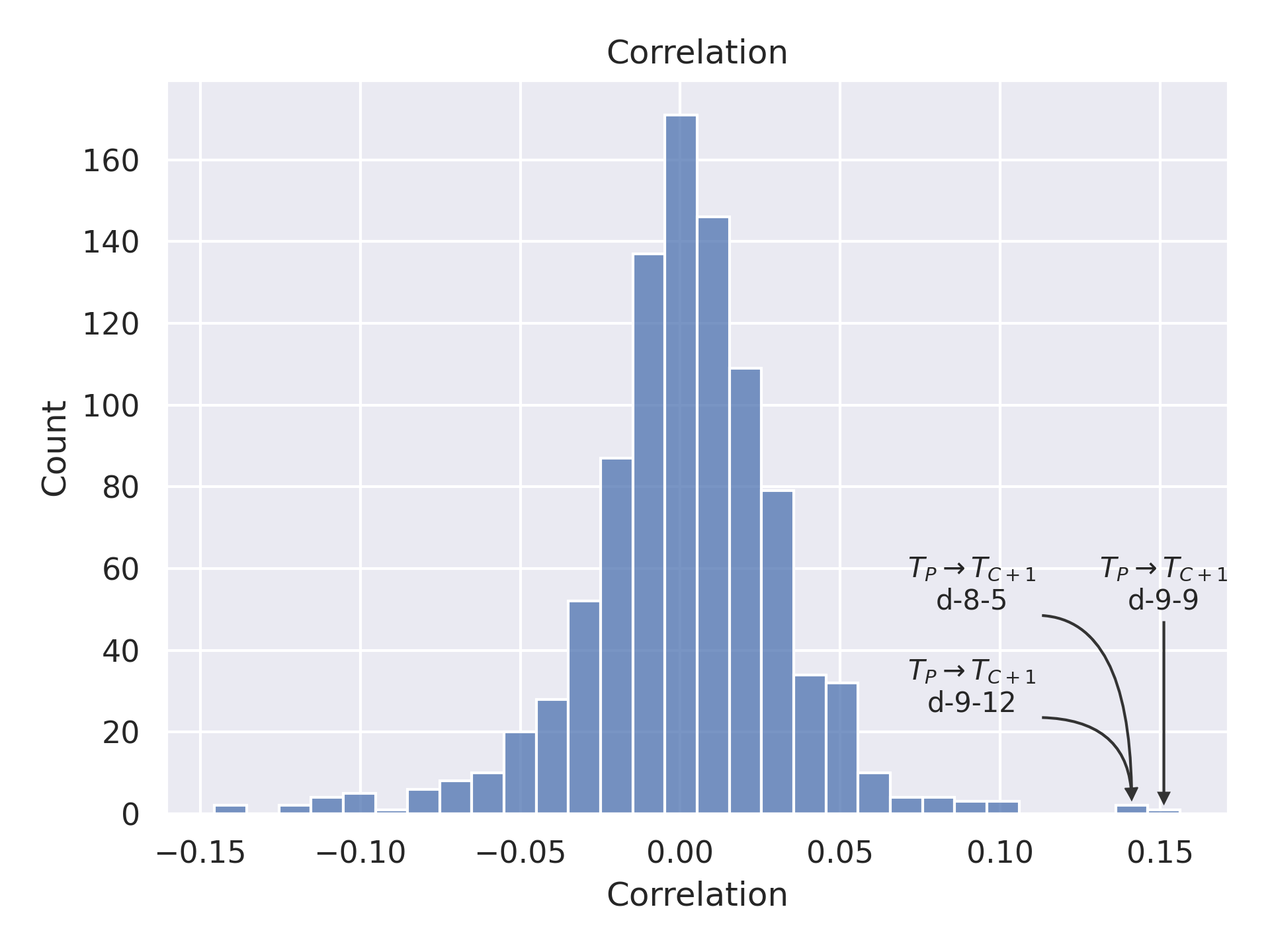}
    \end{subfigure}
    \begin{subfigure}{0.41\linewidth}
        \includegraphics[width=1\linewidth]{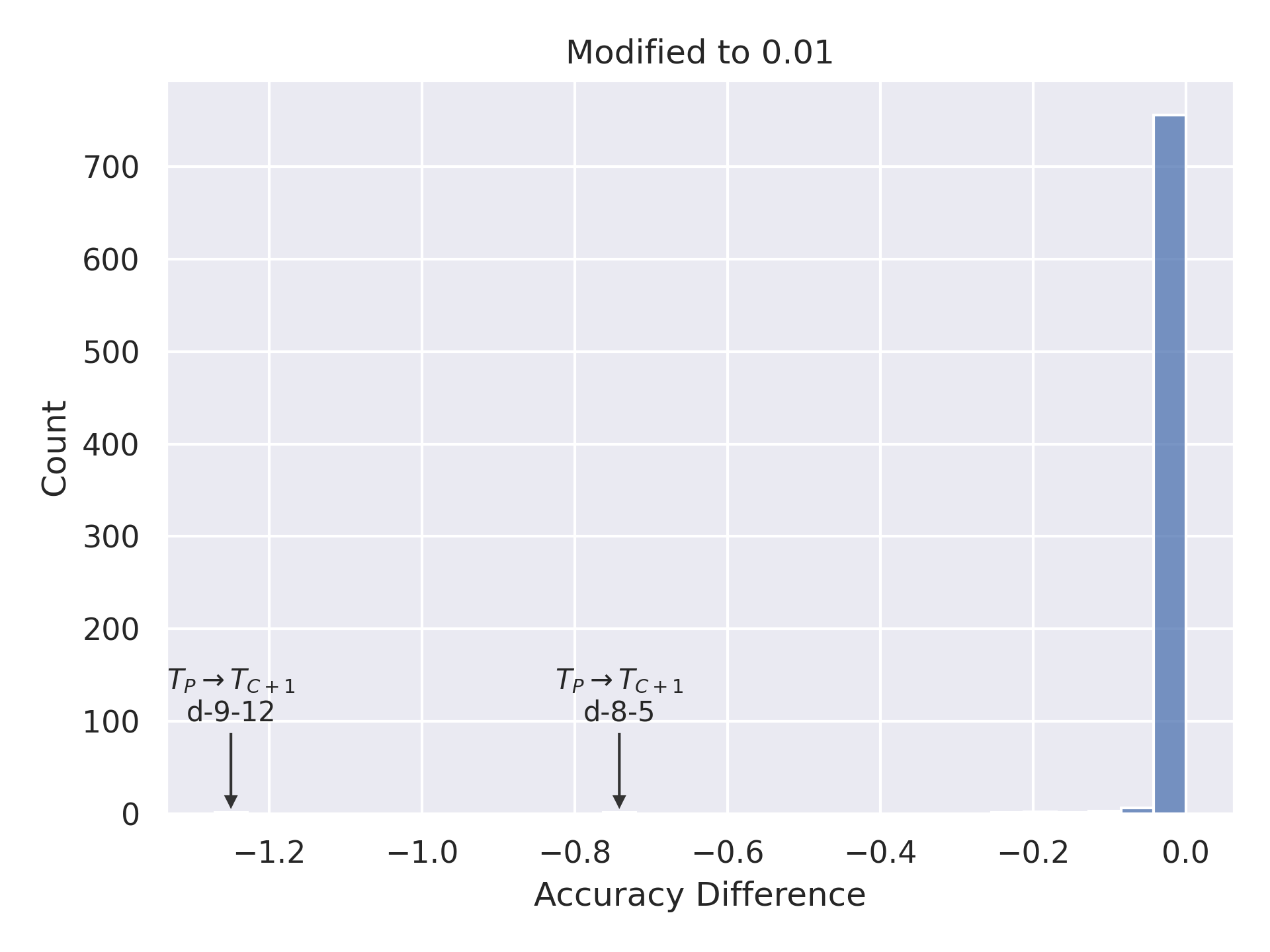}
    \end{subfigure}
    \begin{subfigure}{0.41\linewidth}
        \includegraphics[width=1\linewidth]{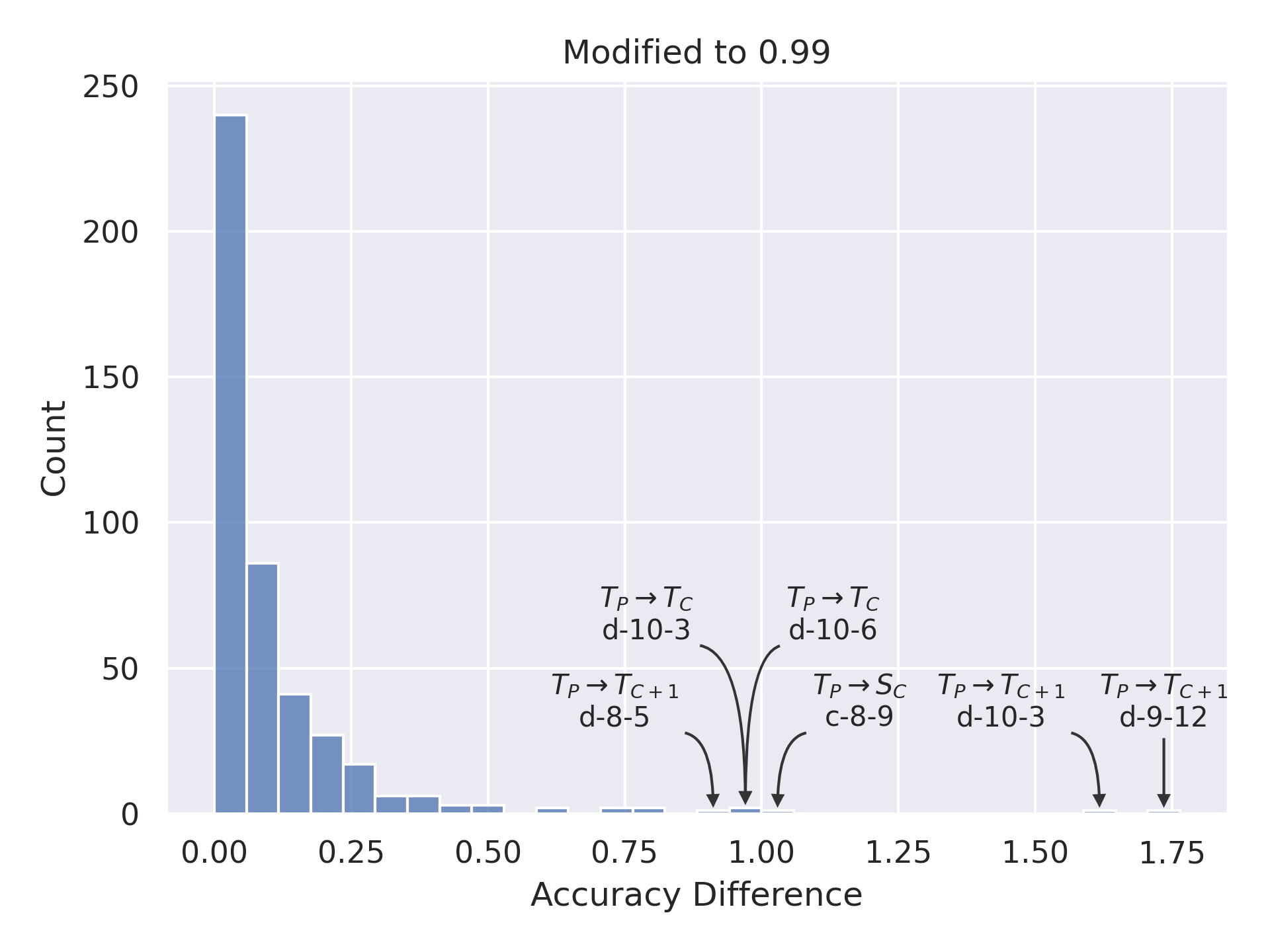}
    \end{subfigure}
    \caption{Histograms of average attention scores, correlations between attention scores and accuracy on \textbf{LCPT} (English-to-French) dataset, and the differences in accuracy for: Modifying Heads to $0.01$ (excluding positive values) and Modifying Heads to $0.99$ (excluding negative values) of each head of the \textbf{context-aware NLLB-200} model with the values of noticeable heads annotated with arrows.}
    \label{fig:hist-nllb-enfr-1}
\end{figure*}

\change{For each model, we generated histograms of the measured metrics and annotate with arrows the values for the most notable heads. Figures~\ref{fig:hist-opus-mt-sentence}, \ref{fig:hist-opus-mt-1}, and \ref{fig:hist-opus-mt-3} show the histograms for OpusMT en-de models (sentence-level, context-aware-1, and context-aware-3 respectively). The histograms for sentence-level NLLB-200 can be found in Figures~\ref{fig:hist-nllb-ende-sentence} and \ref{fig:hist-nllb-enfr-sentence} for the English-to-German and English-to-French directions respectively. The corresponding histograms of the context-aware NLLB-200 model are presented in Figures~\ref{fig:hist-nllb-ende-1} and \ref{fig:hist-nllb-enfr-1} for English-to-German and English-to-French respectively.}

\end{document}